\pdfoutput=1

\documentclass[10pt,twocolumn,letterpaper]{article}

\usepackage[pagenumbers]{cvpr} 

\newcommand{\red}[1]{{\color{red}#1}}

\definecolor{cvprblue}{rgb}{0.21,0.49,0.74}
\usepackage[pagebackref,breaklinks,colorlinks,allcolors=cvprblue]{hyperref}

\usepackage{siunitx}
\usepackage{multirow}
\usepackage{tikz}
\usepackage{orcidlink}

\usepackage{booktabs} 
\usepackage{xcolor}
\usepackage{import}
\usepackage{array}
\usepackage{multirow}
\usepackage{subcaption}
\usepackage{enumitem}
\usepackage{mathtools}
\usepackage[overlay]{textpos}

\definecolor{chatdark}{RGB}{24, 26, 27}

\def\paperID{} 
\def\confName{CVPR}
\def\confYear{2026}

\title{HelixTrack:\\Event‑Based Tracking and RPM Estimation of Propeller-like Objects}

\author{Radim Spetlik \quad Michal Pliska \quad Vojtěch Vrba \quad Jiri Matas\\
Faculty of Electrical Engineering\\
Czech Technical University in Prague
}

\newcommand{\tquadcopterego}[0]{Timestamped Quadcopter with Egomotion}
\newcommand{\tquadcopteregoshort}[0]{TQE}

\newcommand{\ok}{\textcolor{green!50!black}{\checkmark}}
\newcommand{\fail}{\textcolor{red!70!black}{$\times$}}

\newcommand{\helixtrack}[0]{HelixTrack}

\begin{document}
\maketitle
\begin{abstract}
Safety-critical perception for unmanned aerial vehicles 
and rotating machinery requires microsecond‑latency tracking of fast, periodic motion under egomotion and strong distractors. 
Frame‑ and event‑based trackers drift or break on propellers because periodic signatures violate their smooth‑motion assumptions. 
%
We tackle this gap with \helixtrack{}, a fully event-driven method that jointly tracks propeller-like objects and estimates their rotations per minute (RPM). 
Incoming events are back-warped from the image plane into the rotor plane via a homography estimated on the fly. 
A Kalman Filter maintains instantaneous estimates of phase. 
Batched iterative updates refine the object pose by coupling phase residuals to geometry. 
%
%
To our knowledge, no public dataset targets joint tracking and RPM estimation of propeller‑like objects. 
We therefore introduce the \tquadcopterego{} (\tquadcopteregoshort{}) dataset with 13\,high‑resolution event sequences, containing 52 rotating objects in total, captured at distances of 2\,m\,/\,4\,m, with increasing egomotion and microsecond RPM ground truth.  
On \tquadcopteregoshort{}, \helixtrack{} processes full-rate events \(\approx\!11.8\times\) faster than real time and microsecond latency. It consistently outperforms per‑event and aggregation‑based baselines adapted for RPM estimation.

\end{abstract}\vspace{-1em}   
\section{Introduction}
\label{sec:intro}


Tracking propellers and reading their instantaneous RPM is a practical capability for working with and around small aerial vehicles. 
In radio-denied or silent settings, blade phase and RPM provide a communication-free signal for leader-follower behavior, intent (spool-up/spool-down), and collision avoidance, and act as a diagnostic or identification cue when other channels fail. 
However, fast rotors produce dense, strictly periodic patterns and visually similar distractors that violate the motion assumptions of conventional frame-based and event-based trackers, motivating an event-native, propeller-aware solution.

In recent years, event-stream tracking has attracted increasing attention, with research advancing from feature-level tracking to single- and multi-object tracking, supported by new datasets and strong baselines \cite{li_3d_2024, messikommer_data-driven_2025, wang_event_2024, wang_spikemot_2025, wang_mambaevt_2025, sun_reliable_2024}.
%
\begin{figure}
    \def\svgwidth{\hsize}\import{fig/cvpr2026_l/}{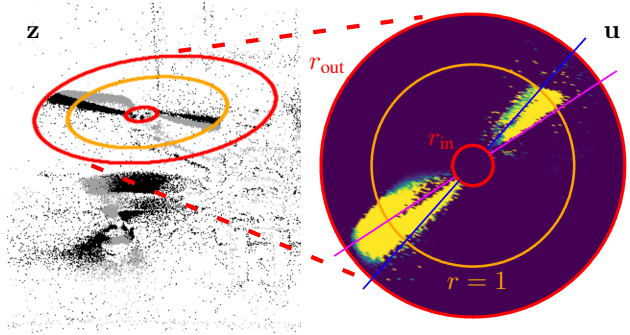}\caption{\emph{HelixTrack} -- tracking and instantaneous rotations per minute estimation of propeller-like objects from asynchronous events.
    Incoming events are back‑warped from the image plane ``$\mathbf{z}$'' into the rotor plane ``$\mathbf{u}$'' by a homography. 
    Rotor‑plane coordinates are gated with $r_\text{in}$ and $r_\text{out}$ to a ring‑shaped band and used for: (i) per-event phase update via Kalman filter, and (ii) batched pose refinement via iterative homography update.
}\label{fig:overview}\vspace{-1em}
\end{figure}
Two families of approaches have proven particularly effective. \emph{Per-event} methods process each event as it arrives and update the state continuously, yielding very low latency and strong robustness in high-speed regimes \cite{zafeiri_event-ecc_2024, wang_asynchronous_2024}. Examples include asynchronous continuous optimization over event constellations \cite{zafeiri_event-ecc_2024} and Kalman-filtered blob tracking that embraces native, event-driven timing \cite{wang_asynchronous_2024}. \emph{Aggregation-based} methods store events in spatiotemporal tensors --- event frames, time surfaces, or voxel grids --- which can be processed by modern deep networks and sequence models \cite{wang_event_2024, wang_mambaevt_2025, wang_spikemot_2025, sun_reliable_2024}.
Such aggregation enables the use of powerful learned backbones (e.g., transformers and state-space models) which have delivered competitive accuracy across single- and multi-object tracking benchmarks \cite{wang_event_2024, wang_mambaevt_2025, wang_spikemot_2025, sun_reliable_2024}. Thus, both paradigms have established themselves as viable choices for event-based tracking, trading off latency, modeling flexibility, and learning capacity.

However, an important set of phenomena remains under-explored: objects that exhibit fast, periodic motion. Rotating propellers are archetypal: they generate dense, structured, highly periodic event patterns that violate the ``smooth translational motion'' assumptions implicit in many trackers. 
Rapid movement of blades, strong distractors (\eg, in the case of multiple propellers), and repetitive textures challenge 
both per-event and aggregation-based object trackers, leading to drift and failures. 
Event-based rate estimation for periodic phenomena has been studied \cite{kolar_eeppr_2025,spetlik_efficient_2025}, but prior work focused on estimating frequency in static scenes where no tracking is required. 
To our knowledge, no prior work explicitly formulates and solves the tracking of objects exhibiting periodic motion while jointly estimating their rotation rate (RPM) from raw events.

We address this gap with \emph{HelixTrack}, an event-based framework for \emph{propeller tracking and RPM estimation}. 
Our key insight is that a rotating blade induces a \emph{helical signature} in the event space when lifted to appropriate spatiotemporal coordinates anchored on the rotor hub. 
\helixtrack{} is a \emph{fully event‑driven}, \emph{real‑time}, \emph{low-latency} framework for tracking and measuring spinning propeller-like objects with an event camera  (see Fig.~\ref{fig:overview}). 
\helixtrack{}:
(i) back‑warps each event through a homography onto the rotor plane while applying various hard and soft gates,  
(ii) uses an Extended Kalman Filter (EKF) to estimate the instantaneous phase, angular speed, and acceleration, and
(iii) iteratively updates the homography, minimizing a set of refined loss terms. 
The outcome is a stream of phase and pose updates from which RPM follows directly.
Conceptually, HelixTrack bridges both families of prior work: 
it runs natively \emph{asynchronously}, retaining the latency benefits of event cameras, 
yet its \emph{lifted, structured representation} admits aggregation for frame‑ or batch‑based pipelines.

We also overcome a common limitation of event-based tracking --- the absence of reliable ground-truth positions. 
Annotating the trajectories of rapidly spinning propeller hubs observed by a moving camera presents significant challenges and requires substantial time investment. 
To address this, we introduce a dataset that provides microsecond-level accurate ground-truth RPM measurements for each propeller, enabling evaluation through the mean absolute error of the predicted RPM, which is a dependable indicator of the tracker's precision.

Beyond drone propellers, the formulation is applicable to other periodic, radially symmetric actuators (e.g., fans, rotors), offering a principled path to track-and-measure behaviors that are challenging for conventional trackers.

In summary, our contributions are: 
\begin{enumerate}[label=(\roman*),leftmargin=1.5em]
\item \textbf{Problem formulation.} We specify the task and evaluation protocol for event-based tracking with joint RPM estimation of propeller-like objects exhibiting periodic, high-speed motion. This fits in-between the general event-based trackers \cite{zafeiri_event-ecc_2024, wang_asynchronous_2024, wang_event_2024, wang_spikemot_2025, wang_mambaevt_2025, sun_reliable_2024} and rate-only estimators that are not object-centric \cite{kolar_eeppr_2025,spetlik_efficient_2025}.
\item The \textbf{HelixTrack model.} We propose an event‑driven tracker that back‑warps events to the rotor plane via a homography, fits a helical phase model, and jointly infers RPM, with a per‑event EKF, and pose, using the Gauss-Newton (GN) algorithm on event batches.
\item The \textbf{TQE Dataset.} We collect 13 high‑resolution sequences captured at 2~m~/~4~m distances with seven egomotion levels, and synchronized IR‑RPM ground truth. 
\item \textbf{Benchmark.} We adapt asynchronous~\cite{wang_asynchronous_2024} and learned~\cite{messikommer_data-driven_2025} trackers for RPM estimation and show that \helixtrack{} achieves the best performance. 
\item \textbf{Ablations \& Runtime analysis.} We ablate loss terms, test tolerance to coarse pose/scale/RPM initialization, and profile speed --- processing at $\approx 11.8\times$ real time with accuracy/throughput trade‑offs under subsampling. 
\end{enumerate}


\noindent 
\helixtrack{} turns an adversarial motion pattern into a feature that can be modeled, localized, and measured.

\section{Related Work}
\label{sec:related}

The review of \emph{tracking} literature relevant to \helixtrack{} is organized by tracking paradigms and sensing modalities.

Event cameras natively support low-latency motion perception, which has motivated blob- and feature-level trackers as well as general single-object tracking (SOT) pipelines.
Asynchronous blob trackers operate directly on the event stream, i.e. without frames, maintaining object hypotheses with simple motion and shape cues \cite{messikommer_data-driven_2025, li_3d_2024, zafeiri_event-ecc_2024, wang_asynchronous_2024}. 
At the feature level, data-driven formulations learn to follow high-speed structures through the spatio-temporal event volume, improving robustness over hand-crafted rules \cite{messikommer_data-driven_2025, li_3d_2024}. 
Benchmarks for event-based visual object tracking (VOT) 
enable comparison across VOT methods \cite{wang_event_2024, wang_mambaevt_2025}. 

Tracking-by-detection and joint detection-tracking have also been explored in the event domain.
Recent work demonstrates end-to-end multi-object tracking on events with learned association and sparsity-aware motion models \cite{wang_spikemot_2025}.
%
Beyond local heuristics, sequence models capture long-range temporal dependencies in asynchronous data.
State-space models and transformer-style architectures have been adapted to event streams for VOT, producing stronger appearance modeling under severe motion and illumination change \cite{sun_reliable_2024, wang_mambaevt_2025}. 
%
Several approaches fuse events with intensity frames (or other modalities) to reduce failure modes of either sensor alone, typically with transformer-based fusion and hybrid feature pipelines \cite{sun_reliable_2024}. 
In contrast, our method is \emph{fully event-driven}, which preserves very low latency.

Event cameras have also been used for geometry-aware tracking --- e.g., 3D feature trajectories and object-grounded motion constraints --- which is particularly relevant when the target exhibits structured dynamics \cite{li_3d_2024}. 
Our formulation continues this line by embedding a physics/geometry prior in the tracker state.

An orthogonal thread optimizes a continuous-time objective directly over events for tracking and registration, offering drift control and accurate sub-frame localization \cite{zafeiri_event-ecc_2024}. 
We instead cast tracking as online filtering with an explicit dynamical phase model, which trades some global optimality for latency and simplicity.

Previous unmanned aerial vehicle (UAV) tracking work has focused on estimating drone position or pose rather than measuring propeller attributes. Frame-based methods often rely on standard cameras and CNN-based pipelines~\cite{vrba_marker-less_2020,zheng_deep_2024,zheng_keypoint-guided_2024}, providing velocities and accelerations with significant latency. Event-based techniques for high-speed UAV interception or localization~\cite{mitrokhin_event-based_2018,stewart_drone_2021, stewart_virtual_2022,sanket_evpropnet_2021} sometimes detect propeller patterns in real time at modest rates (e.g., 35\,Hz~\cite{sanket_evpropnet_2021}). However, they rarely localize individual propellers or measure angular velocity at microsecond temporal precision, and typically leverage low-resolution sensors, hampering performance in cluttered scenes or for small targets. In \cite{ciattaglia_measuring_2024} propeller speeds are measured but no tracking is performed.
\\
\textbf{Relation to our work.}
Unlike detection-oriented pipelines that classify or localize events on a per-frame basis, our tracker maintains the \emph{phase and pose} of a specific object continuously over time, using only asynchronous events.

\section{Method}
\label{sec:method}

\paragraph{High-level summary.}
Given an approximate pose and RPM from~\cite{spetlik_efficient_2025}, the tracker estimates two coupled quantities:  
(i) an image-to-rotor-plane homography (geometry), and  
(ii) a latent phase state $\mathbf{x}(t)=[\phi(t),\,\omega(t),\,\alpha(t)]^\top$ (phase and its first and second derivatives).  
Asynchronously arriving events are processed in two loops:
\begin{enumerate}
  \item \textbf{Per-event phase tracking.} An EKF updates $\mathbf{x}(t)$ from each event using a wrapped phase residual.
  \item \textbf{Batched pose refinement.} Periodically, Gauss-Newton (GN) updates homography parameters by minimizing a robust, weighted objective with analytic Jacobians.
\end{enumerate}
The EKF gives microsecond-scale phase and RPM tracking, while GN accumulates evidence to estimate pose.

\subsection{Problem Setup and Notation}
\label{sec:notation}
Camera observes an event stream $e_i=(\mathbf{z}_i, t_i, p_i)$ with pixel location $\mathbf{z}_i=(x_i,y_i)^\top$, timestamp $t_i$, and polarity $p_i\!\in\!\{+1,-1\}$.  
Vectors are bold; we reserve $\mathbf{x}$ for the phase state and $\mathbf{z}$ for image points.
\\
\textbf{Homography parameterization (image $\leftrightarrow$ rotor plane).}
A planar point $\mathbf{u}=(u_x,u_y)^\top$ in the rotor plane projects to the image via a $6$-DoF homography
\begin{align}\label{eq:homography}
\mathbf{H}(\mathbf{q})\;=&\;
\begin{bmatrix}
s\cos\psi & -\,s\sin\psi & t_x\\[2pt]
s\sin\psi & \ \ s\cos\psi & t_y\\[2pt]
p_{31} & \ \ p_{32} & 1
\end{bmatrix}\\ \label{eq:homography_q}
\mathbf{z} \;\sim\; \mathbf{H}(\mathbf{q})\,[\mathbf{u};1&],\quad
\mathbf{q}=\big[s,\,\psi,\,t_x,\,t_y,\,p_{31},\,p_{32}\big]^\top,
\end{align}
where $s>0$ (in-plane scale), $\psi$ (in-plane rotation), $(t_x,t_y)$ (translation), and $(p_{31},p_{32})$ (small perspective terms).  
Given an event location $\mathbf{z}_i$,
\[
\tilde{\mathbf{u}}_i=\mathbf{H}(\mathbf{q})^{-1}[\mathbf{z}_i;1],\quad
\mathbf{u}_i=\frac{1}{\tilde{u}_{i,3}}\begin{bmatrix}\tilde{u}_{i,1}\\ \tilde{u}_{i,2}\end{bmatrix},\quad
\frac{\partial \mathbf{u}_i}{\partial \mathbf{q}}\in\mathbb{R}^{2\times 6}.
\]

\subsection{Generative Phase Model}
We map each event to a common phase coordinate that is invariant to arbitrary in‑plane camera rotation and respects the $B$‑fold symmetry of a $B$‑blade rotor.
Let $\zeta\in\{\pm1\}$ be the \emph{rotation sign}, estimated from circular statistics over a short prefix of events.
The signed azimuth is
\[
\theta_\zeta(\mathbf{u})=\operatorname{atan2}(\zeta\,u_y,\,u_x)\in(-\pi,\pi].
\]
To absorb the image‑plane rotation $\psi$~(\ref{eq:homography_q}) into the rotor phase $\phi(t)$ (radians), we define the effective phase
\[
\phi_{\text{eff}}(t) = \phi(t) - B\,\psi.
\]
$\phi_{\text{eff}}$ is the tracked phase after accounting for $\psi$.
Given the timestep $t_i$ and the back-wraped position $\mathbf{u}_i$ of an event $e_i$, we form the wrapped \emph{phase residual}
\begin{equation}\label{eq:phase_residual}
\varepsilon_i^{\phi} \;=\; \operatorname{wrap}_\pi\!\big(\phi_\text{eff}(t_i)-B\,\theta_\zeta(\mathbf{u}_i)\big).
\end{equation}
where $\operatorname{wrap}_\pi(\theta)=\mathrm{mod}(\theta+\pi,2\pi)-\pi$. 
By construction, $\varepsilon_i^{\phi}$ is the $\pi$‑wrapped, signed phase error between $\phi_{\text{eff}}(t_i)$ and the blade phase implied by the event, $B\,\theta_\zeta(\mathbf{u}_i)$.
Ignoring wrap discontinuities, the residual is first‑order invariant to global in‑plane rotation:
\[
\frac{\partial \varepsilon_i^{\phi}}{\partial \psi} = \frac{\partial \phi_\text{eff}}{\partial \psi} - B\,\frac{\partial \theta_\zeta}{\partial \psi} = -B -B(-1) = 0,
\]
since a global rotation shifts all azimuths by $-\psi$ so that $\partial\theta_\zeta/\partial\psi=-1$.
\begin{figure*}[ht]
\def\svgwidth{\hsize}\import{fig/tqe/}{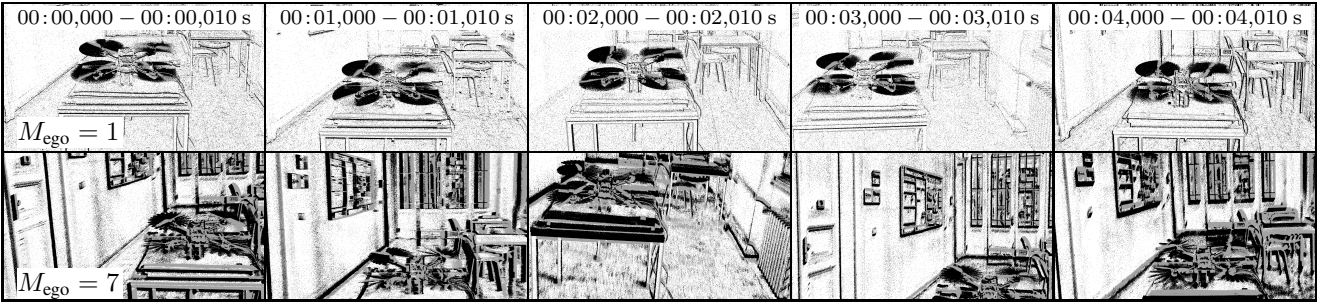}\caption{
Aggregated events from two sequences in the \tquadcopterego{} dataset. Top row: smallest egomotion ($M_{\text{ego}}=1$); bottom row: largest egomotion ($M_{\text{ego}}=7$). Events are aggregated from the first 10 ms of five 1-second intervals. 
}\label{fig:dataset_ego_sequenced_overview}\vspace{-1em}
\end{figure*}

\subsection{Phase-State Dynamics: Constant Acceleration with White Jerk}
\label{sec:phase-dyn}
We want to stay predictive between irregular event times. We therefore use a continuous‑time acceleration prior whose uncertainty grows with elapsed time.
Note that $t$,$t'$ are continuous.
By definition of angular kinematics,
\[
\omega(t)\!=\!\dot{\phi}(t),\quad
\alpha(t)\!=\!\dot{\omega}(t)\!=\!\ddot{\phi}(t),\quad
j(t)\!=\!\dot{\alpha}(t)\!=\!\dddot{\phi}(t),
\]
units $\omega$[rad/s], $\alpha$[rad/s$^2$], $j$[rad/s$^3$].
We model $j(t)$ as ``white jerk'', \ie, zero-mean white Gaussian noise with spectral density $q_{\text{jerk}}$, i.e.\ $\mathbb{E}[j(t)j(t')]=q_{\text{jerk}}\delta(t - t')$,
capturing torque fluctuations giving the continuous-time SDE
\[
\dot{\mathbf{x}}(t)\!=\!\mathbf{A}\,\mathbf{x}(t)+\mathbf{L}\,w(t), \quad w(t)\!\sim\!\mathcal{N}\!\left(0,\,q_{\text{jerk}}\delta(t - t')\right),
\]
\[
\mathbf{A} =
\begin{bmatrix}
0 & 1 & 0\\
0 & 0 & 1\\
0 & 0 & 0
\end{bmatrix}, \quad
\mathbf{L}=\begin{bmatrix}0\\0\\1\end{bmatrix}.
\]
Using $\mathbb{E}[j(t)j(t')]\!=\!q_{\text{jerk}}\delta(t-t')$ and integrating the process noise through the continuous‑time model yields $\mathbf{Q}_d(\Delta t_i)\!=\!\int_0^{\Delta t_i} [\tfrac{\tau^2}{2},\!\tau,\!1] q_{\text{jerk}} [\tfrac{\tau^2}{2},\!\tau,\!1]^\top d\tau$, \;$\Delta t_i\!=\!t_i\!-\!t_{i-1}$, evaluating to
\[
\mathbf{Q}_d
= q_{\text{jerk}} 
\begin{bmatrix}
\tfrac{\Delta t_i^5}{20} & \tfrac{\Delta t_i^4}{8} & \tfrac{\Delta t_i^3}{6}\\
\tfrac{\Delta t_i^4}{8} & \tfrac{\Delta t_i^3}{3} & \tfrac{\Delta t_i^2}{2}\\
\tfrac{\Delta t_i^3}{6} & \tfrac{\Delta t_i^2}{2} & \Delta t_i
\end{bmatrix}.
\]
We discretize the continuous-time state by the event times $t_i$.
Thus, $\mathbf{x}_{i-1}$ is the state immediately \emph{after} assimilating event $e_{i-1}$. It is propagated over $\Delta t_i$ to form
\[
\mathbf{x}_i=\mathbf{F}(\Delta t_i)\,\mathbf{x}_{i-1},
\]
\[
\mathbf{P}_i=\mathbf{F}(\Delta t_i)\,\mathbf{P}_{i-1}\,\mathbf{F}(\Delta t_i)^\top+\mathbf{Q}_d(\Delta t_i),
\]
\[
\mathbf{F}=\exp(\mathbf{A}\Delta t_i)=
\begin{bmatrix}
1 & \Delta t_i & \tfrac{1}{2}\Delta t_i^2\\
0 & 1 & \Delta t_i\\
0 & 0 & 1
\end{bmatrix}.
\]
\subsection{Extended Kalman Filter Update}
We gate events by a von Mises window
$w_{\text{vm}}(\varepsilon^\phi)=\exp\!\{\kappa(\cos(\varepsilon^\phi)-1)\}$, where $\kappa$ controls angular selectivity ($\kappa=0$ means no gating).  
We make the innovation variance heteroscedastic using the gates:
\[
R_i\;=\;\frac{\sigma_0^2}{\max\{w_\text{vm}(\varepsilon_i^{\phi})w_{\text{ring}}(r_i;\sigma_{\text{ring}}),\,\epsilon\}},
\]
where $\sigma_0^2$ is a nominal phase noise, $\varepsilon^\phi_i$ is the phase residual~(\ref{eq:phase_residual}),
and $\epsilon>0$ prevents division by zero.
A smooth spatial gate emphasizes the blade-tip neighborhood (normalized near $r=1$):
\[
w_{\text{ring}}(r;\sigma_{\text{ring}})
=\exp\!\left[-\tfrac{1}{2}\!\left(\tfrac{r-1}{\sigma_{\text{ring}}}\right)^2\right].
\]
The event $e_i$ yields the EKF update
\[
\mathbf{K}_i=\mathbf{P}_i \mathbf{C}^\top\big(\mathbf{C}\mathbf{P}_i \mathbf{C}^\top + R_i\big)^{-1},
\]
\[
\mathbf{x}_i\leftarrow\mathbf{x}_i-\mathbf{K}_i\,\varepsilon^\phi_i,\qquad \mathbf{P}_i\leftarrow(\mathbf{I}-\mathbf{K}_i\mathbf{C})\mathbf{P}_i,
\]
where $\mathbf{C}=[\,1\ \ 0\ \ 0\,]$.

\subsection{Gauss--Newton Pose Refinement}\label{sec:gn_loss}
We refine $\mathbf{q}$ on mini-batches of events $\mathcal{B}\subset\{1,\ldots,N\}$ using GN with analytic Jacobians.
Let $r_i=\|\mathbf{u}_i\|$. In addition to phase residual $\varepsilon_i^{\phi}$, we define the \emph{radial residual}
\[
\varepsilon_i^{r}=r_i-1,
\]
which pushes the average radius of projected events to the unit radius $r=1$.

Next, we add the \emph{polarity residual}
\[
\varepsilon_i^{\text{pol}} =
c_{\text{pol}}\,
\sin\!\Big(\tfrac{1}{2}\operatorname{wrap}_\pi\big(\varepsilon_i^{\phi} - \theta_{\text{pol}}(p)\big)\Big),
\]
to align the event polarity $p$ with the expected edge phase. $\theta_{\text{pol}}(+)=c_+$, $\theta_\text{pol}(-)=c_-$, and $c_\text{pol}$ are constants. 

We focus on an informative annulus $r_{\text{in}}\le r_i\le r_{\text{out}}$ (see Fig.~\ref{fig:overview}) and incorporate \emph{soft annulus residuals}
\[
\varepsilon_i^{\text{in}} = c_b\,\operatorname{softplus}_\tau(r_{\text{in}} - r_i),\quad
\varepsilon_i^{\text{out}} = c_b\,\operatorname{softplus}_\tau(r_i - r_{\text{out}}),
\]
with $\operatorname{softplus}_\tau(s)=\tau\log(1+\mathrm{e}^{s/\tau})$, ($\tau>0$),
and $c_b \in \mathbb{R}$ is a constant.

Since scale and perspective can drift when more events fall just \emph{outside} than \emph{inside} the expected tip radius, 
we add the Balanced Tip Occupancy (BTO) regularizer that encourages a balanced fraction of events around the unit radius \(r{=}1\).
We accumulate the event counts $C_{\text{core}}$, $C_{\text{in}}$, and $C_{\text{out}}$, which are the numbers of events in three narrow radial bands around the tip --- core at \(r=1\), inside at \(1-\Delta\), and outside at \(1+\Delta\).
Rate-invariant inside/outside fractions are
\[
p_{\text{in}}=\frac{C_{\text{in}}}{C_{\text{core}}+\varepsilon},\qquad
p_{\text{out}}=\frac{C_{\text{out}}}{C_{\text{core}}+\varepsilon},
\]
with $\epsilon > 0$. The \emph{BTO penalty} is
\[
\ell_{\text{BTO}}
=
\operatorname{softplus}_\tau\!\big(p_{\text{out}}-\overline{p}_{\text{out}}\big)
+\operatorname{softplus}_\tau\!\big(\underline{p}_{\text{in}}-p_{\text{in}}\big),
\]
$\overline{p}_{\text{out}}$ and $\underline{p}_{\text{in}}$ are constants.
Gradients flow through the radial channel, stabilizing scale and perspective at the blade tips.

\textbf{Objective function for pose estimation.}
Let $\rho_2(s)=\tfrac{1}{2}s^2$ and define event gates $g_i^\phi=\lambda_\phi\,w_i^\phi$, $g_i^r=\lambda_r\,w_i^r$ with
\[
w_i^\bullet \!=\! w_{\text{vm}}(\varepsilon_i^{\phi})\; w_{\text{ring}}(r_i;\sigma_{\text{ring}})\; w_{\text{Huber}}(\varepsilon_i^\bullet;c_\bullet),
\quad \bullet\in\{\phi,r\},
\]
where $w_{\text{Huber}}(\varepsilon;c)=\min\{1,\,c/|\varepsilon|\}$ with threshold $c>0$. The batch objective is
\begin{align}
&\mathcal{L}(\mathbf{q})
= \sum_{i\in\mathcal{B}}
\Big[\, g_i^\phi\,\rho_2\!\big(\varepsilon_i^\phi\big)
     + g_i^r\,\rho_2\!\big(\varepsilon_i^r\big) \,\Big]
\nonumber\\
&\quad
+ \lambda_{\text{pol}} \sum_{i\in\mathcal{B}} \rho_2\!\big(\varepsilon_i^{\text{pol}}\big)
+ \lambda_{\text{band}} \left(\sum_{i\in\mathcal{B}} \rho_2\!\big(\varepsilon_i^{\text{in}}\big)
+  \sum_{i\in\mathcal{B}} \rho_2\!\big(\varepsilon_i^{\text{out}}\big)\right)
\nonumber\\
&\quad
+ \lambda_{\text{BTO}}\,\ell_{\text{BTO}}(p_{\text{in}},p_{\text{out}})
+ \lambda_{\text{reg}}\,\ell_{\text{reg}}(\mathbf{q}).
\label{eq:loss}
\end{align}

\textbf{Interpretation.}
The first line ties phase to geometry (via $\varepsilon_i^\phi$) and centers the working ring at $r{=}1$ (via $\varepsilon_i^r$). The gates $w_{\text{vm}}$ and $w_{\text{ring}}$ downweight azimuthally inconsistent or off-band events, and $w_{\text{Huber}}$ reduces heavy-tailed outliers. The second line adds weak supervision (polarity alignment, soft band barriers). BTO inhibits degenerate “only-inside” or “only-outside” fits. Finally, $\ell_{\text{reg}}$ regularizes weakly observable directions (e.g., $\|p_{31}\|^2+\|p_{32}\|^2$).  
%
We choose $\lambda_\phi\!\approx\!1/\sigma_\phi^2$ and $\lambda_r\!\approx\!1/\sigma_r^2$, and set Huber thresholds $c_\phi,c_r$ at $1\!\sim\!2$ standard deviations.

\textbf{GN system and update.}
Linearizing the residuals in~\eqref{eq:loss} yields normal equations that we omit from the main paper for brevity
(see supplementary Sec.~\ref{sec:full_gn} for the full system). We solve the equations in the least-square sense with Gauss-Newton algorithm and update
\[
\mathbf{q} \leftarrow \mathbf{q} + \Gamma\,\Delta\mathbf{q},\quad
\Gamma=\operatorname{diag}(\gamma_{\text{scale}},\gamma_{\text{rot}},
\gamma_{\text{tx}},\gamma_{\text{ty}},\gamma_{\text{persp}},\gamma_{\text{persp}}),
\]
where $\Gamma$ sets the update step size for each homography parameter.
A pose update is triggered if 
$\big|(\phi-\phi_{\text{base}}-B(\psi-\psi_{\text{base}}))/B\big| \ge \pi$,
defining the latency of the pose estimation. Note that gradient descent in the direction of $\mathcal{L}(\mathbf{q})$ minimization is also possible, in this small 6‑DoF least‑squares tracker, Gauss-Newton is preferable. It exploits the residual/Jacobian structure to take curvature‑aware, parameter‑coupled steps. 

\subsection{RPM Estimation}\label{sec:rpm_estimation}
Instantaneous angular speed and RPM are
\[
\hat{\omega}=|\omega|,\qquad
\widehat{\mathrm{RPM}}=\tfrac{60}{2\pi}\hat{\omega},\qquad
\text{RPM}_\text{shaft}=\widehat{\mathrm{RPM}}/B,
\]
since $\omega=\dot{\phi}$.

\paragraph{Implementation Details}
We process events sequentially:
\begin{enumerate}
  \item \textbf{Predict} $(\mathbf{x}_i,\mathbf{P}_i)$ with $(\mathbf{F},\mathbf{Q}_d)$.
  \item \textbf{Update EKF} with $\varepsilon^\phi_i$ using gates $w_{\text{vm}},\,w_{\text{ring}}$.
  \item \textbf{Accumulate GN terms} using analytic $\partial\mathbf{u}/\partial\mathbf{q}$.
  \item \textbf{Pose step} Apply a batched GN update when triggered.
\end{enumerate}

\paragraph{Assumptions.}
(i) A single planar homography adequately models the rotor;  
(ii) the helical phase model is accurate within the informative ring.

\section{Experiments}\label{sec:experiments}

We first introduce the newly collected \tquadcopterego{} (\tquadcopteregoshort{}) dataset, then detail the evaluation protocol and metrics, and finally present experiments. Unless otherwise specified, results are reported using 4‑fold cross‑validation over the four propellers in \tquadcopteregoshort{} recordings (one training propeller per fold).

\subsection{Dataset}\label{sec:main_dataset}


\paragraph{\tquadcopterego{} (\tquadcopteregoshort{})} dataset contains 13 sequences of a quadcopter with four propellers, recorded at distances of 2 and 4 meters, containing 52 rotating objects in total. Sequences are captured with Metavision EVK4HD event camera. Each sequence shows the quadcopter moving progressively faster, starting with near-hovering to mimic light wind and progressing to rapid, shaky movements similar to drone racing. The level of egomotion of the camera is indicated by $M_\text{ego} \in \{1,\dots,7\}$, where higher values represent a greater intensity of motion (see Fig.~\ref{fig:dataset_ego_sequenced_overview}). Camera settings, such as aperture and focus, were tuned for varying distances. The Arduino-regulated propeller speeds were calibrated to a spiral path, simulating real flight, with motor speeds ranging from 30-80\% of maximum. An infrared sensor aimed at reflective tape on the rotor measured the values, with its analog output passed to the camera's trigger input. In all recordings, the quadcopter is mounted on a table and the camera is held in hand.


\begin{table*}[ht]
    \centering
\fontsize{9.0pt}{10.0pt}\selectfont
\begin{tabular}{ c @{ } c @{ } c r@{ }l r r @{\hspace{1em}} r r @{\hspace{1em}} r @{\hspace{1em}} r r @{\hspace{1em}} r r}
\toprule
& & & & & \multicolumn{3}{c}{MArE ($\%$) $\downarrow$} & \multicolumn{3}{c}{MAE (RPM) $\downarrow$} & \multicolumn{3}{c}{RMSE (RPM) $\downarrow$} \\

\cmidrule(l){6-8}\cmidrule(l){9-11}\cmidrule(l){12-14}
$d$ & $M_{\text{ego}}$ & $l$ (s) & \multicolumn{2}{c}{propeller} & ours & \footnotesize AEB~\cite{wang_asynchronous_2024} & \footnotesize DeepEv~\cite{messikommer_data-driven_2025} & ours & \footnotesize AEB~\cite{wang_asynchronous_2024} & \footnotesize DeepEv~\cite{messikommer_data-driven_2025} & ours & \footnotesize AEB~\cite{wang_asynchronous_2024} & \footnotesize DeepEv~\cite{messikommer_data-driven_2025} \\

\midrule
\multirow{12}*{$2$ m\hspace{0.3em}} & \multirow{4}*{1} & \multirow{4}*{28.7} & rear & right & \bfseries 1.3 & 61.4 & 8564.8 & 93.7 & 5980.8 & 652423.7 & 215.9 & 7186.8 & 785896.2 \\
 &  &  & rear & left & 1.3 & \bfseries 0.9 & 402.4 & 81.4 & 73.0 & 32090.3 & 241.4 & 241.2 & 46466.0 \\
 &  &  & front & left & \bfseries 1.5 & 35.3 & 4132.0 & 118.4 & 2487.0 & 303141.3 & 194.1 & 3302.9 & 359054.3 \\
 &  &  & front & right & \bfseries 1.2 & 65.5 & 94.6 & 77.8 & 5650.8 & 8759.2 & 186.3 & 7673.3 & 9700.4 \\
\cmidrule{2-14} 
 & \multirow{4}*{3} & \multirow{4}*{27.7} & rear & right & \bfseries 1.2 & 1198.7 & 5013.6 & 85.1 & 102482.5 & 386360.6 & 223.2 & 192935.8 & 437954.9 \\
 &  &  & rear & left & 31.7 & \bfseries 15.0 & 2059.8 & 2137.8 & 1696.4 & 151319.8 & 4518.8 & 3482.9 & 184206.2 \\
 &  &  & front & left & \bfseries 4.2 & 467.5 & 5236.2 & 230.2 & 37691.4 & 376231.1 & 763.8 & 81016.3 & 448776.7 \\
 &  &  & front & right & \bfseries 4.8 & 1347.9 & 195.7 & 267.9 & 110056.2 & 14595.8 & 1031.9 & 202948.8 & 18910.5 \\
\cmidrule{2-14}
 & \multirow{4}*{5} & \multirow{4}*{28.8} & rear & right & \bfseries 62.3 & 531.3 & 106.7 & 6347.2 & 41415.3 & 9244.0 & 7842.7 & 79321.0 & 10241.1 \\
 &  &  & rear & left & \bfseries 100.9 & 480.0 & 117.9 & 9689.7 & 30491.9 & 10624.2 & 10899.5 & 67097.7 & 12239.0 \\
 &  &  & front & left & \bfseries 65.0 & 389.8 & 99.3 & 5045.3 & 33458.3 & 8963.0 & 6266.7 & 69659.2 & 9836.3 \\
 &  &  & front & right & \bfseries 97.7 & 1122.0 & 398.5 & 8871.4 & 77151.0 & 32523.2 & 9710.6 & 111267.4 & 56823.0 \\
\midrule
\multirow{12}*{$4$ m\hspace{0.3em}} & \multirow{4}*{1} & \multirow{4}*{28.7} & rear & right & \bfseries 73.1 & 285.5 & 834.5 & 6616.3 & 21029.7 & 57801.3 & 8517.8 & 46759.1 & 73901.3 \\
 &  &  & rear & left & \bfseries 1.2 & 57.2 & 89.6 & 76.4 & 6038.7 & 8341.9 & 158.4 & 20115.0 & 9422.8 \\
 &  &  & front & left & \bfseries 1.4 & 48.9 & 88.7 & 104.8 & 4676.4 & 7853.7 & 255.3 & 16769.4 & 8925.9 \\
 &  &  & front & right & \bfseries 1.3 & 165.4 & 835.4 & 87.8 & 13041.8 & 58054.3 & 319.4 & 36853.1 & 74833.5 \\
\cmidrule{2-14}
 & \multirow{4}*{3} & \multirow{4}*{28.9} & rear & right & 70.9 & \bfseries 58.8 & 95.8 & 5770.0 & 5904.9 & 9773.3 & 7008.1 & 11479.0 & 10725.7 \\
 &  &  & rear & left & \bfseries 1.3 & 46.8 & 1801.4 & 85.0 & 3960.0 & 140787.4 & 211.7 & 4845.4 & 181653.8 \\
 &  &  & front & left & \bfseries 1.3 & 303.1 & 1937.2 & 109.4 & 21349.0 & 146814.9 & 210.8 & 51357.9 & 174482.0 \\
 &  &  & front & right & 79.8 & \bfseries 39.7 & 97.4 & 7832.8 & 4091.3 & 9594.5 & 9394.6 & 5306.3 & 10465.4 \\
\cmidrule{2-14}
 & \multirow{4}*{5} & \multirow{4}*{27.2} & rear & right & \bfseries 95.3 & 994.6 & 97.8 & 9688.6 & 83816.5 & 9942.3 & 10614.2 & 127547.1 & 10817.5 \\
 &  &  & rear & left & \bfseries 93.9 & 1043.3 & 94.8 & 9316.3 & 76882.9 & 9332.0 & 10400.6 & 133235.2 & 10351.1 \\
 &  &  & front & left & \bfseries 1.7 & 1092.8 & 3339.1 & 119.1 & 81598.8 & 260136.5 & 241.1 & 126119.1 & 310936.4 \\
 &  &  & front & right & \bfseries 88.3 & 1126.9 & 95.6 & 8752.7 & 93336.6 & 9351.7 & 9711.9 & 134050.9 & 10281.2 \\
\bottomrule
\end{tabular}\caption{Benchmark on selected recordings of \tquadcopterego{} (\tquadcopteregoshort{}). 
    For each presented method and setting (camera-target distance $d$, egomotion level $M_\text{ego}$), the average shaft-RPM mean absolute relative error (MArE), mean absolute error (MAE), and root mean squared error (RMSE) are reported over 4‑fold cross‑validation across the four propellers (one training propeller per fold).
    Both AEB-Tracker~\cite{wang_asynchronous_2024} and DeepEv~\cite{messikommer_data-driven_2025} methods were modified to allow RPM estimation (see Sec.~\ref{sec:existing_trackers_modifications}). \tquadcopteregoshort{} ground-truth RPM has $\mu\pm\sigma\!=\!11193\pm3540$, $\min$ $917$, $\max$ $15768$.
    Duration $l$ of recording reported in seconds.
    $M_\text{ego} = 1$ roughly means ``slow camera movement'', while $M_\text{ego} = 5$ means ``fast camera movement'' (see Fig.~\ref{fig:dataset_ego_sequenced_overview}). 
    As $M_\text{ego}$ increases, the camera experiences stronger egomotion, leading to increase of the number of events generated in the background.
    Complete results are provided in supplementary Table~\ref{tab:benchmark_complete}.
    }\label{tab:benchmark}\vspace{-1em}
\end{table*}

\subsection{Accuracy metrics}

We evaluate a tracker by the mean absolute error (MAE), mean absolute relative error (MArE), and root mean squared error (RMSE) between its instantaneous shaft speed and the infrared (IR) sensor ground truth. The tracker internally estimates angular speed ($\omega(t)$) (rad/s) (Sec.~\ref{sec:rpm_estimation}), and we convert it to RPM and to shaft RPM.
The IR setup produces time-synchronized, microsecond-accurate shaft-speed measurements for each propeller (the comparator output is synchronized with the camera clock), which we treat as the ground truth ($\mathrm{RPM}_{\text{GT}}(t)$). We compute the per-sequence errors as
\[
\mathrm{MAE}_{\text{seq}}
=\frac{1}{|\mathcal{T}|}\sum_{t\in\mathcal{T}}
\left|\widehat{\mathrm{RPM}}_{\text{shaft}}(t)-\mathrm{RPM}_{\text{GT}}(t)\right|,
\]
\[
\mathrm{MArE}_{\text{seq}}
=\frac{1}{|\mathcal{T}|}\sum_{t\in\mathcal{T}}
\frac{\left|\widehat{\mathrm{RPM}}_{\text{shaft}}(t)-\mathrm{RPM}_{\text{GT}}(t)\right|}{\left|\mathrm{RPM}_{\text{GT}}(t)\right|},
\]
\[
\mathrm{RMSE}_{\text{seq}}
=\sqrt{\frac{1}{|\mathcal{T}|}\sum_{t\in\mathcal{T}}
\left(\widehat{\mathrm{RPM}}_{\text{shaft}}(t)-\mathrm{RPM}_{\text{GT}}(t)\right)^{2}},
\]
where ($\mathcal{T}$) is the set of tracker update times (asynchronous for per‑event methods, discrete for aggregation baselines). Unless stated otherwise, all results are reported as shaft RPM.

\subsection{Benchmarking against existing trackers}\label{sec:existing_trackers_modifications}
In our benchmarks, we include the asynchronous blob tracker~\cite{wang_asynchronous_2024} and the data‑driven feature tracker~\cite{messikommer_data-driven_2025} because they represent complementary, strong event‑native baselines that map directly to rotor‑RPM estimation with minimal changes. The tracker~\cite{wang_asynchronous_2024}~operates purely asynchronously on raw events and tracks compact ``event blobs''. In contrast, the deep tracker~\cite{messikommer_data-driven_2025} produces high‑rate point trajectories. Both methods have mature public implementations, are designed for fast, high‑dynamic‑range motion, and together bracket the model‑based vs.\ learned design space, enabling comparisons with minimal adaptations to the propeller-tracking setting of RPM estimation. 
The next two sections detail these task-specific modifications.

\subsubsection{AEB-Tracker modifications}\label{sec:aebtracker_modifications}
We adapted the original event tracker~\cite{wang_asynchronous_2024} to rotor targets by (i) using an ellipse-aligned gate for event association and (ii) estimating RPM directly from the tracked ellipse orientation. Specifically, for an event at $(c,r)$ and track $i$ with center $(z_{x_i},z_{y_i})$, semi-axes $(\lambda_{1,i},\lambda_{2,i})$ and orientation $\theta_i$, we compute the ellipse-frame residual
\[
\begin{bmatrix}e_x\\ e_y\end{bmatrix}
=
\begin{bmatrix}\cos\theta_i & \sin\theta_i \\[2pt] -\sin\theta_i & \cos\theta_i\end{bmatrix}
\begin{bmatrix}c-z_{x_i}\\ r-z_{y_i}\end{bmatrix},
\]
\[
d_i=\sqrt{\Big(\tfrac{e_x}{\lambda_{1,i}}\Big)^2+\Big(\tfrac{e_y}{\lambda_{2,i}}\Big)^2},
\]
and accept the event if $d_i<\tau$ (adaptive $\tau$). This replaces the original Euclidean pixel gate, making the association geometry-aware for slender blades.
We also initialize $(z_x,z_y,\lambda_1,\lambda_2)$ to match the rotor footprint.

\emph{RPM from orientation.} The filter tracks the ellipse orientation $\theta(t)\!\in\![0,\pi)$. We unwrap it with
$\Delta\theta_k=\mathrm{wrap}_{(-\pi/2,\pi/2]}(\theta_k-\theta_{k-1})$ and $\Phi_k=\Phi_{k-1}+\Delta\theta_k$.
Whenever $|\Phi|$ increases by $2\pi$ between times $t_1$ and $t_2$ we update an instantaneous RPM estimate
\[
\widehat{\mathrm{RPM}}=\frac{B\cdot 60}{\,t_2-t_1\,},
\]
where the factor $B$ compensates the $\pi$-periodicity of $\theta$. 
Equivalently, $\widehat{\mathrm{RPM}}(t)=\frac{60}{\pi}\,\dot\theta(t)$ using the unwrapped $\theta$.

In addition to the task-specific changes described above, we tuned the tracker’s parameters one by one. After qualitatively confirming that this setup produced correct tracks for about $10$\,ms at the start of each sequence, we ran a grid search around these values (see supplementary Sec.~\ref{sec:aeb_tracker_grid_search}).

\subsubsection{DeepEv Tracker Modifications}
Given DeepEv~\cite{messikommer_data-driven_2025} tracklets $\mathcal{T}\!=\!\{(t_i,\mathbf{z}_i)\}_{i=1}^N$ with $\mathbf{z}_i\!=\!(z_{x_i},z_{y_i})$, we adopt the same EKF used by \helixtrack{} for phase estimation and reuse the homography $\mathbf{H}\!\in\!\mathbb{R}^{3\times3}$ estimated by \helixtrack{}. To preserve the estimated perspective and scale while anchoring the warp to the DeepEv image-plane measurements, we overwrite the translation entries of $H$ for each observation $\mathbf{z}_i$:
\begin{align}
\mathbf{H}_{13} &\leftarrow z_{x_i}\, \mathbf{H}_{33},\\
\mathbf{H}_{23} &\leftarrow z_{y_i}\, \mathbf{H}_{33}.
\end{align}
This keeps the projective scale $\mathbf{H}_{33}$ fixed and injects the DeepEv positions into the homography. Our rationale is that, even when our homography estimate degrades in some recordings, accurate DeepEv tracklets should produce results comparable to those of \helixtrack{}.

We could not fine-tune the model (no ground-truth tracks or camera poses are available), and we could not modify the method due to its deep network nature. We used the EDS-fine-tuned weights, which showed gains on higher-motion data in~\cite{messikommer_data-driven_2025}. We qualitatively confirmed the correctness of our modifications on the first 10ms of recordings.

\subsection{Benchmark}
All three methods were initialized with the same starting positions and scale estimates provided  by~\cite{spetlik_efficient_2025}.
In deployment, initialization can be performed by a standard RGB camera, it does not require a high frame rate or low latency. When a tracker fails, \ie, begins to report a wrong RPM, we do not reinitialize. \helixtrack{} parameters were found by element-wise descent and a grid search around these values (see supplementary Sec.~\ref{sec:helixtrack_params}). In every fold, the parameters that minimized the training error of \helixtrack{} were identical, unlike those of AEB-Tracker.

\begin{table}[ht]
    \setlength{\tabcolsep}{5pt}
\fontsize{9.6pt}{11.3pt}\selectfont
\begin{tabular}{c @{ } r@{$\pm$}l r@{$\pm$}l r@{$\pm$}l}
\toprule
Event & \multicolumn{2}{r}{} & \multicolumn{2}{r}{} & \multicolumn{2}{r}{} \\
Stride & \multicolumn{2}{c}{MAE (RPM) $\downarrow$} &  \multicolumn{2}{c}{$\text{RTF}_\text{est}$ $\uparrow$} &  \multicolumn{2}{c}{$U_\text{evt}/U_\text{GN}$} \\ 
\midrule
1 & ${105.6}$&${48.8}$ & ${13.6}$&${3.6}$ & ${775.00}$&${249.17}$ \\ 
2 & ${1382.2}$&${3277.8}$ & ${30.3}$&${12.5}$ & ${1520.84}$&${1477.22}$ \\ 
3 & ${1339.7}$&${2933.5}$ & ${42.8}$&${12.2}$ & ${1081.24}$&${854.28}$ \\ 
4 & ${2492.1}$&${3520.2}$ & ${53.1}$&${22.8}$ & ${559.09}$&${645.35}$ \\ 
\bottomrule
\end{tabular}\caption{Runtime-accuracy trade‑off for event subsampling. We report mean $\pm$ standard deviation across 4-fold cross-validation on the 23 successfully tracked propellers (MAE $\leq$ 300 RPM) from \tquadcopterego{} dataset. Metrics: Mean Absolute Error (MAE, RPM; lower is better), the estimated real‑time factor $\text{RTF}_\text{est}$ (higher is better) computed as sequence duration divided by the total compute time (per‑event EKF updates plus batched Gauss-Newton (GN) solves), and the mean per‑update costs $U_\text{evt}/U_\text{GN}$ for a single event update and a single GN pose solve, respectively. ``Event stride = k'' means only every k-th event is processed. See Sec.~\ref{sec:runtime_analysis} for the details and definitions.}\label{tab:speed}
\end{table}

Despite our best‑effort adaptations for RPM estimation, the results in Table~\ref{tab:benchmark} show that the strongly periodic propeller setting, outside the intended design space of AEB-Tracker and DeepEv, leads to very high failure rates for both methods. 
The best performance is achieved by \helixtrack{}. 
As expected, errors grow with stronger egomotion $M_\text{ego}$ and with larger capture distance $d$. 
Increasing $M_\text{ego}$ raises the number of background events, further stressing the trackers.
In the sequences where \helixtrack{} fails, we observe abrupt camera movements that coincide with a very low rotation speed ($\approx$ 1352 RPM). 
Under these conditions, the phase advance needed to trigger a pose update --- half a shaft rotation for a two‑blade rotor (i.e., $2\pi$ in helical phase) --- accumulates too slowly relative to the image‑plane motion, so the pose estimate lags and the track can break.

Visualization of \helixtrack{} tracklets and final projections of per-propeller unit circles is available in Fig.~\ref{fig:tracklets}, complete results and all parameters in supplementary Sec.~\ref{sec:benchmark_complete_results} and Table~\ref{tab:benchmark_complete}, and qualitative results in the supplementary video, the \texttt{C++} implementation in the supplementary codes.

\begin{figure*}[ht]
\def\svgwidth{\hsize}\import{fig/tracklets_img/}{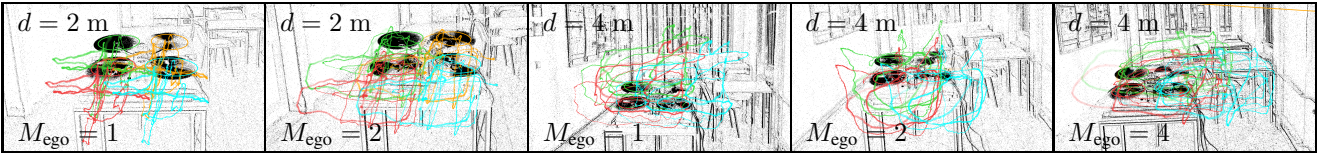}\caption{
Projection of per‑propeller unit circles under the final homography estimate, overlaid with HelixTrack trajectories on selected \tquadcopterego{} sequences. Propellers are color‑coded: rear left green, rear right orange, front left red, front right cyan. Markers encode time: size increases and opacity decreases backward in time, so older states appear larger and more transparent.
}\label{fig:tracklets}
\end{figure*}

\subsection{Runtime Analysis}\label{sec:runtime_analysis}
The \emph{event stride} means processing every $k$-th event (stride~=~1,2,3,4). The $\text{RTF}_\text{est}$ is computed as
\[
\text{RTF}_{\text{est}} \;=\; \frac{\sum_s \text{duration}_s}{U_{\text{evt}} \;+\; U_{\text{GN}}},
\]
\ie, the ratio of total sequence time to the sum $U_{\text{evt}}$ of all per‑event update times and the sum $U_{\text{GN}}$ of all GN‑solve times (all in $\mu$s). We also report $U_\text{evt}/U_\text{GN}$ exposing the cost profile of the loops. All runtime experiments were executed single-threaded on machines equipped with AMD EPYC 7543 processors. The single-propeller tracker used under $20$ MB of memory, including all benchmark storage routines.

The results reported in Table~\ref{tab:speed} are computed over the 23 successfully tracked propellers (MAE $\leq$ 300 RPM) from \tquadcopteregoshort{} (Sec.~\ref{sec:main_dataset}), indicated by \ok{} in supplementary Table~\ref{tab:benchmark_complete}. Processing all events (stride~=~1) achieves $\text{RTF}_\text{est} \approx 13.6 \pm 3.6$ (full dataset $\text{RTF}_\text{est} \approx 11.8 \pm 4.4$), which is comfortably faster than real time, with MAE = $105.6 \pm 48.8$ RPM. Subsampling increases throughput, but degrades accuracy.

\begin{figure}
    \def\svgwidth{\hsize}\import{fig/init_ablation/}{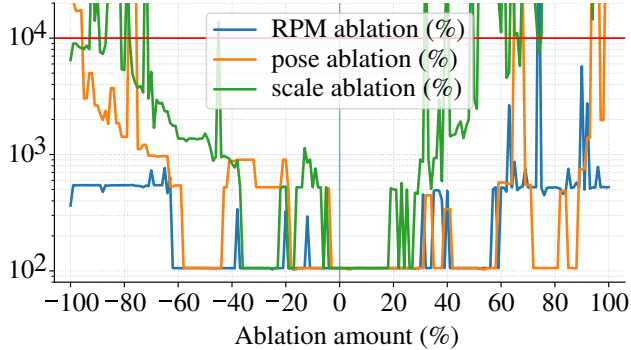}\caption{Robustness to initialization errors. We perturb the initial tracker position, in‑plane scale, and initial RPM by $\pm \{1,2,\dots,100\}\%$ around the nominal initialization and run \helixtrack{} end‑to‑end. For position, we average shifts ``toward'' and ``away from'' the nearest distractor (propeller). The plot reports shaft‑RPM Mean Absolute Error (MAE) across propellers with MAE $\leq 300$ (lower is better); red line marks 10,000 RPM.
}\label{fig:init_ablation}
\end{figure}

\subsection{Initial State Estimate Ablation}\label{sec:initial_state_ablation}

Trackers are provided with a coarse initial state (position, radius/scale, weak perspective, and RPM), supplied by the propeller detector~\cite{spetlik_efficient_2025}. We evaluate how much initialization quality matters by ablating each component independently and measuring the MAE of the estimated shaft RPM across 23 successfully tracked propellers (MAE $\leq$ 300 RPM) from \tquadcopterego{} (Sec.~\ref{sec:main_dataset}), marked with \ok{} in supplementary Table~\ref{tab:benchmark_complete}. We vary RPM and scale by $\pm\{1,2,\dots,100\}\%$ of their nominal values. For position, we translate along the image‑plane radial direction by a given fraction of the estimated blade radius and average ``toward‑distractor'' and ``away‑from‑distractor'' shifts. 

Fig.~\ref{fig:init_ablation} shows that the system is tolerant to errors in initialization but exhibits sharp, asymmetric failure modes for larger perturbations.

\subsection{Loss ablation}
A detailed ablation of \((\lambda_\phi, \lambda_r, \lambda_{\text{pol}}, \lambda_{\text{band}}, \lambda_{\text{BTO}})\) terms (\ref{eq:loss}) revealed that the \emph{phase} \(\lambda_\phi\) and \emph{radial} \(\lambda_r\) data terms provide the backbone for stable pose/phase coupling. The \emph{band} constraints \(\lambda_{\text{band}}\) and \emph{polarity} \(\lambda_{\text{pol}}\) alignment significantly boost robustness under distance and high \(M_{\text{ego}}\); and \emph{BTO} \(\lambda_{\text{BTO}}\) offers a small but reliable safety margin against occupancy imbalance. Together, these components are necessary to keep the asynchronous EKF and batched GN updates coherent. See the full results in supplementary Sec.~\ref{sec:loss_ablation_full}.

\subsection{Limitations}

\textbf{Initialization.} Coarse hub position/scale/RPM is required; small errors are tolerated, but larger perturbations ($> 30\%$) often cause failures (Sec.~\ref{sec:initial_state_ablation}). \\
\noindent\textbf{View and geometry.} A single planar homography and a helical phase on a visible annulus are assumed; edge‑on/side views (severe foreshortening), strong non‑planarity, or heavy occlusion violate this model and degrade tracking. \\
\noindent\textbf{Scene priors.} Known blade count and dominance of rotor events in the ring is assumed. However, the blade count may be estimated during initialization.\\
\noindent\textbf{Reinitialization.} \helixtrack{} does not perform automatic reinitialization after failure. 
Although feasible, doing so would obscure the evaluation by mixing tracking accuracy with recovery behavior.

\section{Conclusion}\label{sec:conclusion}

We presented \helixtrack{}, a geometry‑aware, event‑native tracker that turns the periodicity of a spinning propeller into a modeling advantage. By lifting events to rotor‑plane coordinates and enforcing a simple helical phase law, a per‑event EKF and batched Gauss-Newton pose updates jointly deliver trajectories and instantaneous RPM from raw events. 


Performance of \helixtrack{} on the proposed TQE benchmark exposed the difficulty of \emph{single‑propeller} tracking in the presence of near‑identical, co‑rotating distractors and egomotion. Representative asynchronous and learned trackers frequently fail or experience order‑of‑magnitude errors as motion and distance increase (Table~\ref{tab:benchmark}). 
At the same time, \helixtrack{} produces instantaneous angular‑speed and RPM estimates at event resolution, \ie, \emph{microsecond‑latency} updates, while sustaining $\,\approx 11.8\times\,$ real‑time throughput on TQE (Table~\ref{tab:speed}), pairing precise timing with practical speed. Pose updates are triggered adaptively whenever the accumulated helical phase exceeds $2\pi$. This defines an event-driven, self-timed pose-estimation latency typically on the order of tens of \emph{ms}.

%



{
    \small
    \bibliographystyle{ieeenat_fullname}
    \bibliography{references}

@inproceedings{ciattaglia_measuring_2024,
	title = {Measuring {UAV} {Propeller} {RPM} with {FMCW} {Radar}: {Validation} with {Calibrated} {Accelerometers}},
	doi = {10.1109/SAS60918.2024.10636506},
	booktitle = {{IEEE} {Sensors} {Applications} {Symposium} ({SAS})},
	author = {Ciattaglia, Gianluca and Iadarola, Grazia and Senigagliesi, Linda and Gambi, Ennio and Spinsante, Susanna},
	year = {2024},
	pages = {1--6},
}

@inproceedings{spetlik_efficient_2025,
	title = {Efficient {Real}-{Time} {Quadcopter} {Propeller} {Detection} and {Attribute} {Estimation} with {High}-{Resolution} {Event} {Camera}},
	isbn = {978-3-031-95911-0},
	doi = {10.1007/978-3-031-95911-0_16},
	booktitle = {Image {Analysis}},
	author = {Spetlik, Radim and Uhrová, Tereza and Matas, Jiří},
	editor = {Petersen, Jens and Dahl, Vedrana Andersen},
	year = {2025},
	pages = {217--230},
}

@article{wang_mambaevt_2025,
	title = {{MambaEVT}: {Event} {Stream} based {Visual} {Object} {Tracking} using {State} {Space} {Model}},
	issn = {1558-2205},
	doi = {10.1109/TCSVT.2025.3588533},
	journal = {IEEE Transactions on Circuits and Systems for Video Technology},
	author = {Wang, Xiao and Wang, Chao and Wang, Shiao and Wang, Xixi and Zhao, Zhicheng and Zhu, Lin and Jiang, Bo},
	year = {2025},
}

@inproceedings{li_3d_2024,
	title = {{3D} {Feature} {Tracking} via {Event} {Camera}},
	doi = {10.1109/CVPR52733.2024.01795},
	booktitle = {{IEEE}/{CVF} {Conference} on {Computer} {Vision} and {Pattern} {Recognition}},
	author = {Li, Siqi and Zhou, Zhikuan and Xue, Zhou and Li, Yipeng and Du, Shaoyi and Gao, Yue},
	year = {2024},
	pages = {18974--18983},
}

@inproceedings{wang_event_2024,
	title = {Event {Stream}-{Based} {Visual} {Object} {Tracking}: {A} {High}-{Resolution} {Benchmark} {Dataset} and {A} {Novel} {Baseline}},
	doi = {10.1109/CVPR52733.2024.01821},
	booktitle = {{IEEE}/{CVF} {Conference} on {Computer} {Vision} and {Pattern} {Recognition}},
	author = {Wang, Xiao and Wang, Shiao and Tang, Chuanming and Zhu, Lin and Jiang, Bo and Tian, Yonghong and Tang, Jin},
	year = {2024},
	pages = {19248--19257},
}

@article{wang_spikemot_2025,
	title = {{SpikeMOT}: {Event}-{Based} {Multi}-{Object} {Tracking} {With} {Sparse} {Motion} {Features}},
	volume = {13},
	issn = {2169-3536},
	doi = {10.1109/ACCESS.2024.3523411},
	journal = {IEEE Access},
	author = {Wang, Song and Wang, Zhu and Li, Can and Qi, Xiaojuan and So, Hayden Kwok-Hay},
	year = {2025},
	pages = {214--230},
}

@article{sun_reliable_2024,
	title = {Reliable object tracking by multimodal hybrid feature extraction and transformer-based fusion},
	volume = {178},
	issn = {0893-6080},
	doi = {10.1016/j.neunet.2024.106493},
	journal = {Neural Networks},
	author = {Sun, Hongze and Liu, Rui and Cai, Wuque and Wang, Jun and Wang, Yue and Tang, Huajin and Cui, Yan and Yao, Dezhong and Guo, Daqing},
	year = {2024},
	pages = {106493},
}

@misc{zafeiri_event-ecc_2024,
	title = {Event-{ECC}: {Asynchronous} {Tracking} of {Events} with {Continuous} {Optimization}},
	doi = {10.48550/arXiv.2409.14564},
	author = {Zafeiri, Maria and Evangelidis, Georgios and Psarakis, Emmanouil},
	year = {2024},
}

@article{wang_asynchronous_2024,
	title = {Asynchronous {Blob} {Tracker} for {Event} {Cameras}},
	volume = {40},
	issn = {1941-0468},
	doi = {10.1109/TRO.2024.3454410},
	journal = {IEEE Transactions on Robotics},
	author = {Wang, Ziwei and Molloy, Timothy and van Goor, Pieter and Mahony, Robert},
	year = {2024},
	pages = {4750--4767},
}

@article{messikommer_data-driven_2025,
	title = {Data-{Driven} {Feature} {Tracking} for {Event} {Cameras} {With} and {Without} {Frames}},
	volume = {47},
	issn = {1939-3539},
	doi = {10.1109/TPAMI.2025.3536016},
	number = {5},
	journal = {IEEE Transactions on Pattern Analysis and Machine Intelligence},
	author = {Messikommer, Nico and Fang, Carter and Gehrig, Mathias and Cioffi, Giovanni and Scaramuzza, Davide},
	year = {2025},
	pages = {3706--3717},
}

@inproceedings{kolar_eeppr_2025,
	title = {{EEPPR}: event-based estimation of periodic phenomena rate using correlation in {3D}},
	volume = {13517},
	doi = {10.1117/12.3055033},
	booktitle = {Seventeenth {International} {Conference} on {Machine} {Vision} ({ICMV} 2024)},
	publisher = {SPIE},
	author = {Kolář, Jakub and Špetlík, Radim and Matas, Jiří},
	year = {2025},
	pages = {223--230},
}

@misc{zheng_deep_2024,
	title = {Deep {Learning} for {Event}-based {Vision}: {A} {Comprehensive} {Survey} and {Benchmarks}},
	url = {http://arxiv.org/abs/2302.08890},
	author = {Zheng, Xu and Liu, Yexin and Lu, Yunfan and Hua, Tongyan and Pan, Tianbo and Zhang, Weiming and Tao, Dacheng and Wang, Lin},
	year = {2024},
}

@inproceedings{sanket_evpropnet_2021,
	title = {{EVPropNet}: {Detecting} {Drones} {By} {Finding} {Propellers} {For} {Mid}-{Air} {Landing} {And} {Following}},
	booktitle = {Robotics: {Science} and {Systems} 2021},
	author = {Sanket, Nitin J. and Chahat, Deep Singh and Parameshwara, Chethan M. and Fermüller, Cornelia and de Croon, Guido C.H.E. and Aloimonos, Yiannis},
	year = {2021},
}

@inproceedings{stewart_drone_2021,
	title = {Drone {Virtual} {Fence} {Using} a {Neuromorphic} {Camera}},
	isbn = {978-1-4503-8691-3},
	doi = {10.1145/3477145.3477264},
	booktitle = {International {Conference} on {Neuromorphic} {Systems} 2021},
	author = {Stewart, Terrence and Drouin, Marc-Antoine and Gagne, Guillaume and Godin, Guy},
	year = {2021},
	pages = {1--9},
}

@inproceedings{stewart_virtual_2022,
	title = {A {Virtual} {Fence} for {Drones}: {Efficiently} {Detecting} {Propeller} {Blades} with a {DVXplorer} {Event} {Camera}},
	isbn = {978-1-4503-9789-6},
	doi = {10.1145/3546790.3546800},
	booktitle = {Proceedings of the {International} {Conference} on {Neuromorphic} {Systems} 2022},
	author = {Stewart, Terrence and Drouin, Marc-Antoine and Picard, Michel and Djupkep Dizeu, Frank Billy and Orth, Anthony and Gagné, Guillaume},
	year = {2022},
	pages = {1--7},
}

@article{vrba_marker-less_2020,
	title = {Marker-{Less} {Micro} {Aerial} {Vehicle} {Detection} and {Localization} {Using} {Convolutional} {Neural} {Networks}},
	volume = {5},
	issn = {2377-3766},
	doi = {10.1109/LRA.2020.2972819},
	number = {2},
	journal = {IEEE Robotics and Automation Letters},
	author = {Vrba, Matouš and Saska, Martin},
	year = {2020},
	pages = {2459--2466},
}

@article{zheng_keypoint-guided_2024,
	title = {Keypoint-{Guided} {Efficient} {Pose} {Estimation} and {Domain} {Adaptation} for {Micro} {Aerial} {Vehicles}},
	volume = {40},
	issn = {1941-0468},
	doi = {10.1109/TRO.2024.3400938},
	journal = {IEEE Transactions on Robotics},
	author = {Zheng, Ye and Zheng, Canlun and Shen, Jiahao and Liu, Peidong and Zhao, Shiyu},
	year = {2024},
	pages = {2967--2983},
}

@inproceedings{mitrokhin_event-based_2018,
	title = {Event-{Based} {Moving} {Object} {Detection} and {Tracking}},
	doi = {10.1109/IROS.2018.8593805},
	booktitle = {2018 {IEEE}/{RSJ} {International} {Conference} on {Intelligent} {Robots} and {Systems} ({IROS})},
	author = {Mitrokhin, Anton and Fermüller, Cornelia and Parameshwara, Chethan and Aloimonos, Yiannis},
	year = {2018},
	pages = {1--9},
}
}
\clearpage
\setcounter{page}{1}
\maketitlesupplementary

\newcommand{\beginsupplement}{ %
  \setcounter{figure}{0}       %
  \setcounter{table}{0}        %
  \setcounter{equation}{0}     %
  \setcounter{section}{0}      %
  %
  \renewcommand{\thefigure}{{S}\arabic{figure}}%
  \renewcommand{\thetable}{{S}\arabic{table}}%
  \renewcommand{\theequation}{{S}\arabic{equation}}%
  \renewcommand{\thesection}{{S}\arabic{section}}%
  %
  \crefname{figure}{Fig.\,S}{Figs.\,S}%
  \crefname{table}{Table\,S}{Tables\,S}%
  \crefname{equation}{Eq.\,S}{Eqs.\,S}%
}
\beginsupplement

\section{Full Gauss-Newton System with Jacobians}\label{sec:full_gn}

In this section, we describe the full Gauss-Newton System with Jacobians of the terms mentioned in Sec.~\ref{sec:method}.

\paragraph{Jacobians for all residuals.}
Let $\sigma(x)=\frac{1}{1+\mathrm{e}^{-x}}$ and adopt $\tfrac{d}{dx}\operatorname{wrap}_\pi(x)=1$ a.e.
For polarity,
\[
\delta_i=\operatorname{wrap}_\pi(\varepsilon_i^\phi-d_{\text{pol}}),\quad
\varepsilon_i^{\text{pol}} = c_{\text{pol}}\sin(\tfrac{1}{2}\delta_i),
\]
\[
\mathbf{J}_i^{\text{pol}} = c_{\text{pol}}\cdot\tfrac{1}{2}\cos(\tfrac{1}{2}\delta_i)\,\mathbf{J}_i^{\phi}.
\]
For soft band barriers,
\[
\varepsilon_i^{\text{in}}=c_b\,\operatorname{softplus}_\tau(r_{\text{in}}-r_i),\quad
\varepsilon_i^{\text{out}}=c_b\,\operatorname{softplus}_\tau(r_i-r_{\text{out}}),
\]
\[
\mathbf{J}_i^{\text{in}}=-c_b\,\sigma\!\left(\frac{r_{\text{in}}-r_i}{\tau}\right)\mathbf{J}_i^{r},\quad
\mathbf{J}_i^{\text{out}}=\;\,c_b\,\sigma\!\left(\frac{r_i-r_{\text{out}}}{\tau}\right)\mathbf{J}_i^{r}.
\]

\paragraph{Balanced Tip Occupancy (BTO).}
Define soft occupancies with temperature $\tau_{\text{occ}}$,
\[
m_i^{\text{in}}=\sigma\!\Big(\frac{r_{\text{in}}-r_i}{\tau_{\text{occ}}}\Big), \quad
m_i^{\text{out}}=\sigma\!\Big(\frac{r_i-r_{\text{out}}}{\tau_{\text{occ}}}\Big),
\]
\[
p_{\text{in}}=\tfrac{1}{|\mathcal{B}|}\sum_i m_i^{\text{in}},\quad
p_{\text{out}}=\tfrac{1}{|\mathcal{B}|}\sum_i m_i^{\text{out}}.
\]
Then
\[
\frac{\partial p_{\text{in}}}{\partial \mathbf{q}}=\tfrac{1}{|\mathcal{B}|}\sum_i\Big[-\tfrac{1}{\tau_{\text{occ}}}m_i^{\text{in}}(1-m_i^{\text{in}})\mathbf{J}_i^{r}\Big],
\]
\[
\frac{\partial p_{\text{out}}}{\partial \mathbf{q}}=\tfrac{1}{|\mathcal{B}|}\sum_i\Big[\ \tfrac{1}{\tau_{\text{occ}}}m_i^{\text{out}}(1-m_i^{\text{out}})\mathbf{J}_i^{r}\Big].
\]
Let 
\[\phi(s)=\operatorname{softplus}_\tau(s), \quad \phi'(s)=\sigma(s/\tau),\]
\[\phi''(s)=\sigma(s/\tau)(1-\sigma(s/\tau))/\tau\]
and
\[
s_{\text{in}}^{\text{lo}}=\underline{p_{\text{in}}}-p_{\text{in}},\quad
s_{\text{in}}^{\text{hi}}=p_{\text{in}}-\overline{p_{\text{in}}},
\]
\[
s_{\text{out}}^{\text{lo}}=\underline{p_{\text{out}}}-p_{\text{out}},\quad
s_{\text{out}}^{\text{hi}}=p_{\text{out}}-\overline{p_{\text{out}}}.
\]
Hence
\[
\mathbf{J}_{s_{\text{in}}^{\text{lo}}}= -\,\tfrac{\partial p_{\text{in}}}{\partial \mathbf{q}},\;
\mathbf{J}_{s_{\text{in}}^{\text{hi}}}= \ \tfrac{\partial p_{\text{in}}}{\partial \mathbf{q}},\;
\mathbf{J}_{s_{\text{out}}^{\text{lo}}}= -\,\tfrac{\partial p_{\text{out}}}{\partial \mathbf{q}},\;
\mathbf{J}_{s_{\text{out}}^{\text{hi}}}= \ \tfrac{\partial p_{\text{out}}}{\partial \mathbf{q}}.
\]

\paragraph{Complete GN system.}
With frozen gates $g_i^\phi=\lambda_\phi w_i^\phi$ and $g_i^r=\lambda_r w_i^r$,
\begin{align}
\mathbf{H}&_{\text{GN}}=
\sum_i g_i^\phi(\mathbf{J}_i^{\phi})^\top \mathbf{J}_i^{\phi}
+\sum_i g_i^r(\mathbf{J}_i^{r})^\top \mathbf{J}_i^{r}\\
&+\lambda_{\text{pol}}\sum_i(\mathbf{J}_i^{\text{pol}})^\top \mathbf{J}_i^{\text{pol}}\\
&+\lambda_{\text{band}}\sum_i\Big[(\mathbf{J}_i^{\text{in}})^\top \mathbf{J}_i^{\text{in}}+(\mathbf{J}_i^{\text{out}})^\top \mathbf{J}_i^{\text{out}}\Big]\\
&+\lambda_{\text{BTO}}\!\!\sum_{q\in\{\text{in,out}\}}\!\Big[
\phi''(s_q^{\text{lo}})\,(\mathbf{J}_{s_q^{\text{lo}}})^\top\mathbf{J}_{s_q^{\text{lo}}}
\\&\qquad\qquad+\phi''(s_q^{\text{hi}})\,(\mathbf{J}_{s_q^{\text{hi}}})^\top\mathbf{J}_{s_q^{\text{hi}}}
\Big]\\
&+\mathbf{H}_{\text{reg}},
\end{align}
\begin{align}
\mathbf{g}&=
\sum_i g_i^\phi\,\varepsilon_i^\phi(\mathbf{J}_i^{\phi})^\top+\sum_i g_i^r\,\varepsilon_i^r(\mathbf{J}_i^{r})^\top\\
&+\lambda_{\text{pol}}\sum_i \varepsilon_i^{\text{pol}}(\mathbf{J}_i^{\text{pol}})^\top\\
&+\lambda_{\text{band}}\sum_i\Big[\varepsilon_i^{\text{in}}(\mathbf{J}_i^{\text{in}})^\top+\varepsilon_i^{\text{out}}(\mathbf{J}_i^{\text{out}})^\top\Big]\\
&+\lambda_{\text{BTO}}\!\!\sum_{q\in\{\text{in,out}\}}\!\Big[
\phi'(s_q^{\text{lo}})\,(\mathbf{J}_{s_q^{\text{lo}}})^\top
+\phi'(s_q^{\text{hi}})\,(\mathbf{J}_{s_q^{\text{hi}}})^\top
\Big]\\
&+\mathbf{g}_{\text{reg}}.
\end{align}

\section{Dataset Details}
\label{sec:dataset}

This section gives a detailed overview of \tquadcopterego{} dataset. 

The average number of ground-truth timetamps is 4\,770, maximum 5\,116, minimum 4\,906, and median is 4\,876.

The average number of processed events is $1.7 \times10^9$, maximum $3.1 \times 10^9$, minimum $1.0 \times 10^9$, and median is $1.6\times10^9$.

The average event rate is $60,4 \times 10^6$ (events / second).

\begin{figure*}[ht]
\renewcommand{\arraystretch}{0}\centering
\begin{tabular}{|@{}c@{}|@{}c@{}|@{}c@{}|@{}c@{}|@{}c@{}|@{}c@{}|}
\hline
 \rotatebox{90}{\parbox{5em}{\centering $M_\text{ego}=1$}} &\includegraphics[width=0.195\linewidth]{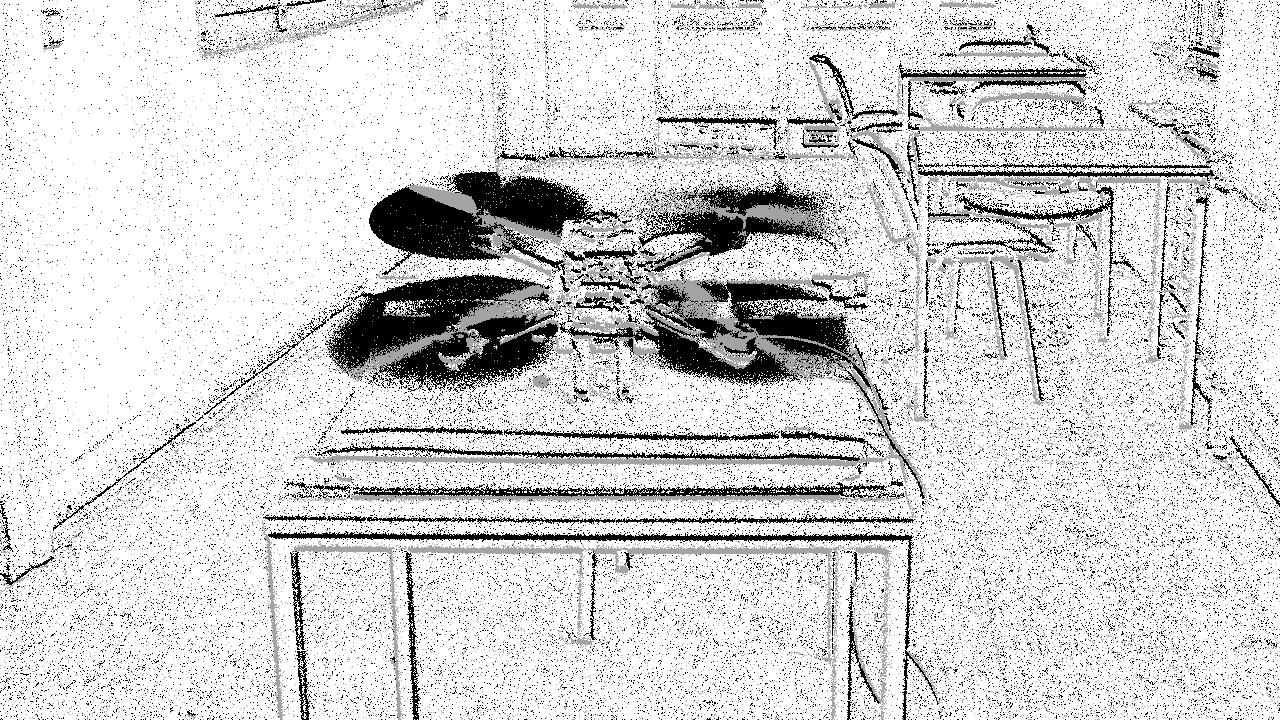} & \includegraphics[width=0.195\linewidth]{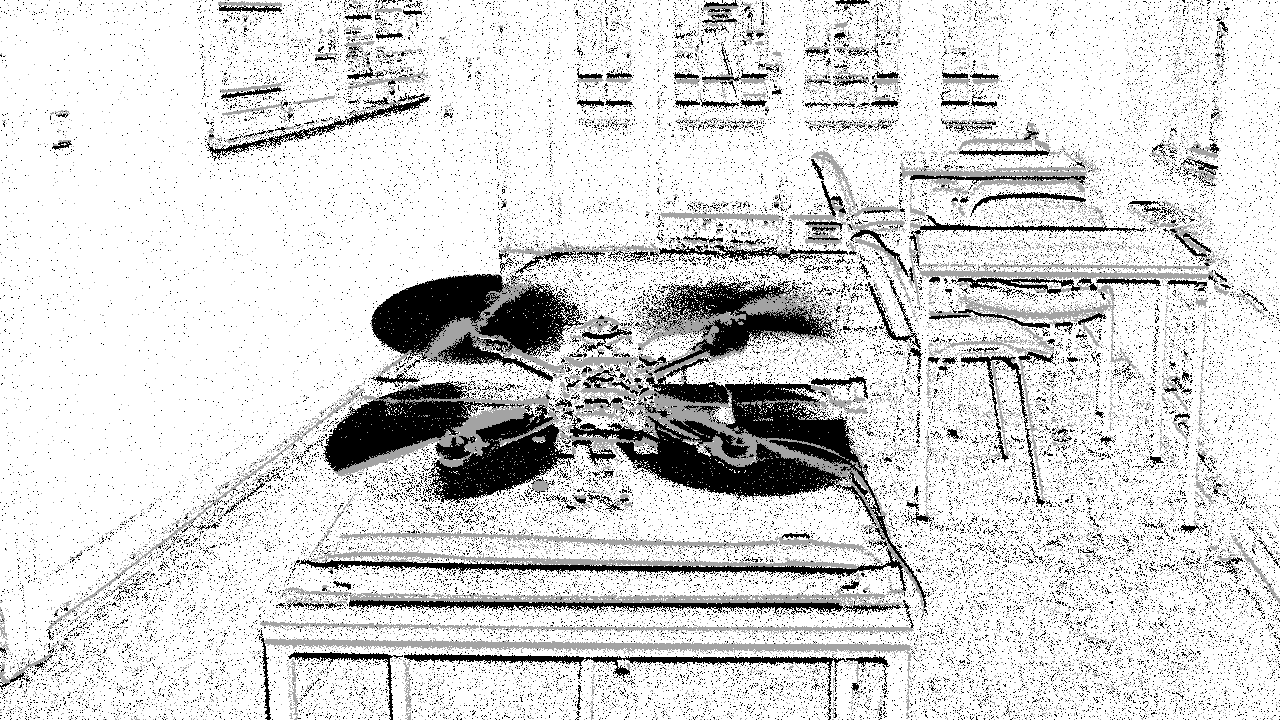} & \includegraphics[width=0.195\linewidth]{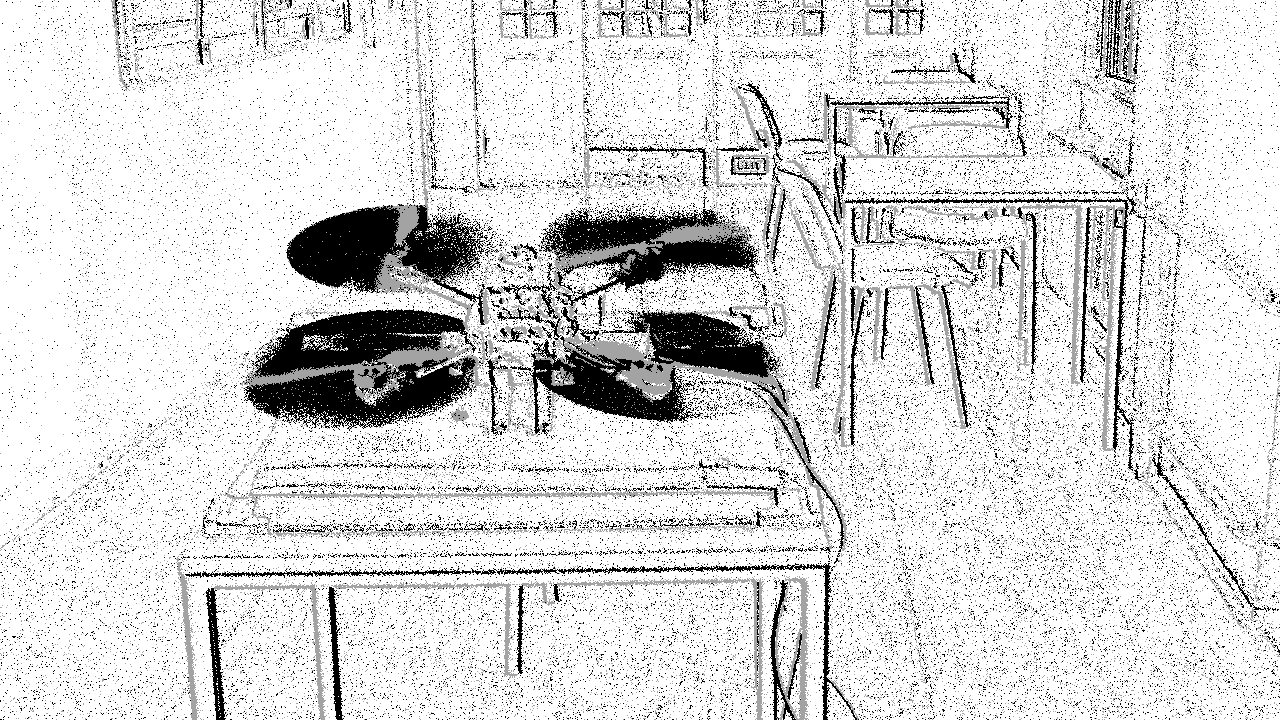} & \includegraphics[width=0.195\linewidth]{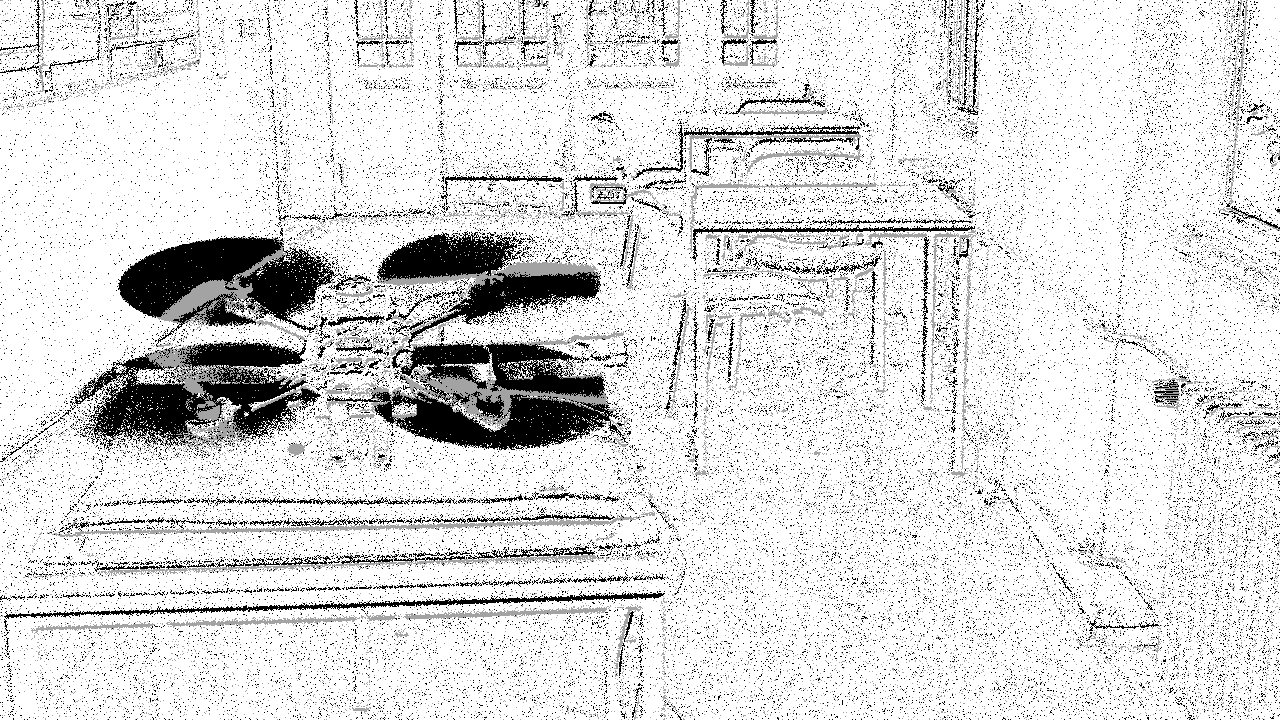} & \includegraphics[width=0.195\linewidth]{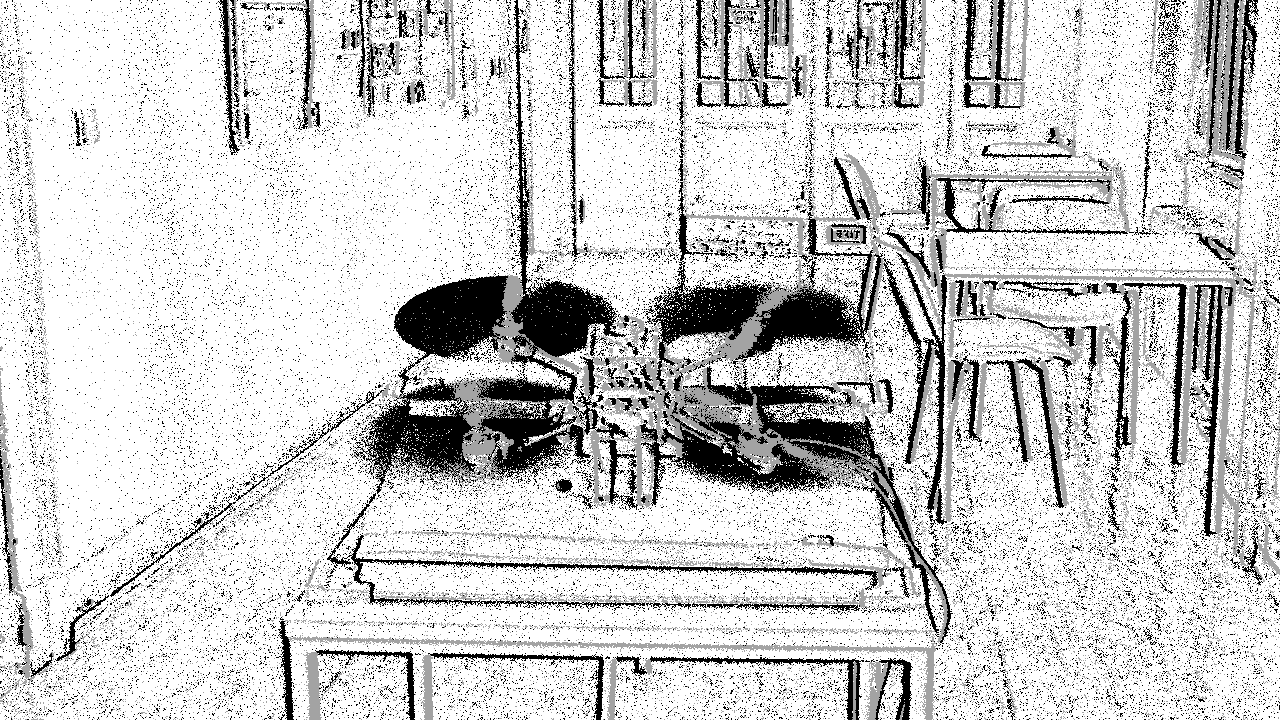}  \\
\hline
 \rotatebox{90}{\parbox{5em}{\centering $M_\text{ego}=2$}} &\includegraphics[width=0.195\linewidth]{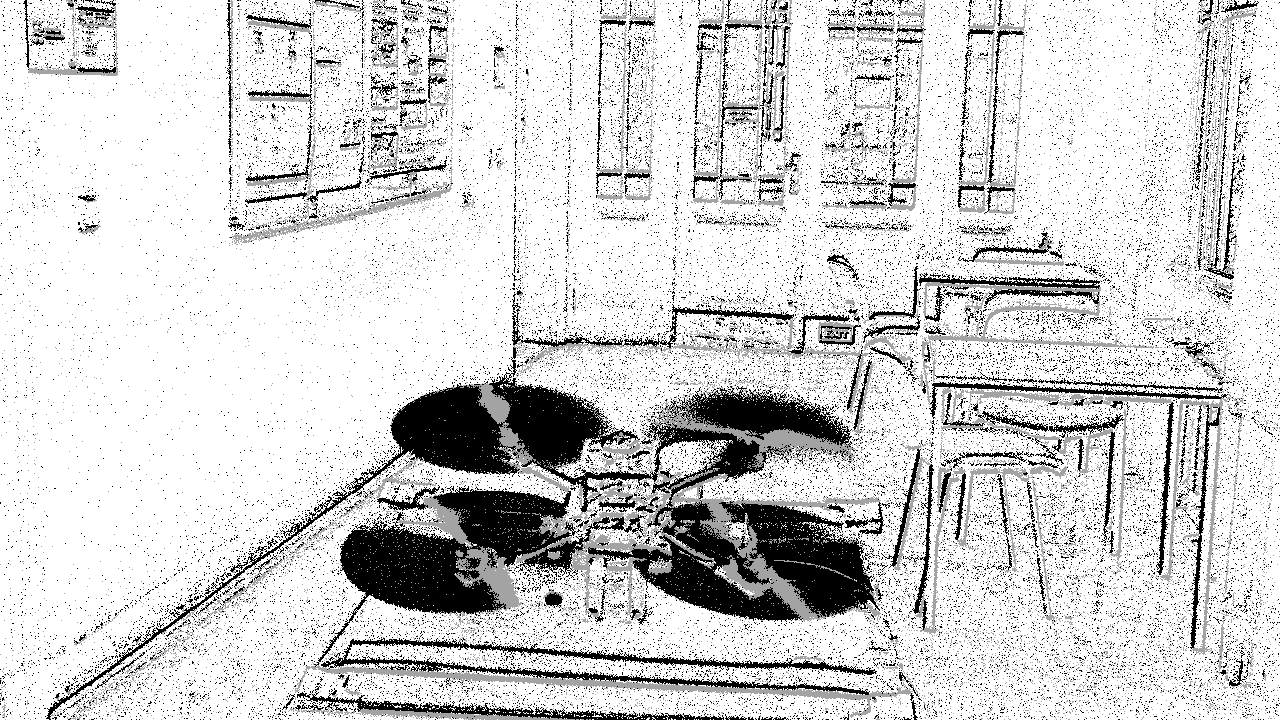} & \includegraphics[width=0.195\linewidth]{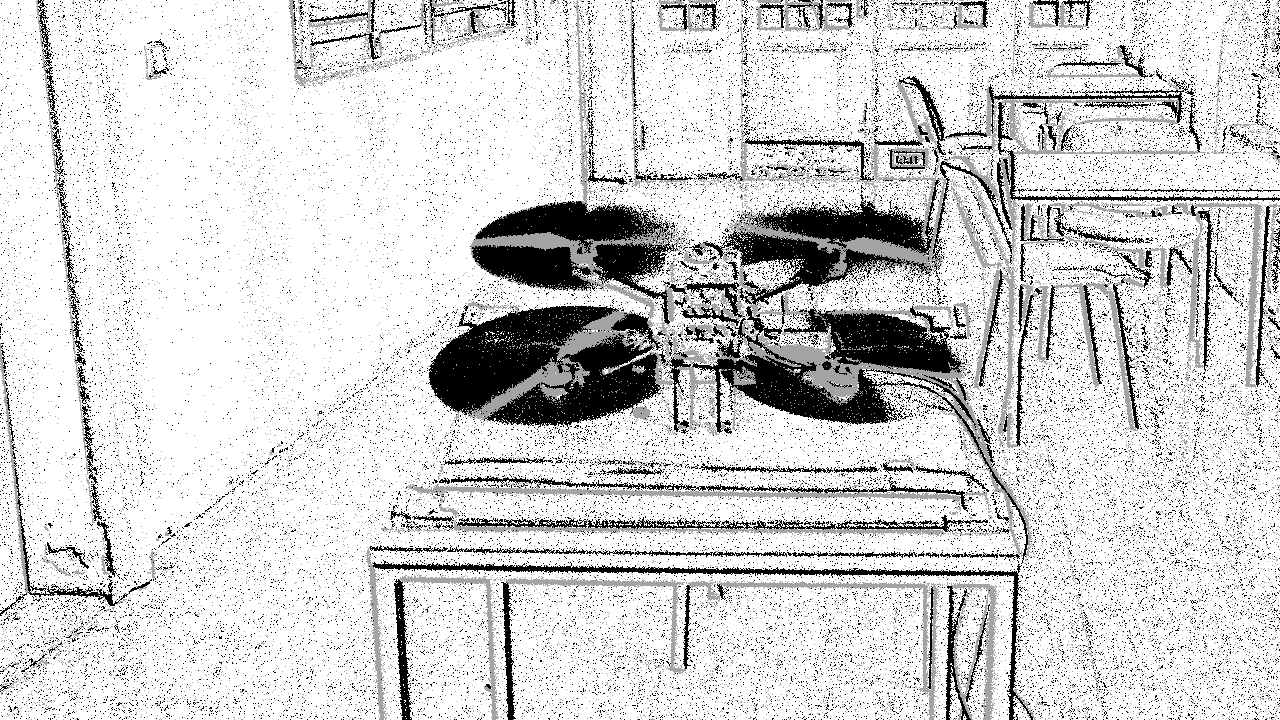} & \includegraphics[width=0.195\linewidth]{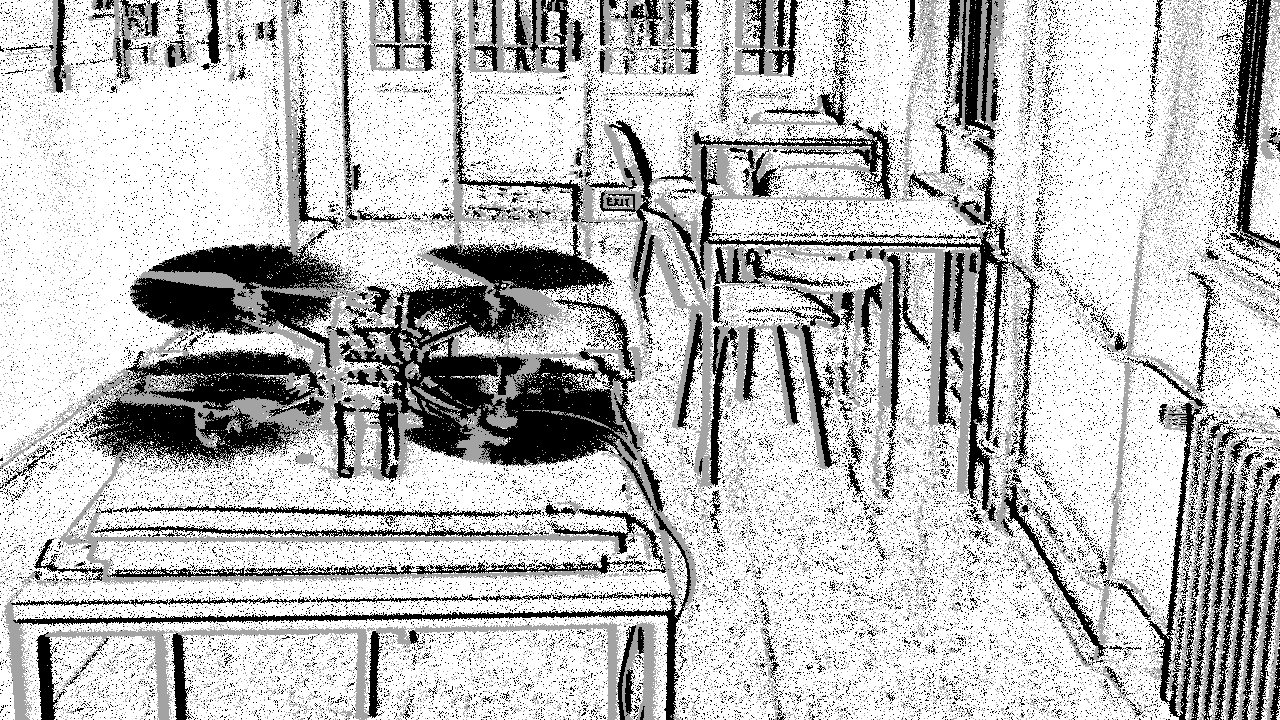} & \includegraphics[width=0.195\linewidth]{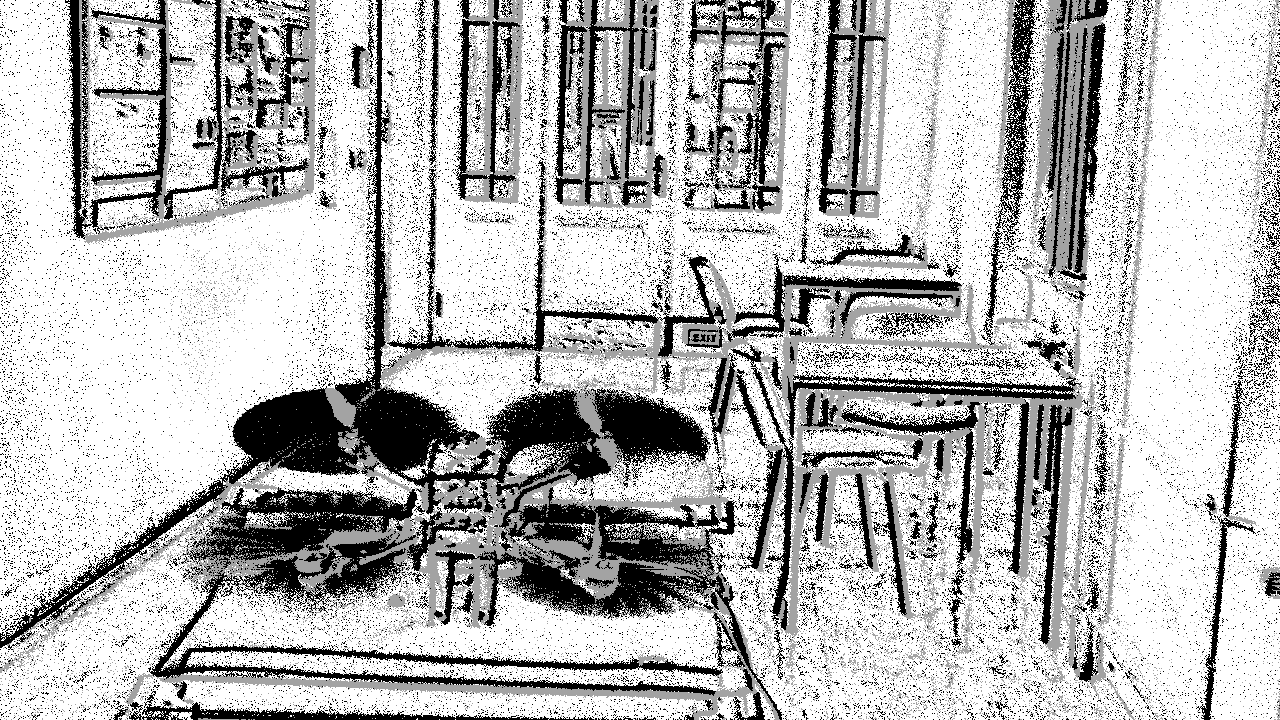} & \includegraphics[width=0.195\linewidth]{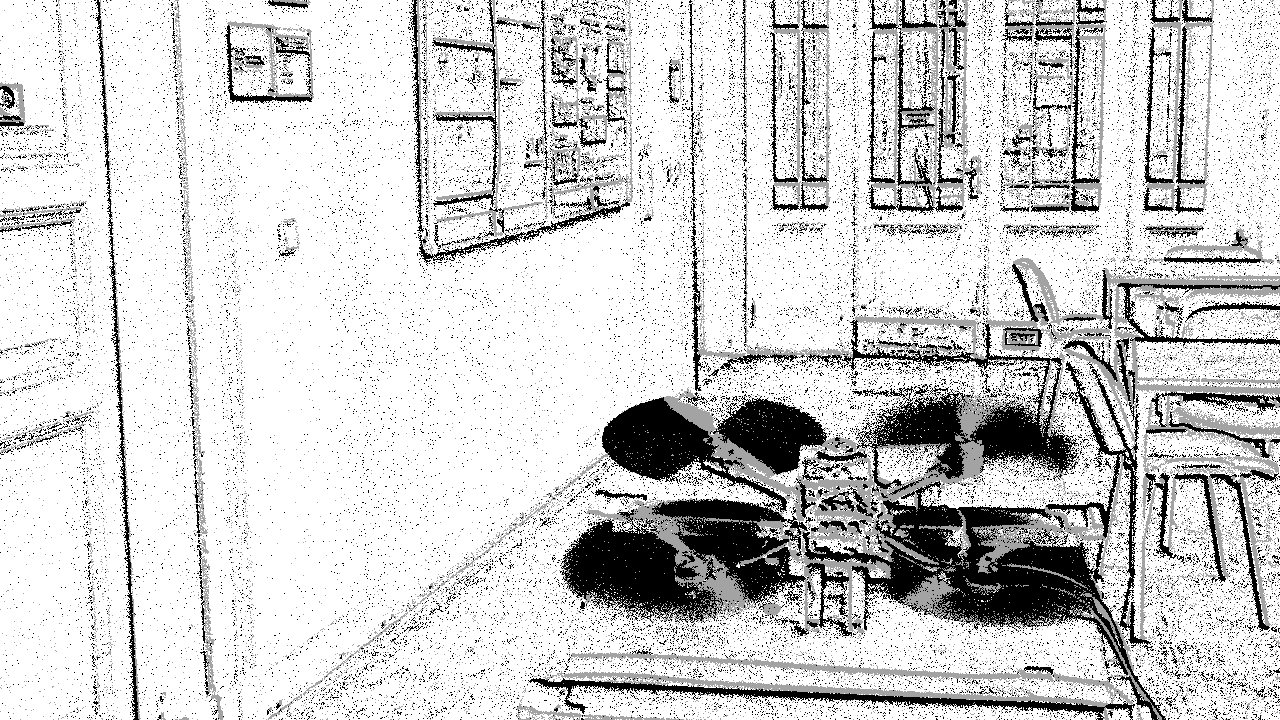}  \\
\hline
 \rotatebox{90}{\parbox{5em}{\centering $M_\text{ego}=3$}} &\includegraphics[width=0.195\linewidth]{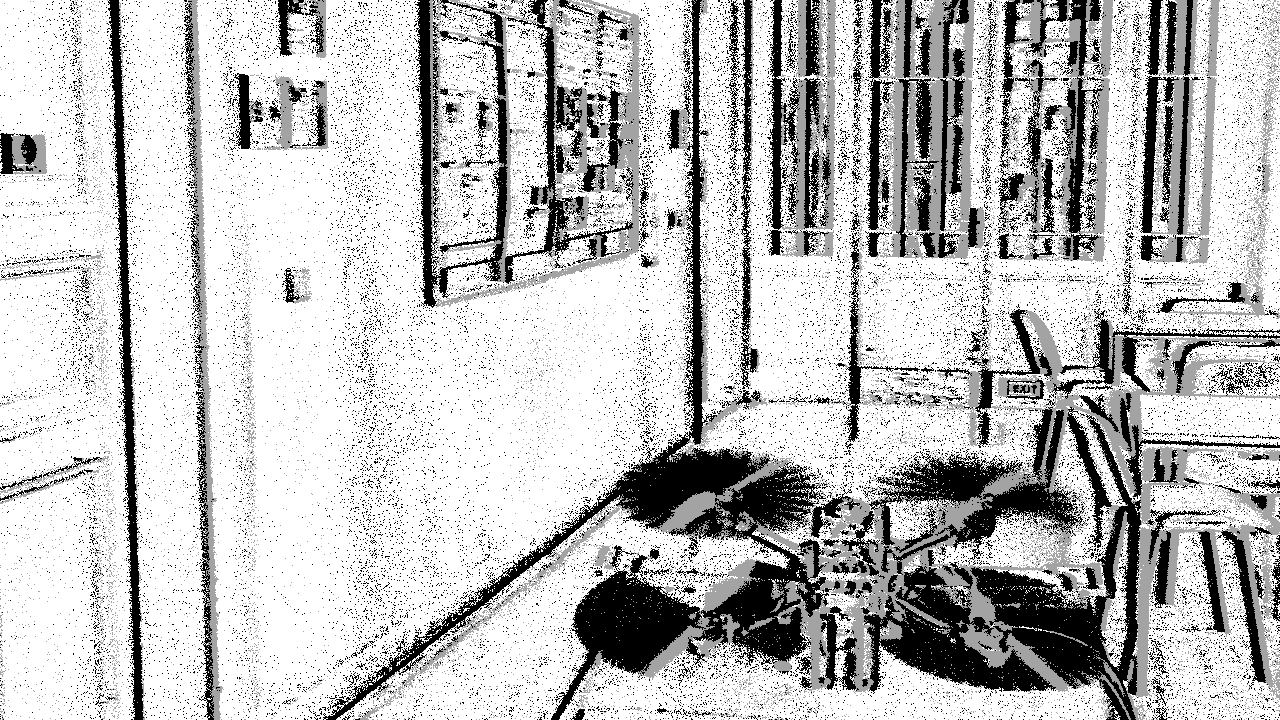} & \includegraphics[width=0.195\linewidth]{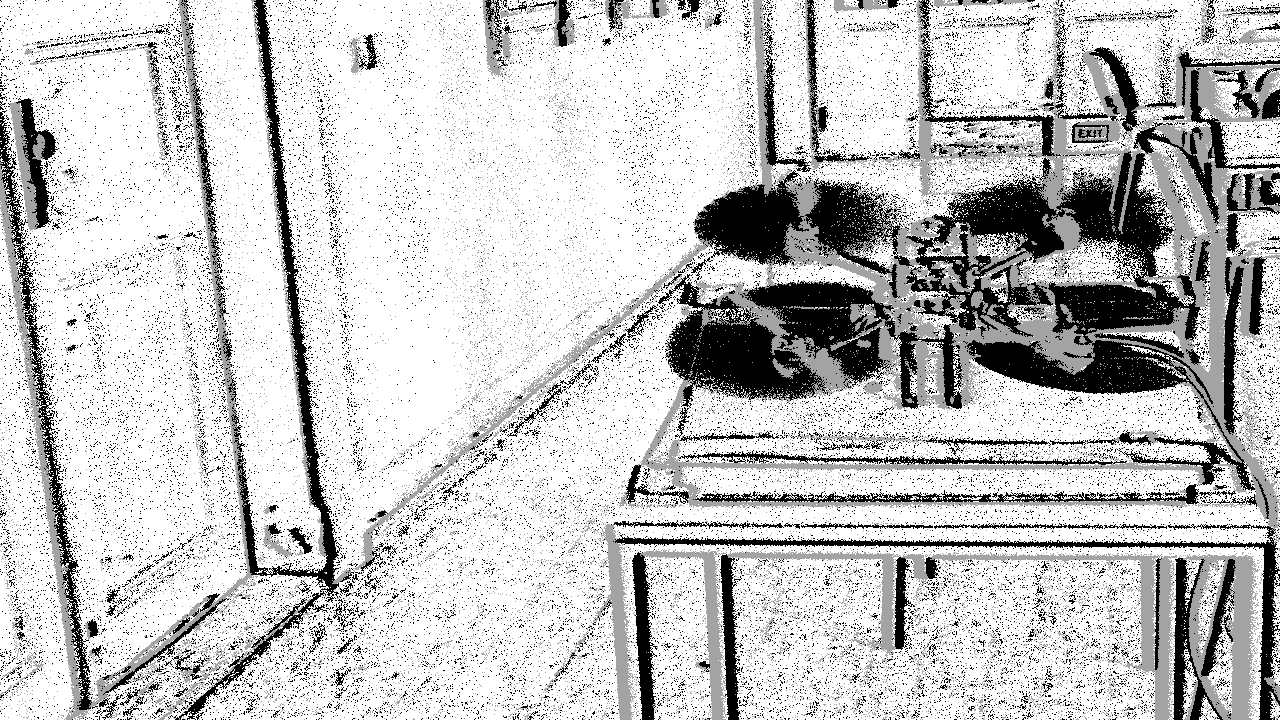} & \includegraphics[width=0.195\linewidth]{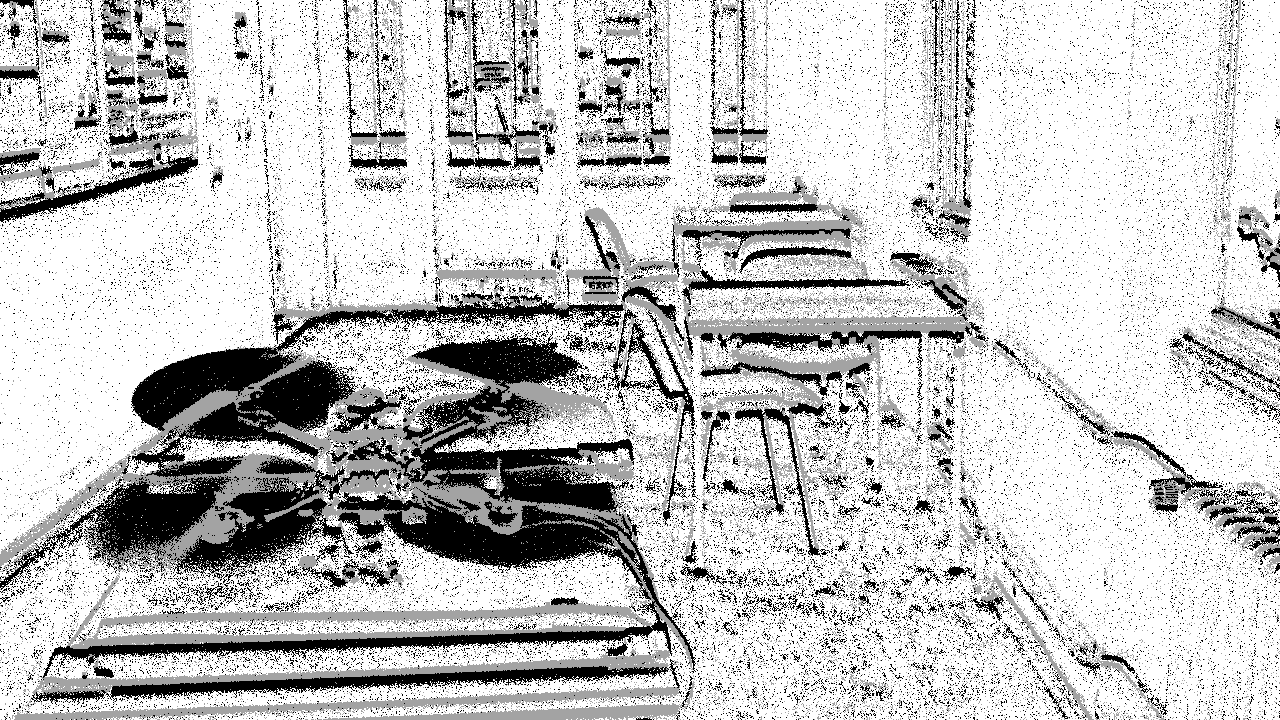} & \includegraphics[width=0.195\linewidth]{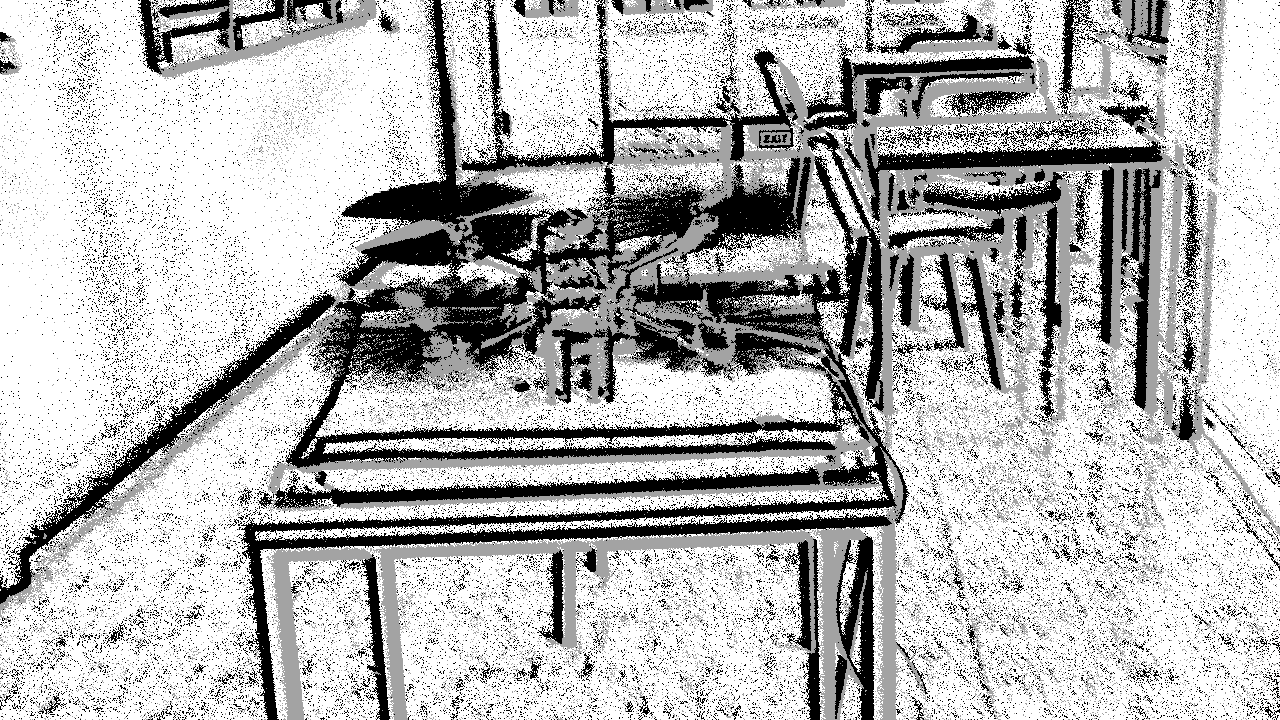} & \includegraphics[width=0.195\linewidth]{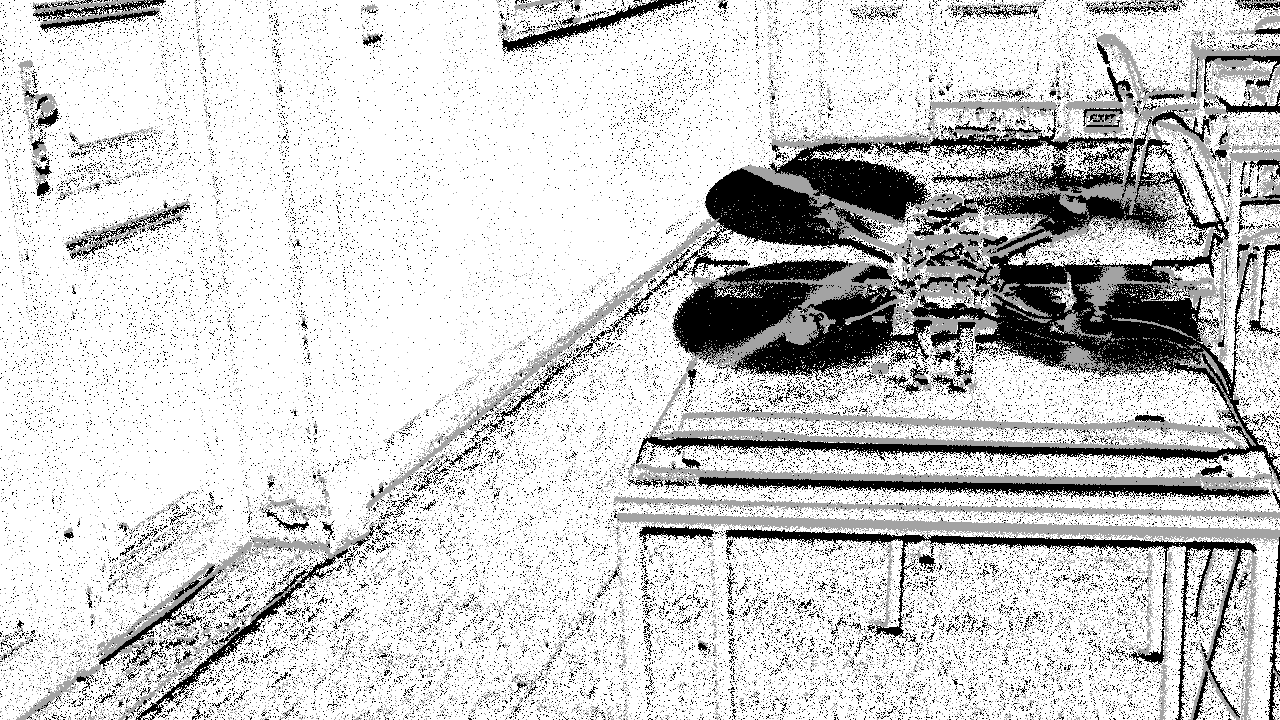}  \\
\hline
 \rotatebox{90}{\parbox{5em}{\centering $M_\text{ego}=4$}} &\includegraphics[width=0.195\linewidth]{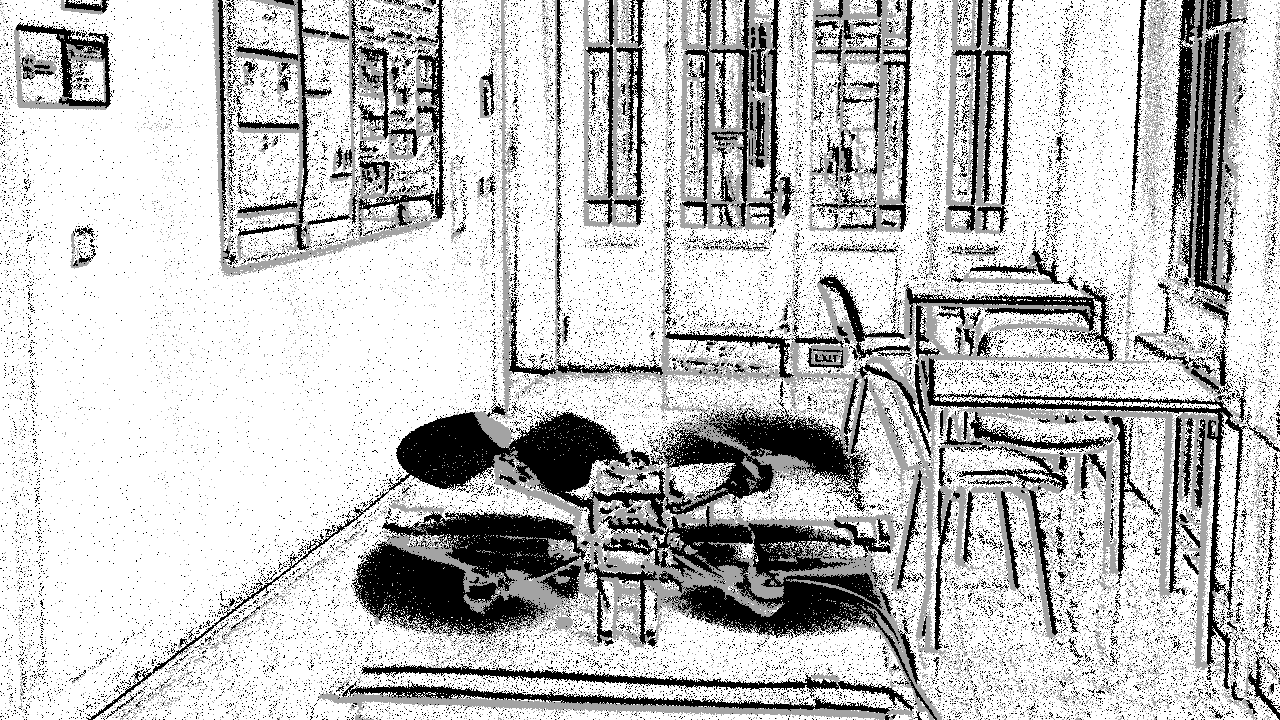} & \includegraphics[width=0.195\linewidth]{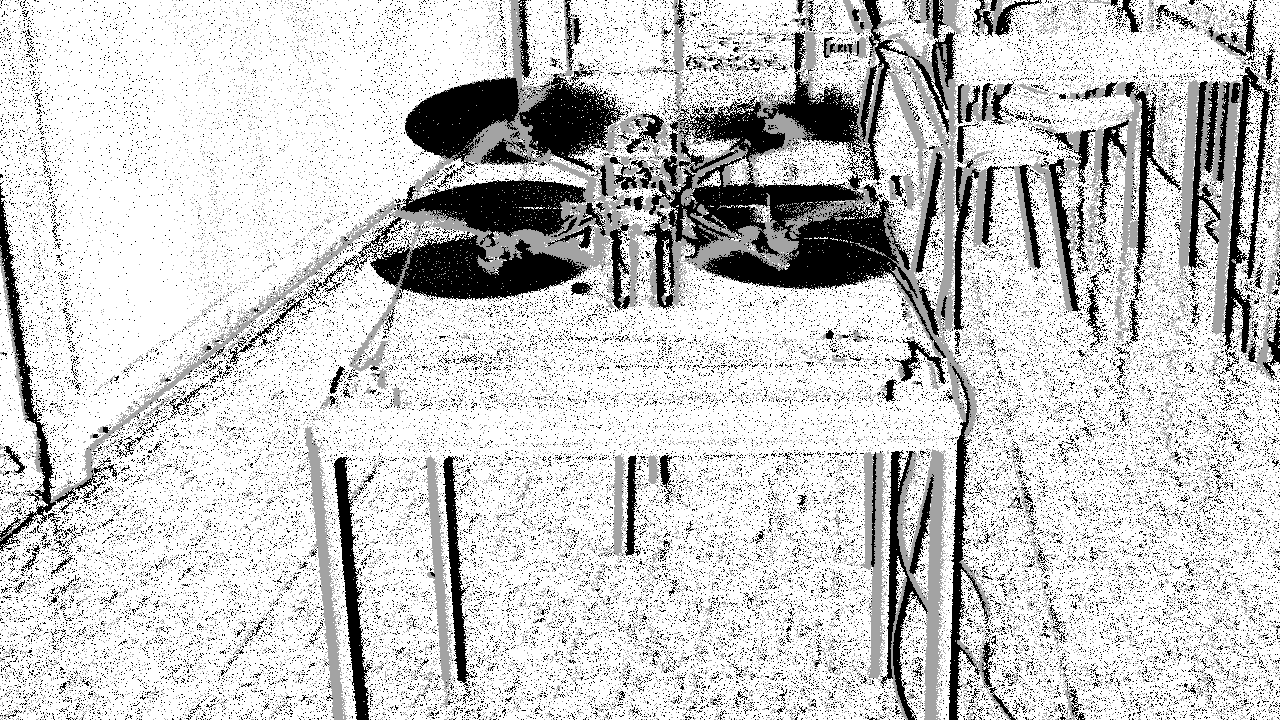} & \includegraphics[width=0.195\linewidth]{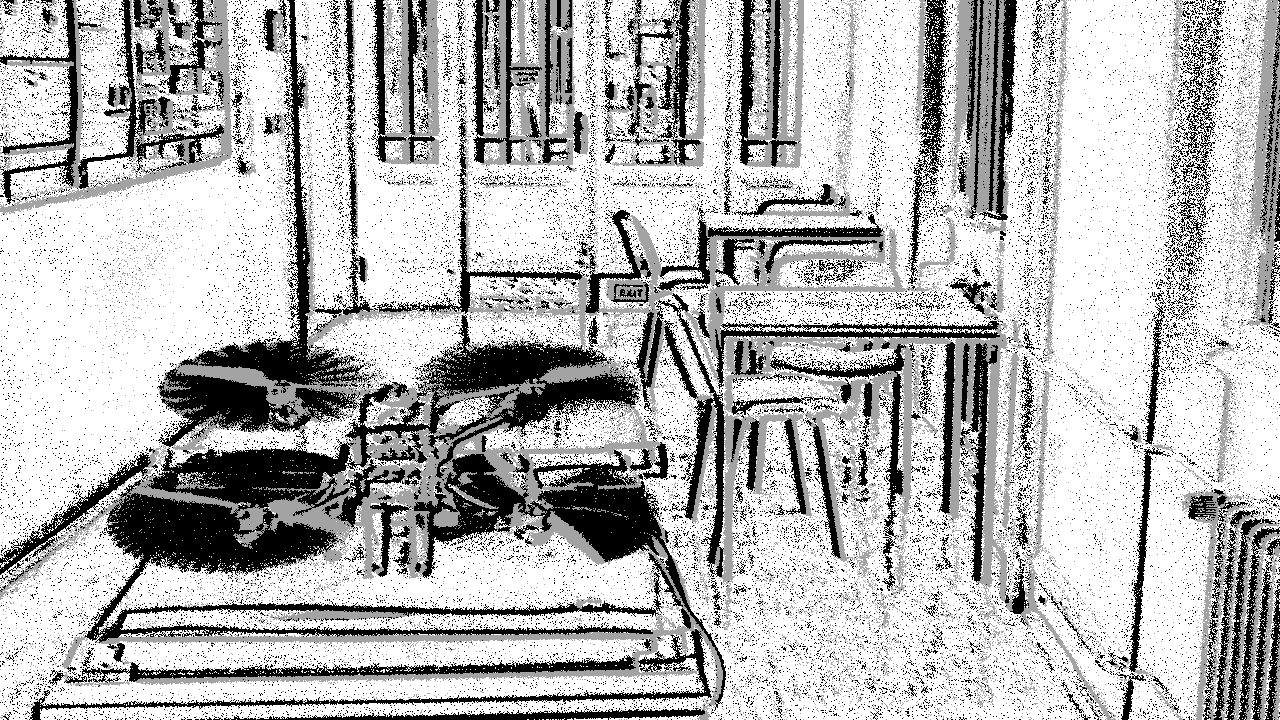} & \includegraphics[width=0.195\linewidth]{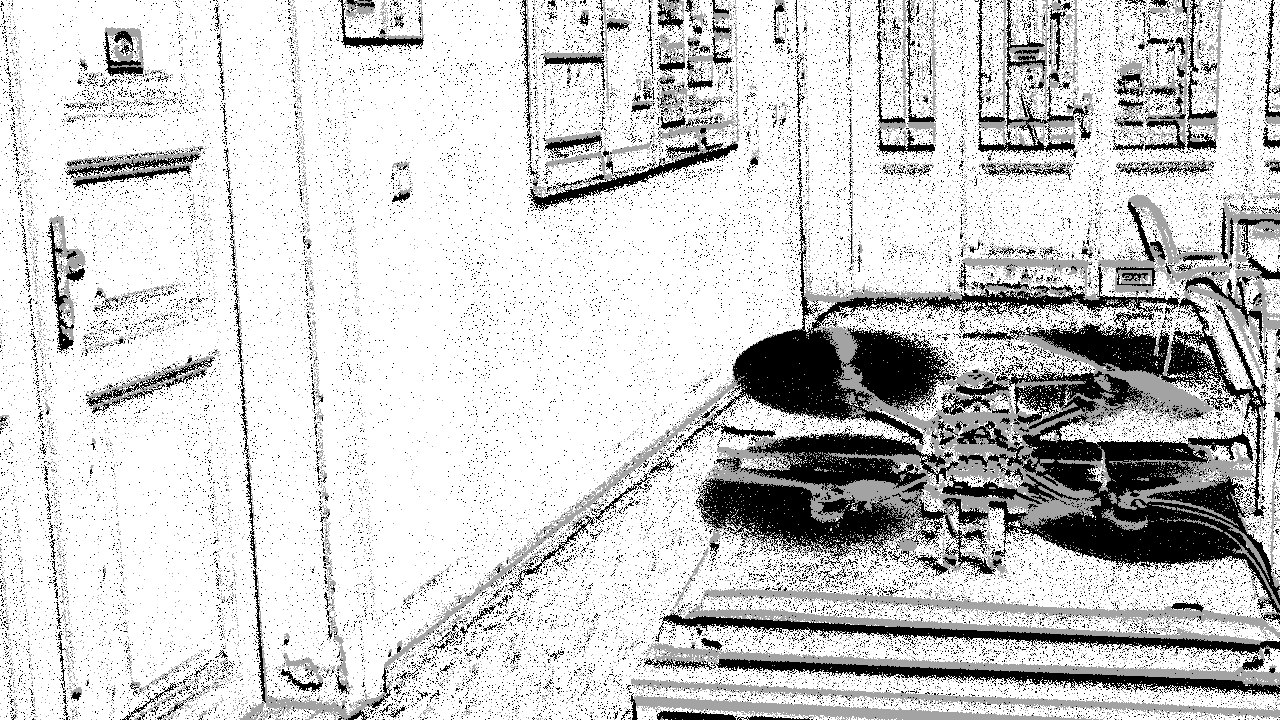} & \includegraphics[width=0.195\linewidth]{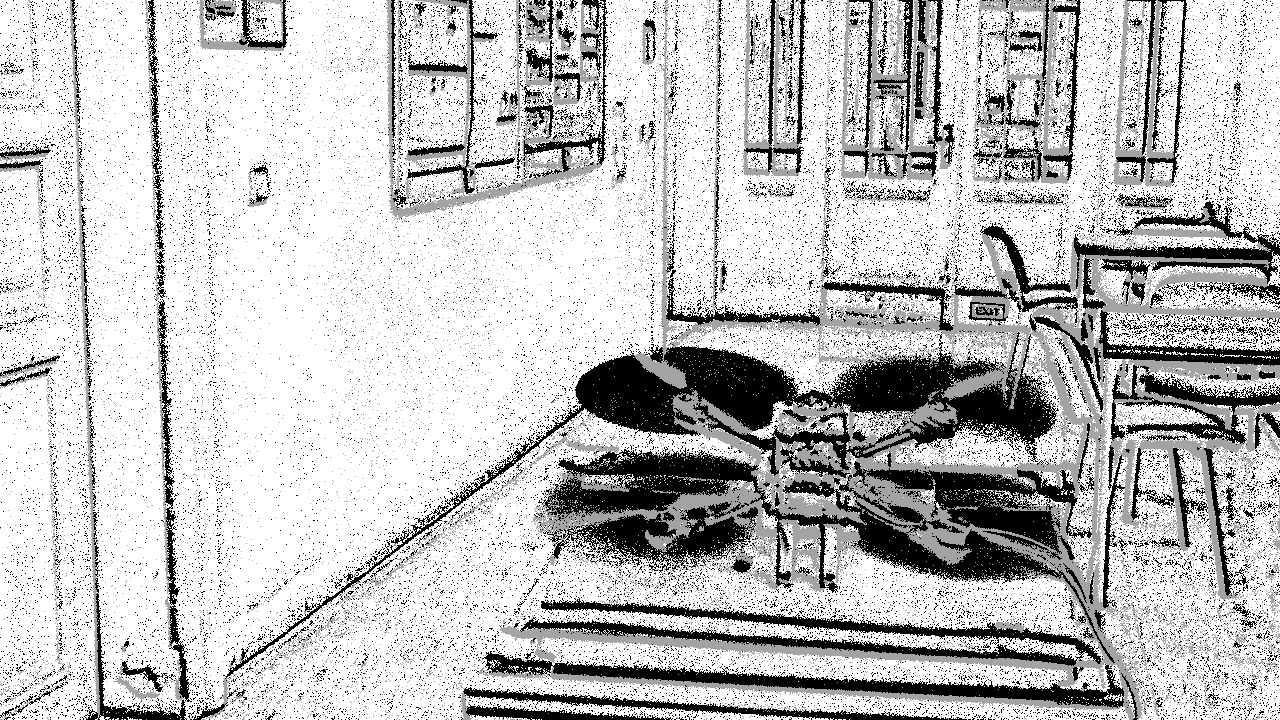}  \\
\hline
 \rotatebox{90}{\parbox{5em}{\centering $M_\text{ego}=5$}} &\includegraphics[width=0.195\linewidth]{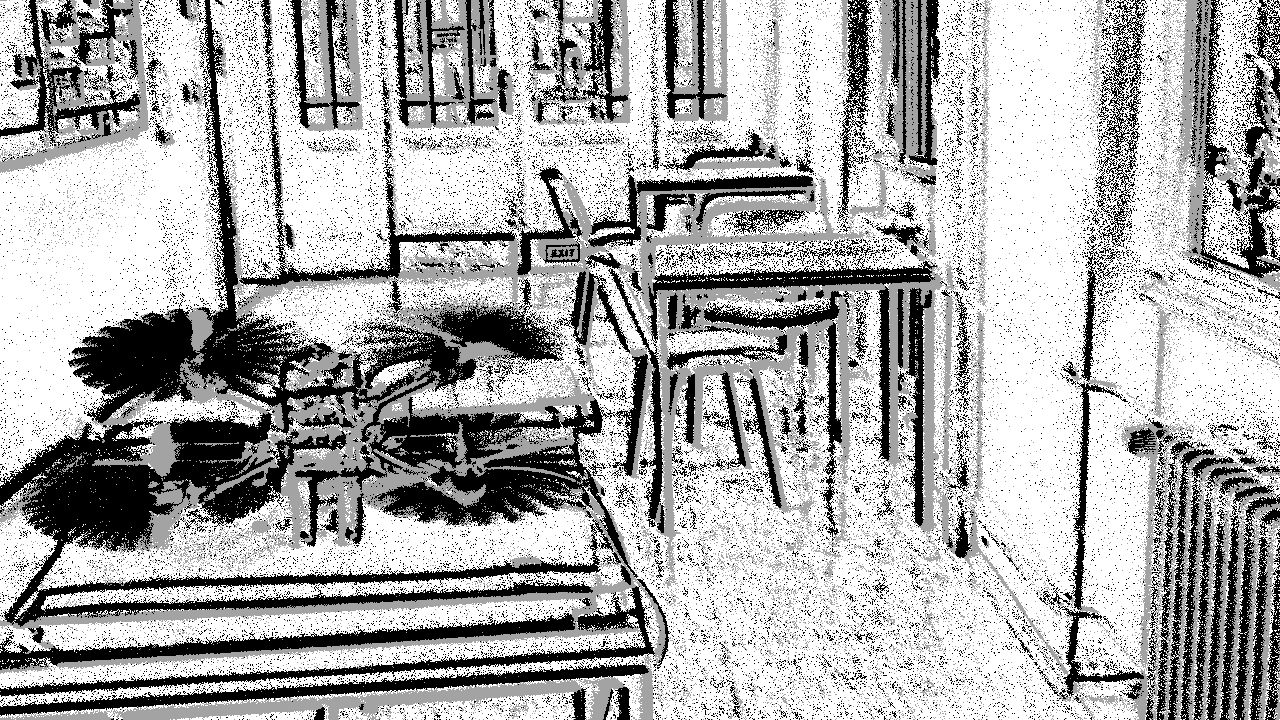} & \includegraphics[width=0.195\linewidth]{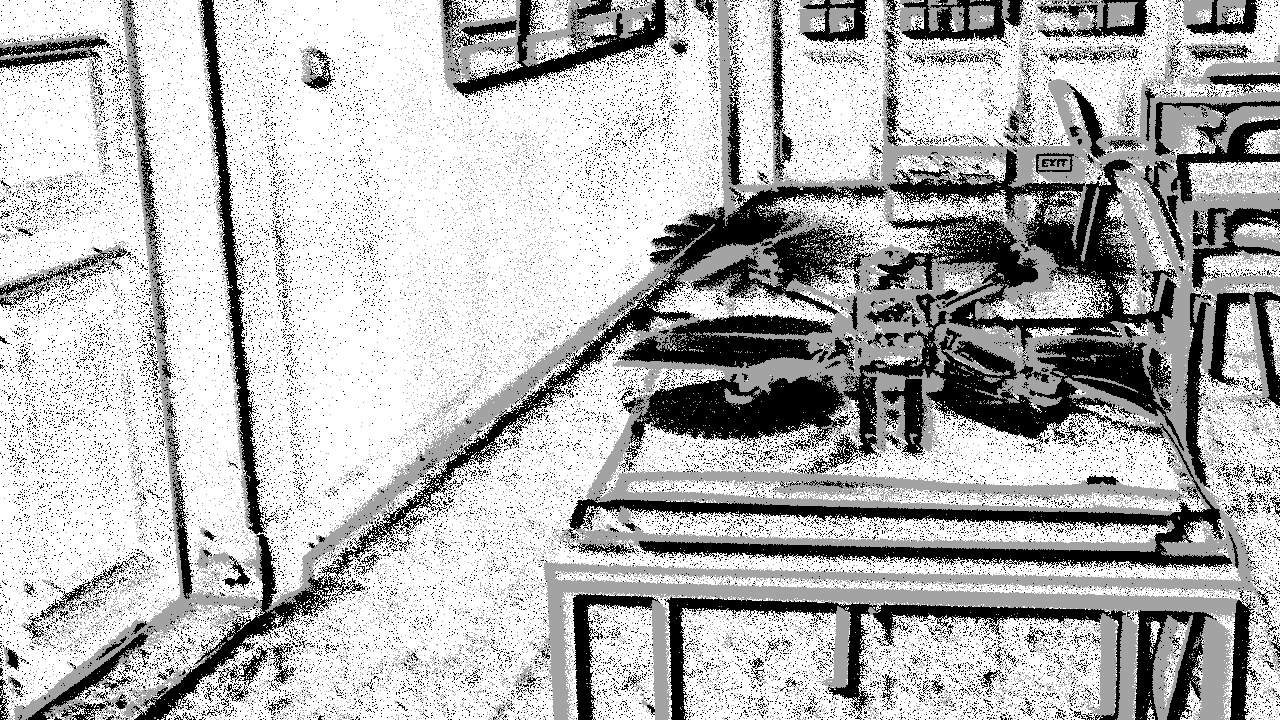} & \includegraphics[width=0.195\linewidth]{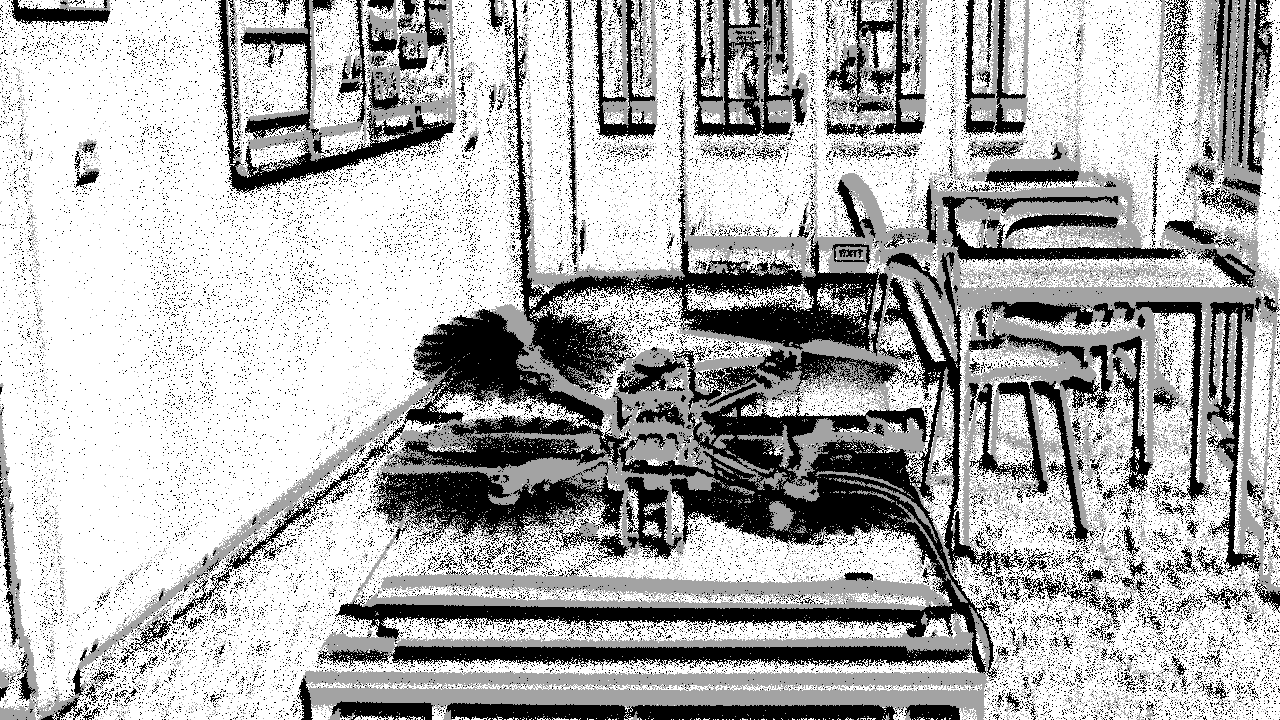} & \includegraphics[width=0.195\linewidth]{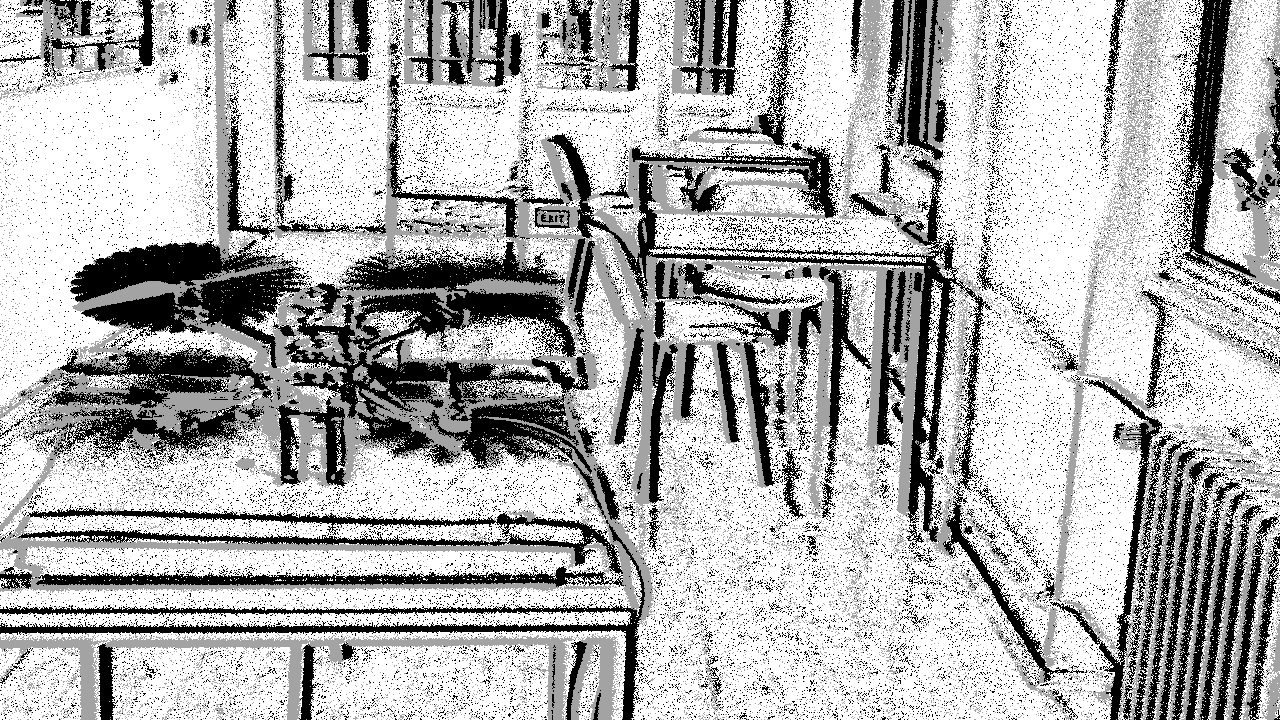} & \includegraphics[width=0.195\linewidth]{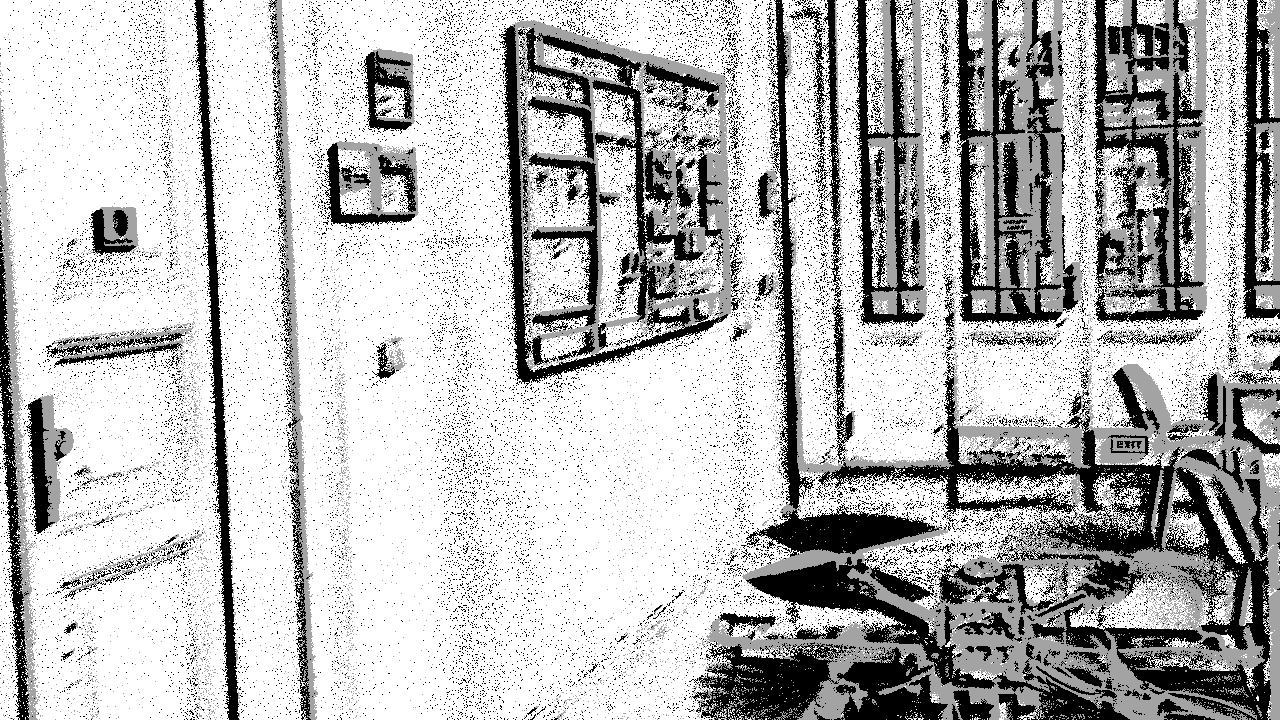}  \\
\hline
 \rotatebox{90}{\parbox{5em}{\centering $M_\text{ego}=6$}} &\includegraphics[width=0.195\linewidth]{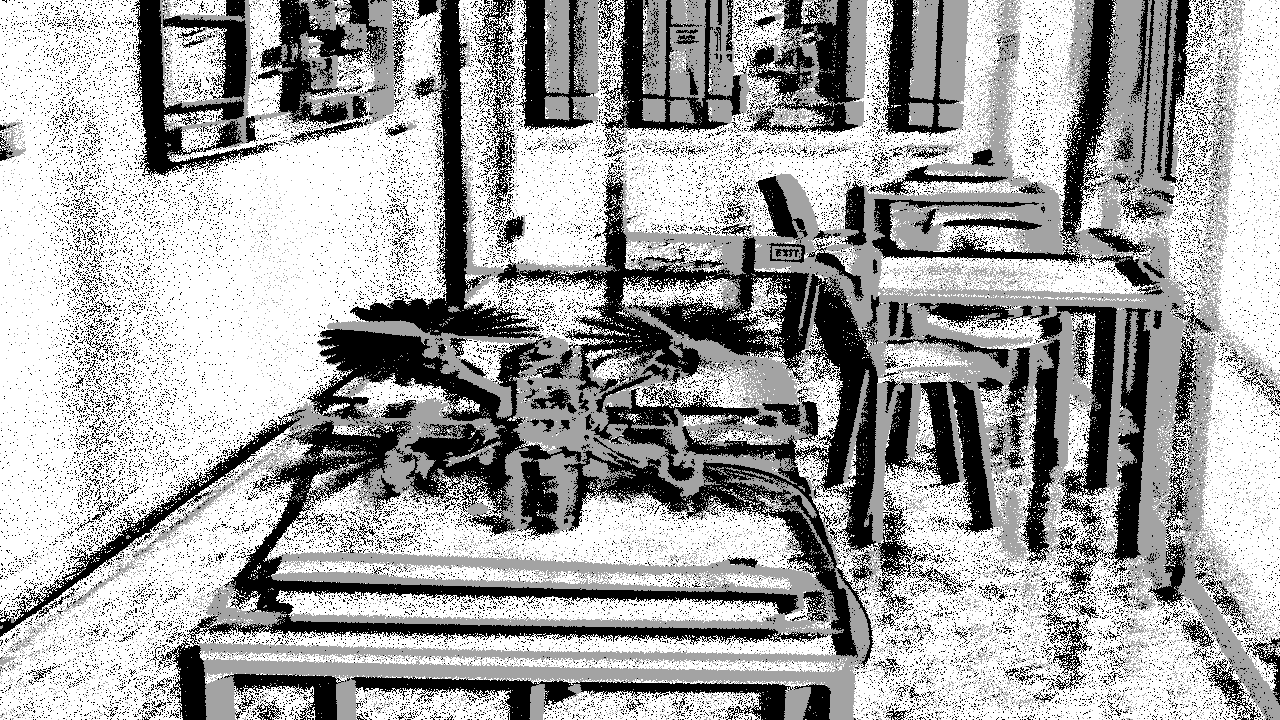} & \includegraphics[width=0.195\linewidth]{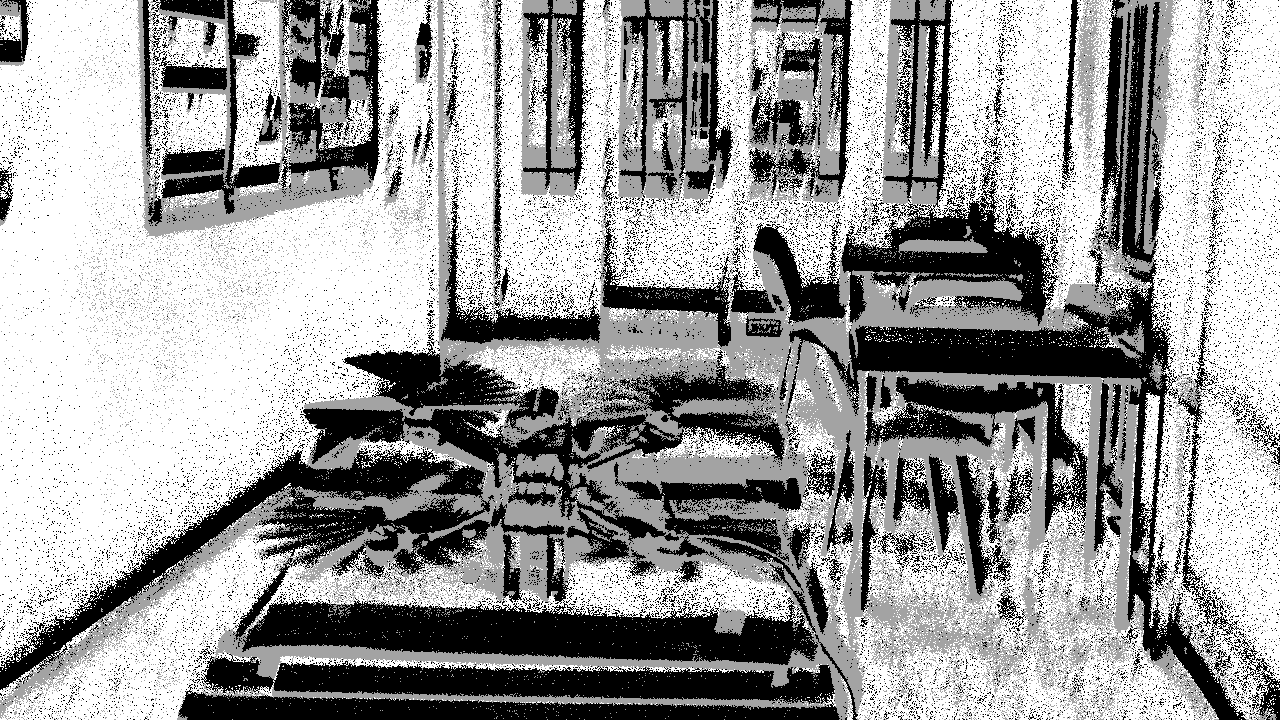} & \includegraphics[width=0.195\linewidth]{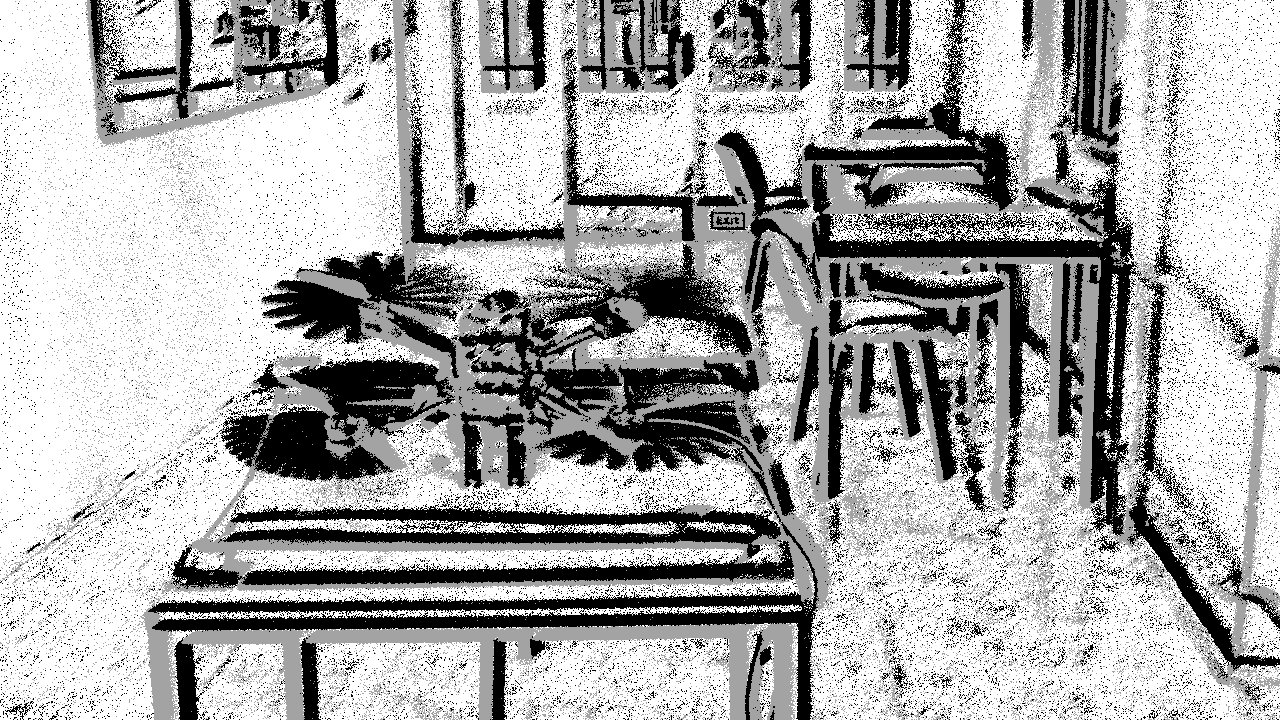} & \includegraphics[width=0.195\linewidth]{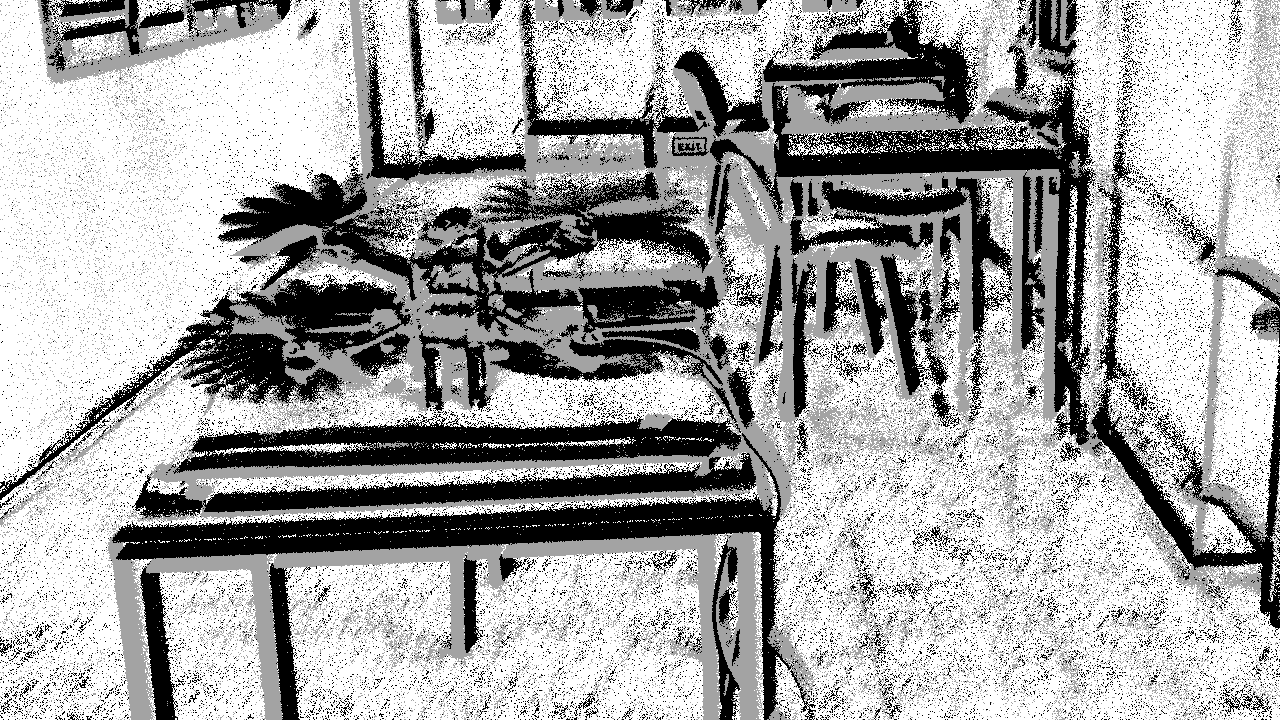} & \includegraphics[width=0.195\linewidth]{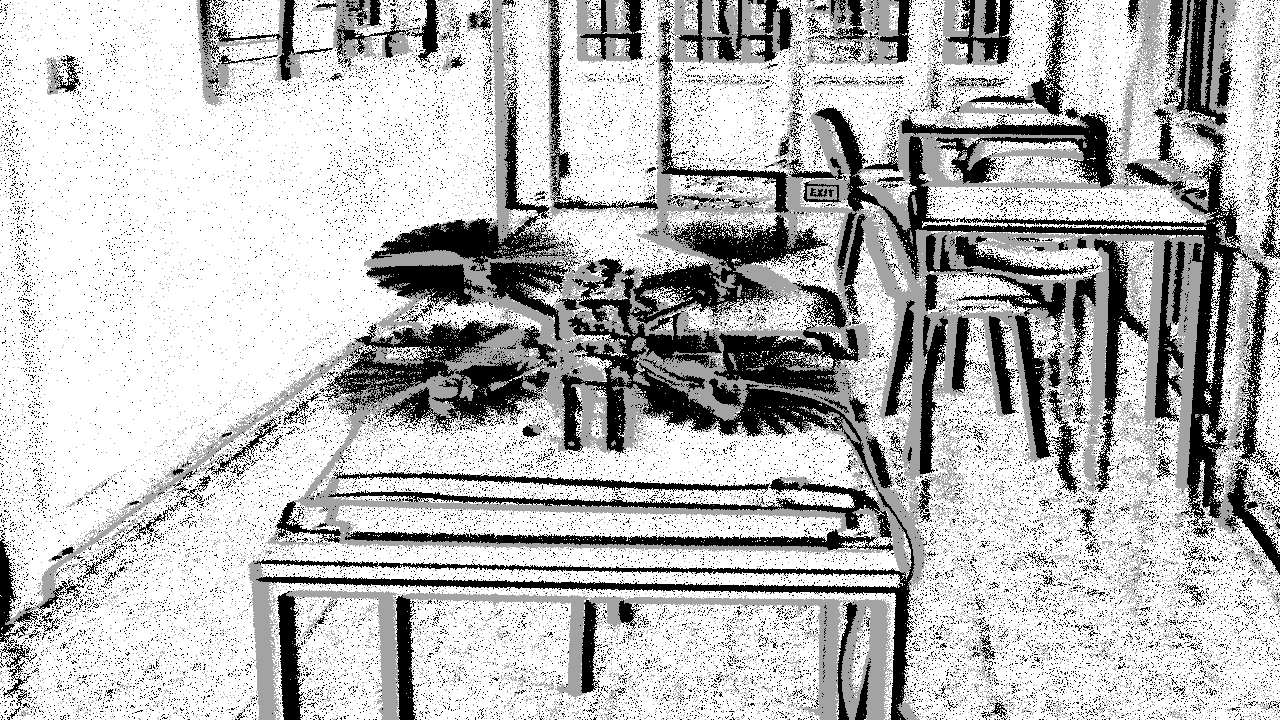}  \\
\hline
 \rotatebox{90}{\parbox{5em}{\centering $M_\text{ego}=7$}} &\includegraphics[width=0.195\linewidth]{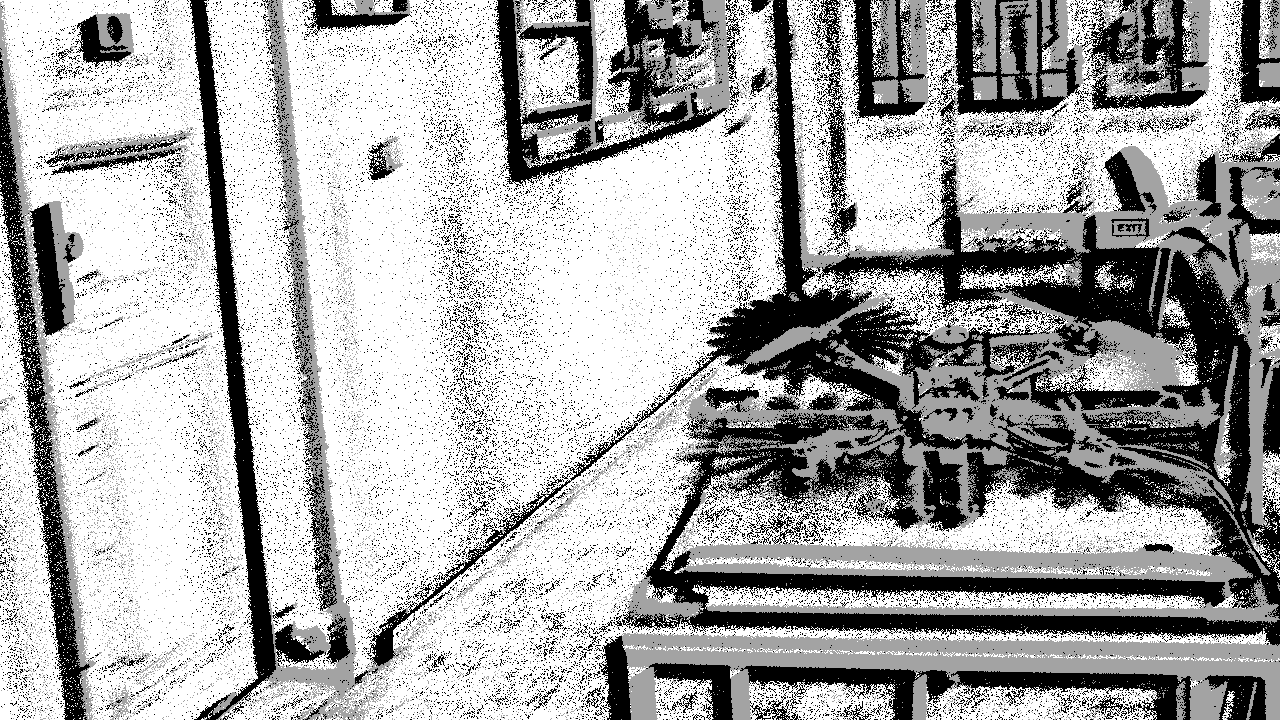} & \includegraphics[width=0.195\linewidth]{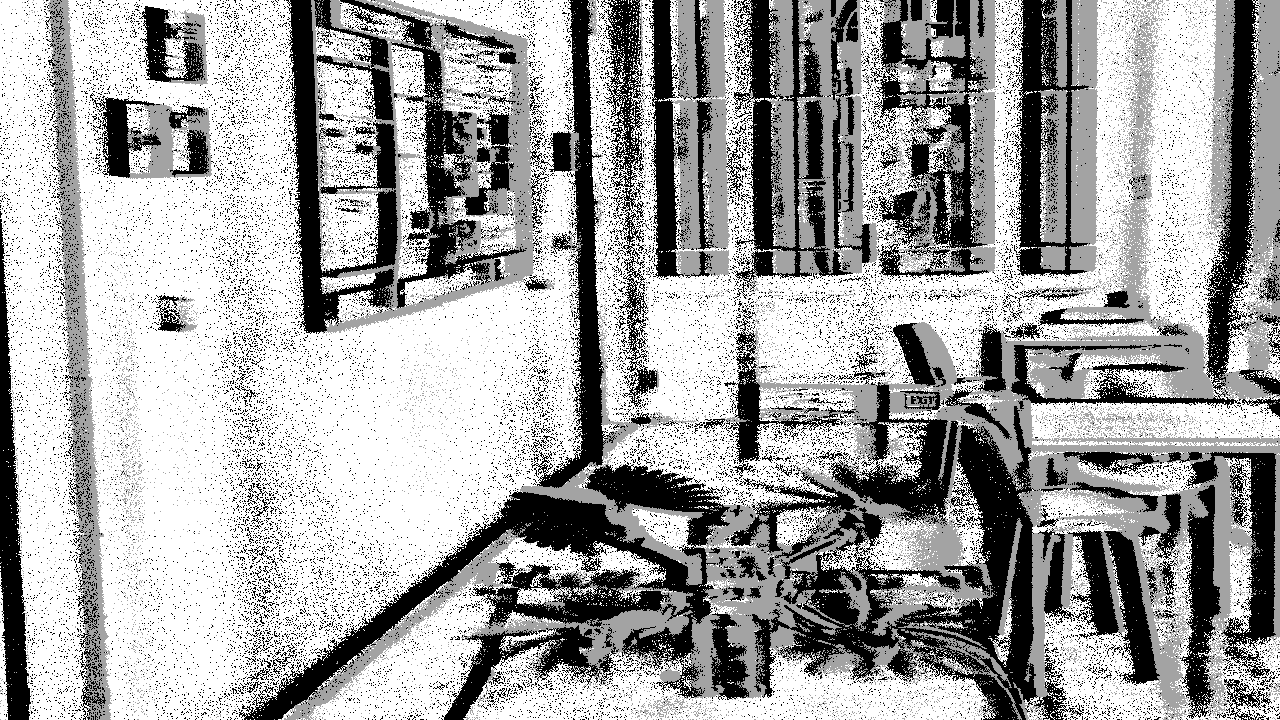} & \includegraphics[width=0.195\linewidth]{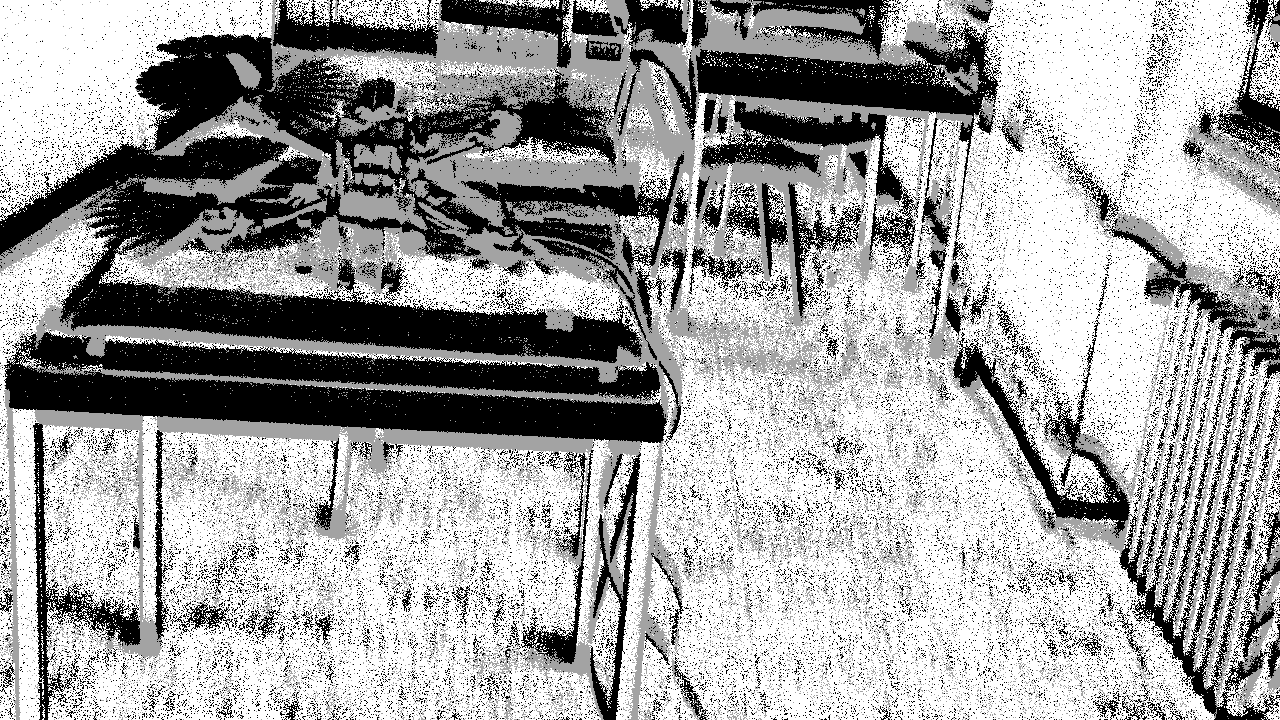} & \includegraphics[width=0.195\linewidth]{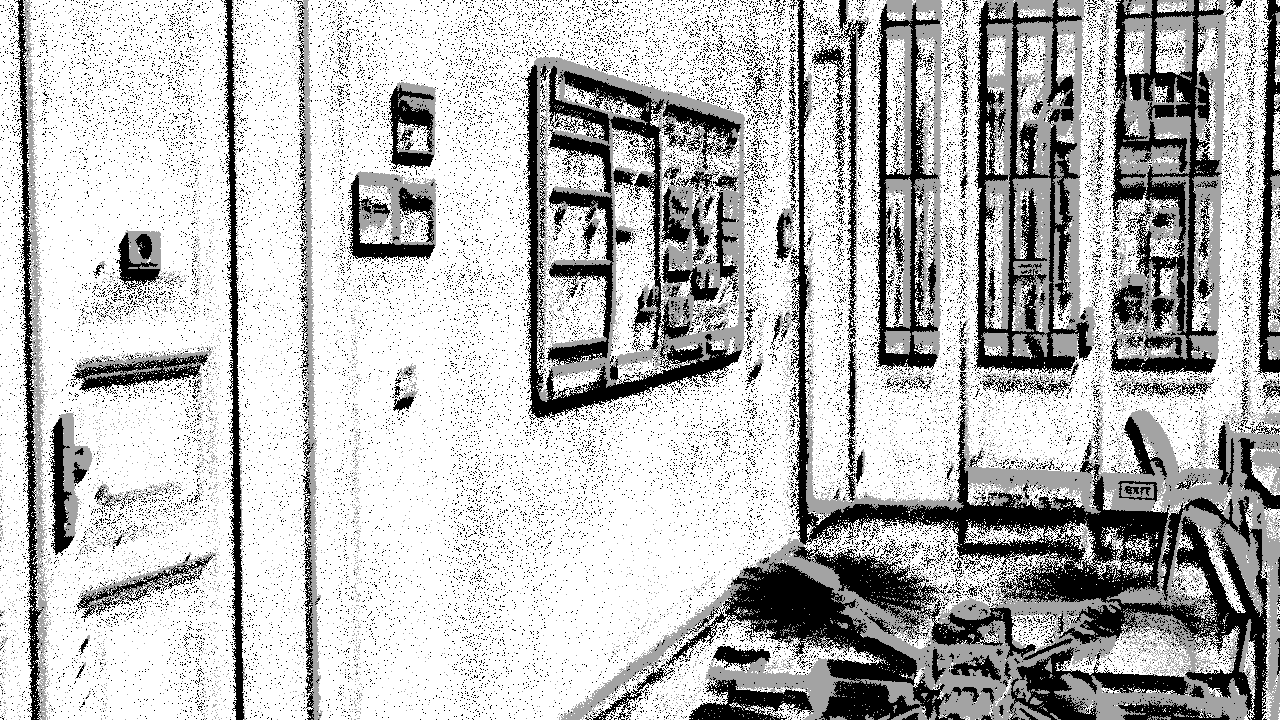} & \includegraphics[width=0.195\linewidth]{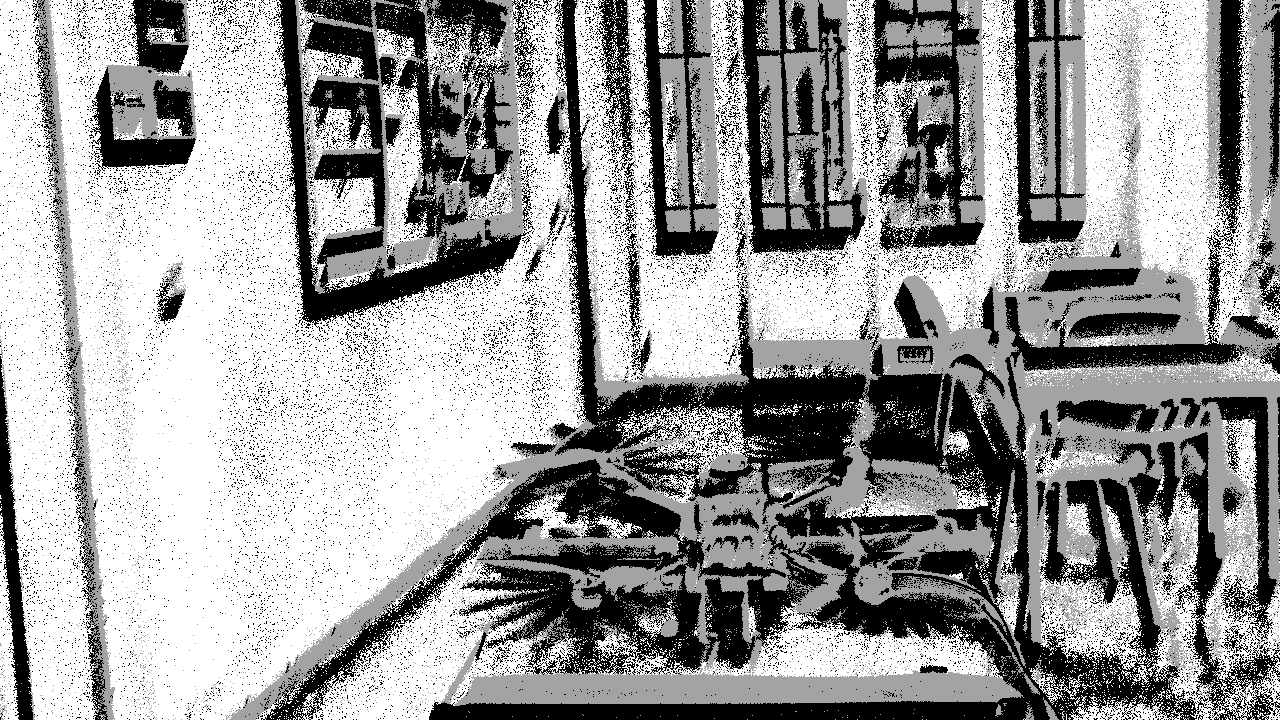}  \\
\hline
\end{tabular}

\caption{Aggregated events from all sequences in the \tquadcopterego{} dataset with camera-to-quadcopter distance of 2 m,
 from the smallest (top row, $M_{\text{ego}}$=1), to the largest (bottom row, $M_\text{ego}$=7) egomotion intensity.
 Events from the first 10 ms of five 1-second intervals are shown.
 Positive events displayed in black, negative in gray.
 }
\label{fig:dataset_ego_sequenced_overview_full_dist2}
\end{figure*}

\begin{figure*}[ht]
\renewcommand{\arraystretch}{0}\centering
\begin{tabular}{|@{}c@{}|@{}c@{}|@{}c@{}|@{}c@{}|@{}c@{}|@{}c@{}|}
\hline
 \rotatebox{90}{\parbox{5em}{\centering $M_\text{ego}=1$}} &\includegraphics[width=0.195\linewidth]{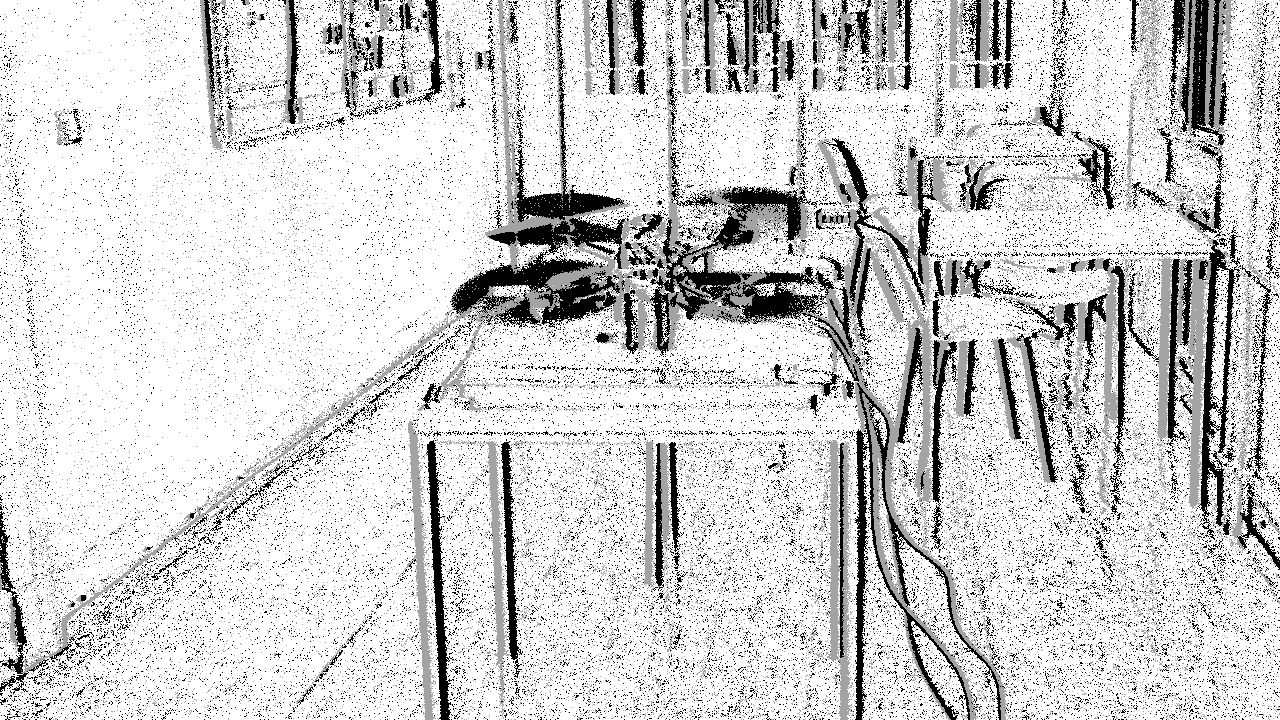} & \includegraphics[width=0.195\linewidth]{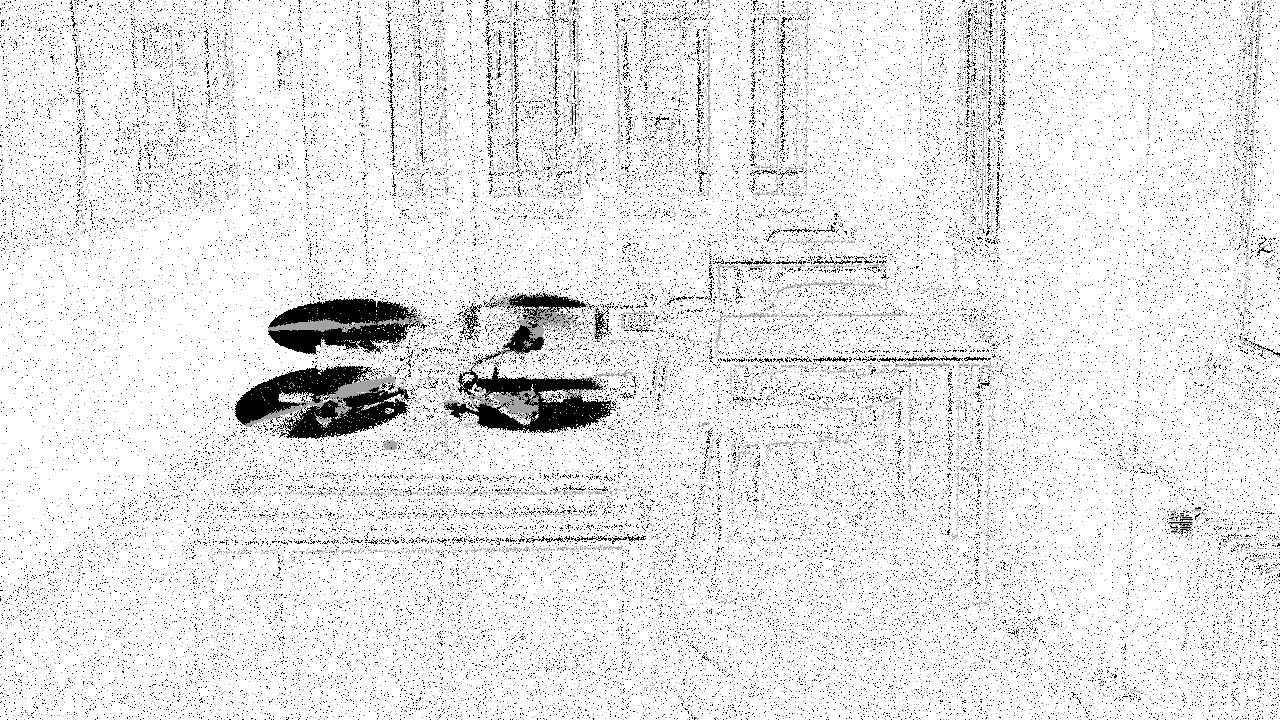} & \includegraphics[width=0.195\linewidth]{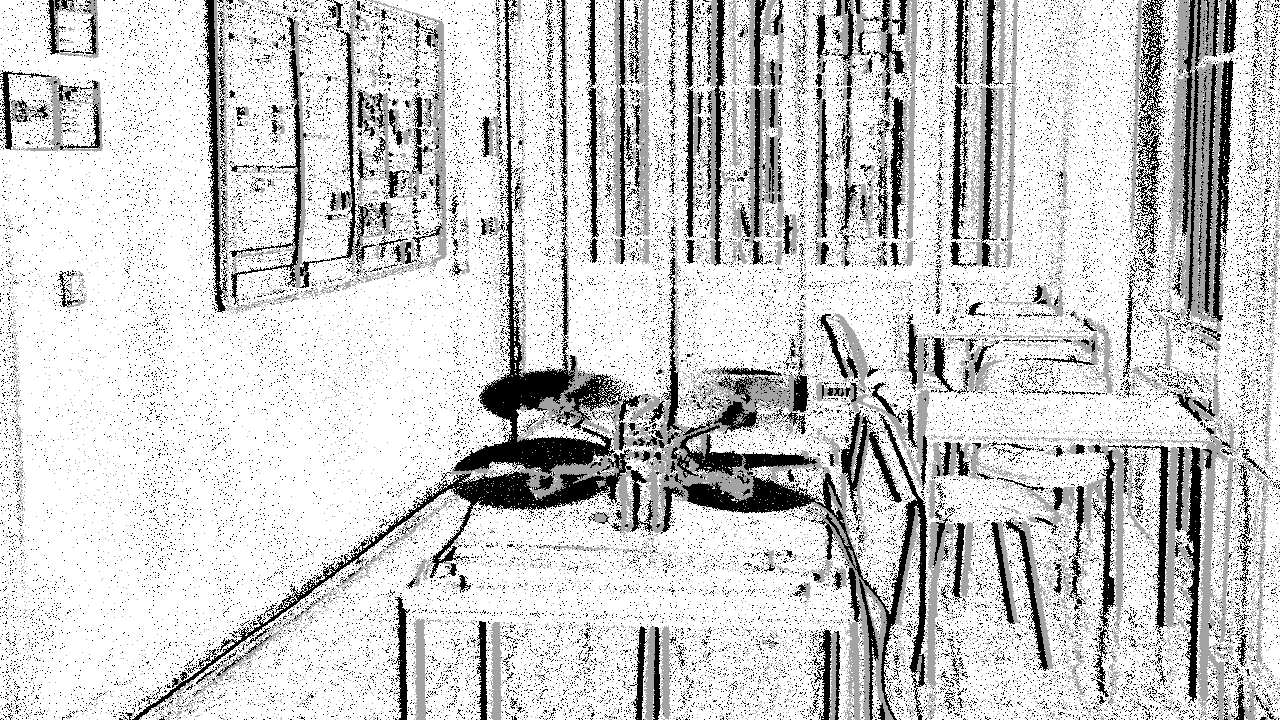} & \includegraphics[width=0.195\linewidth]{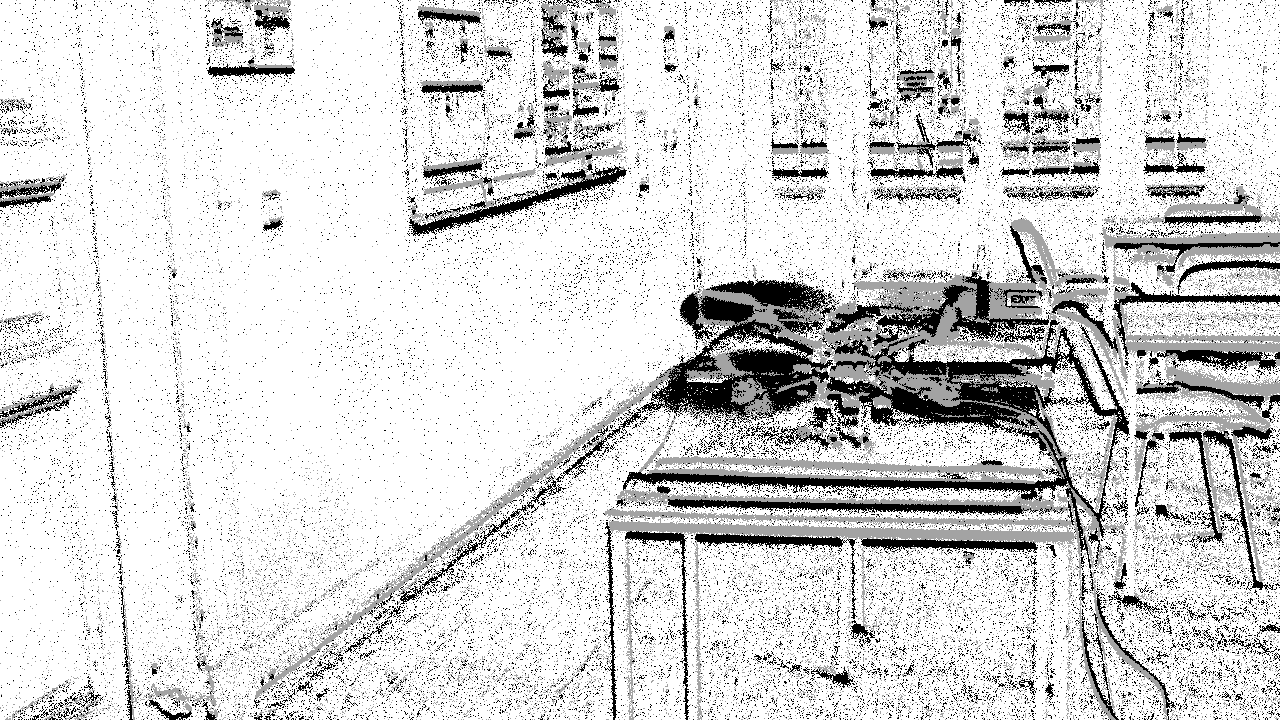} & \includegraphics[width=0.195\linewidth]{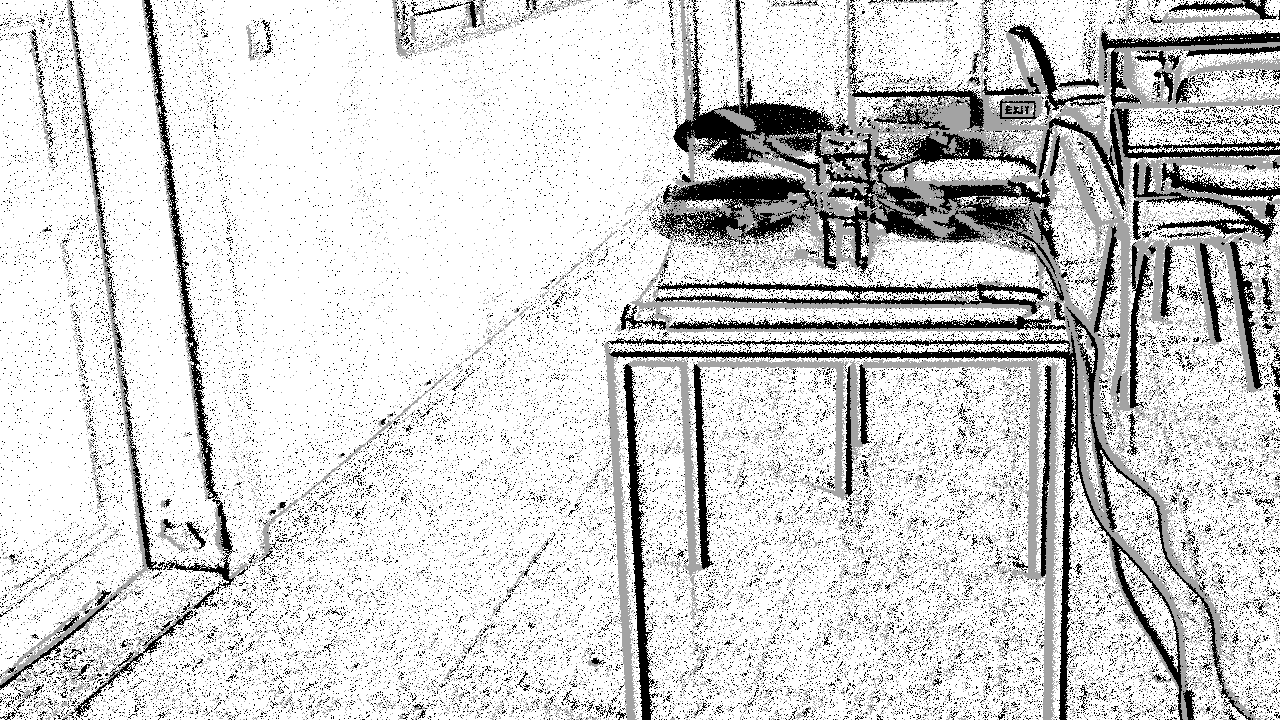}  \\
\hline
 \rotatebox{90}{\parbox{5em}{\centering $M_\text{ego}=2$}} &\includegraphics[width=0.195\linewidth]{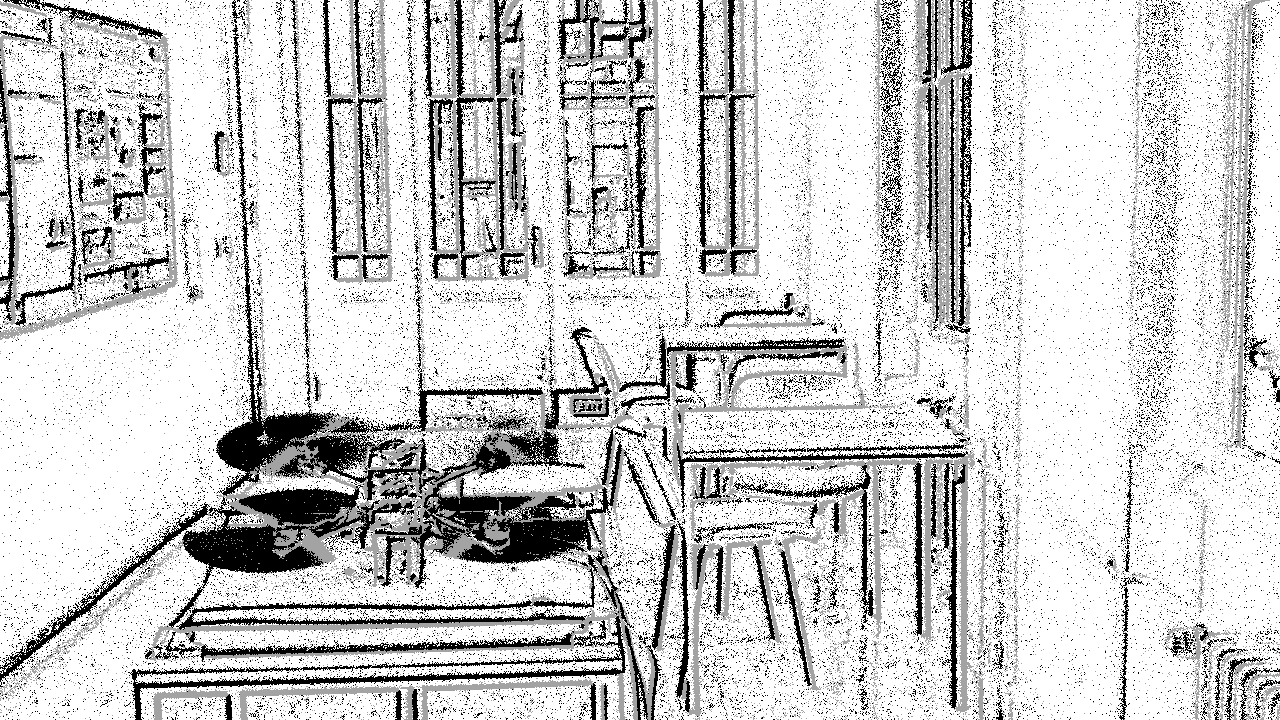} & \includegraphics[width=0.195\linewidth]{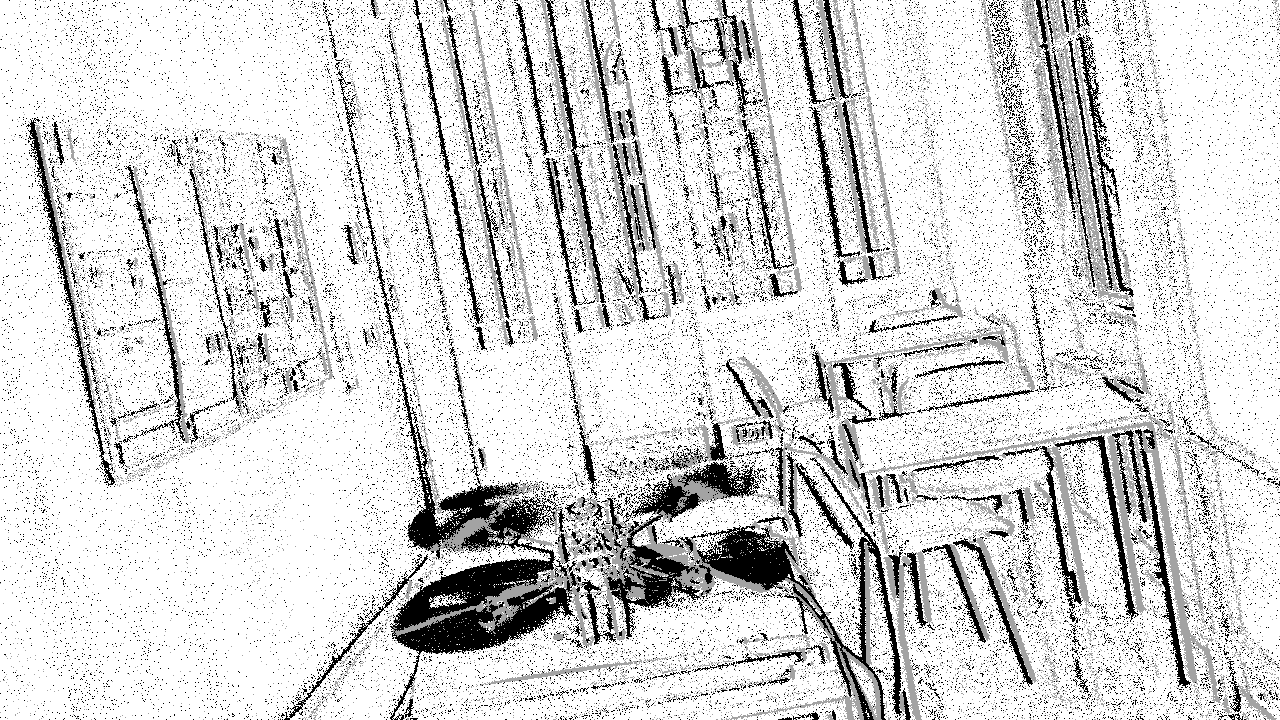} & \includegraphics[width=0.195\linewidth]{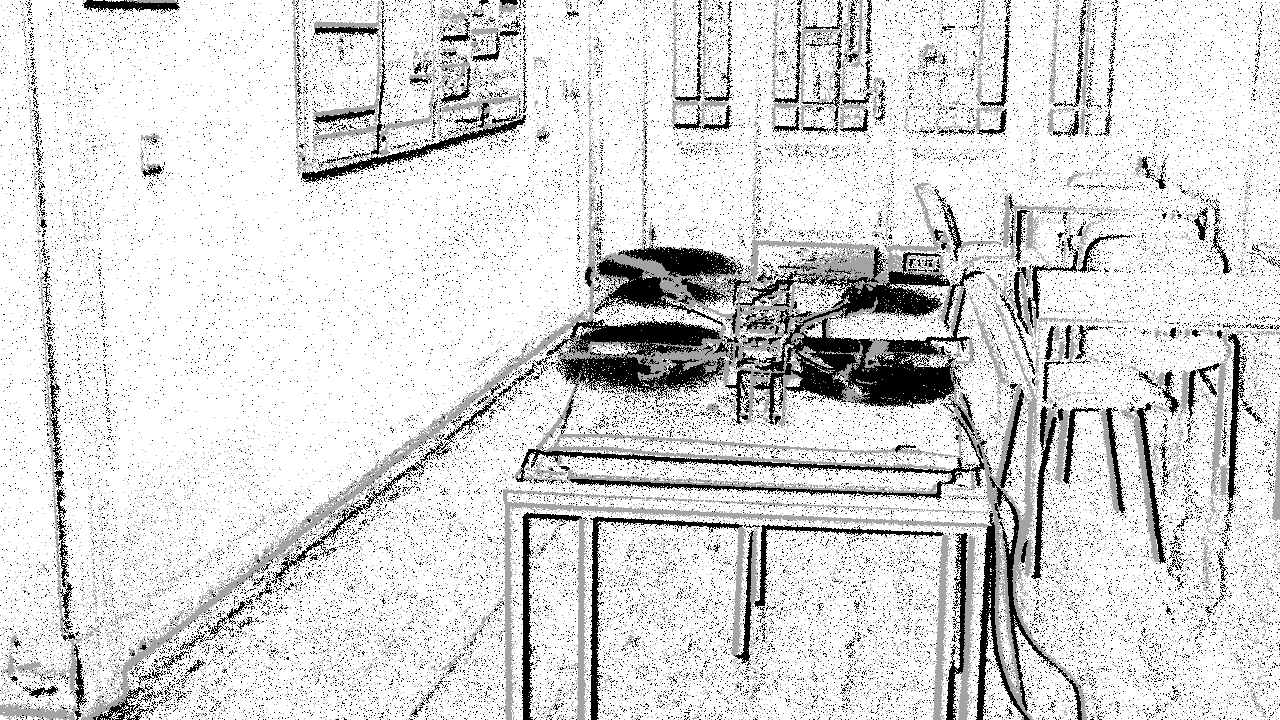} & \includegraphics[width=0.195\linewidth]{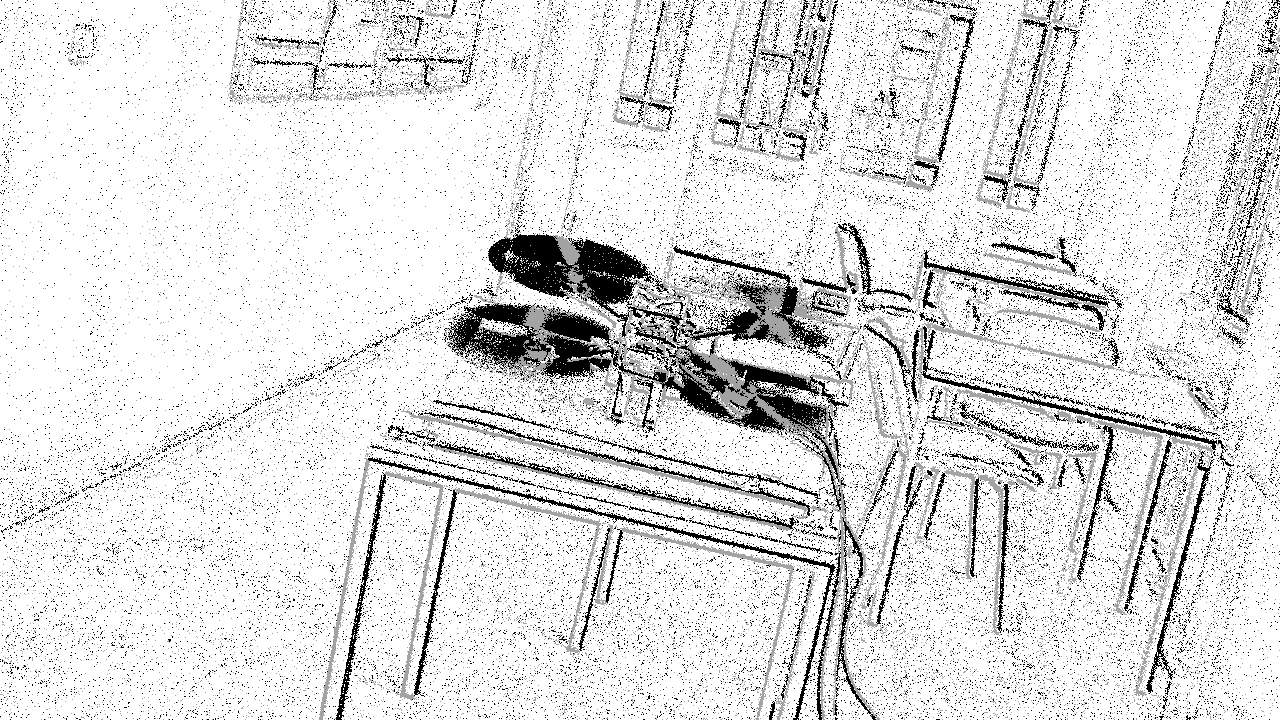} & \includegraphics[width=0.195\linewidth]{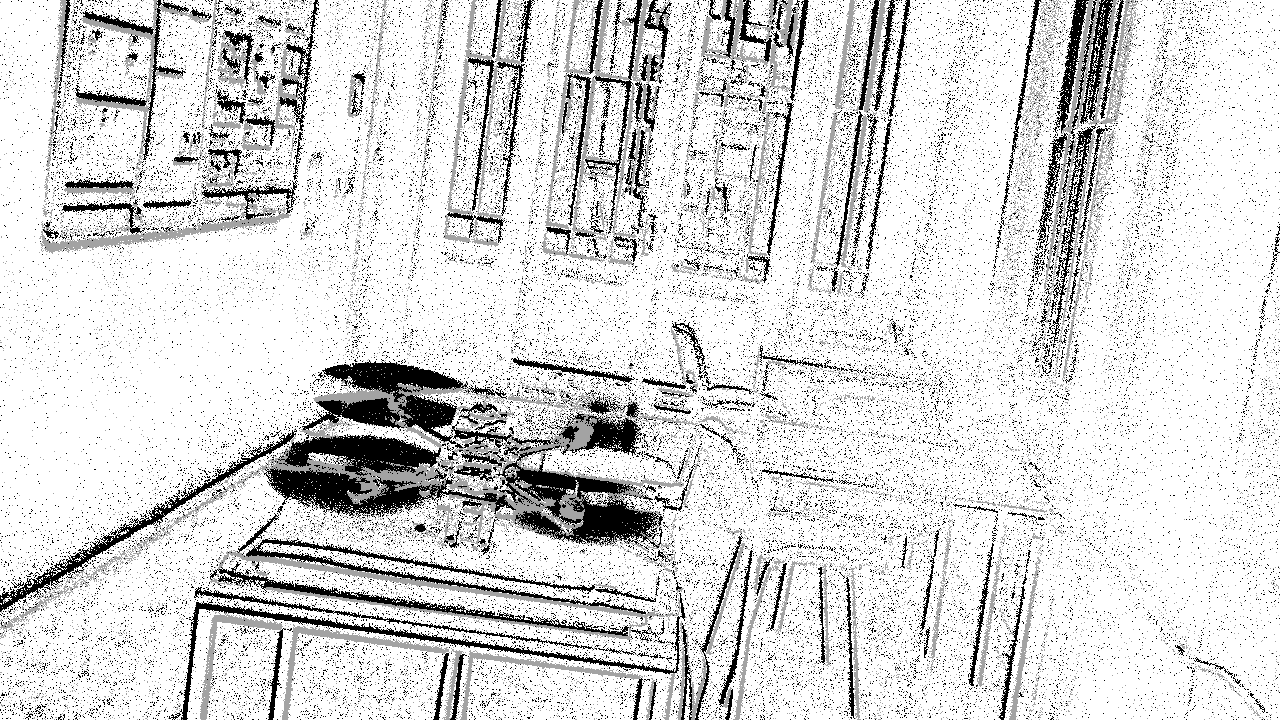}  \\
\hline
 \rotatebox{90}{\parbox{5em}{\centering $M_\text{ego}=3$}} &\includegraphics[width=0.195\linewidth]{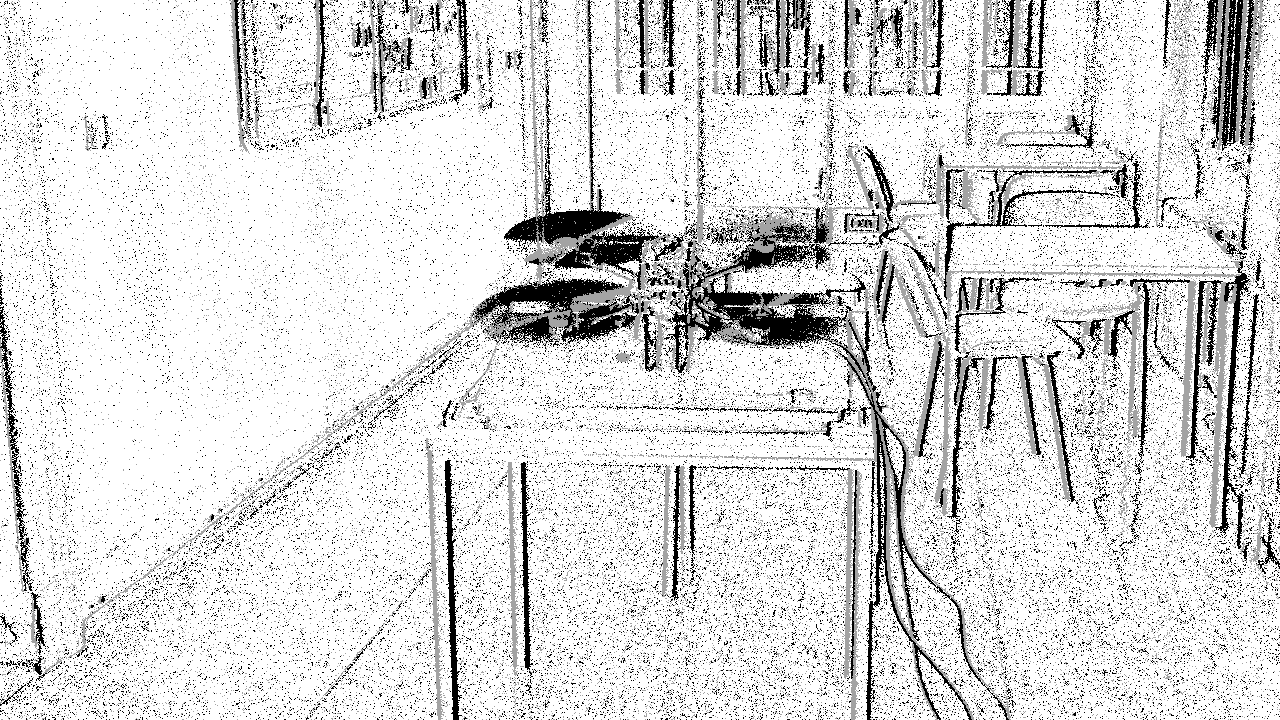} & \includegraphics[width=0.195\linewidth]{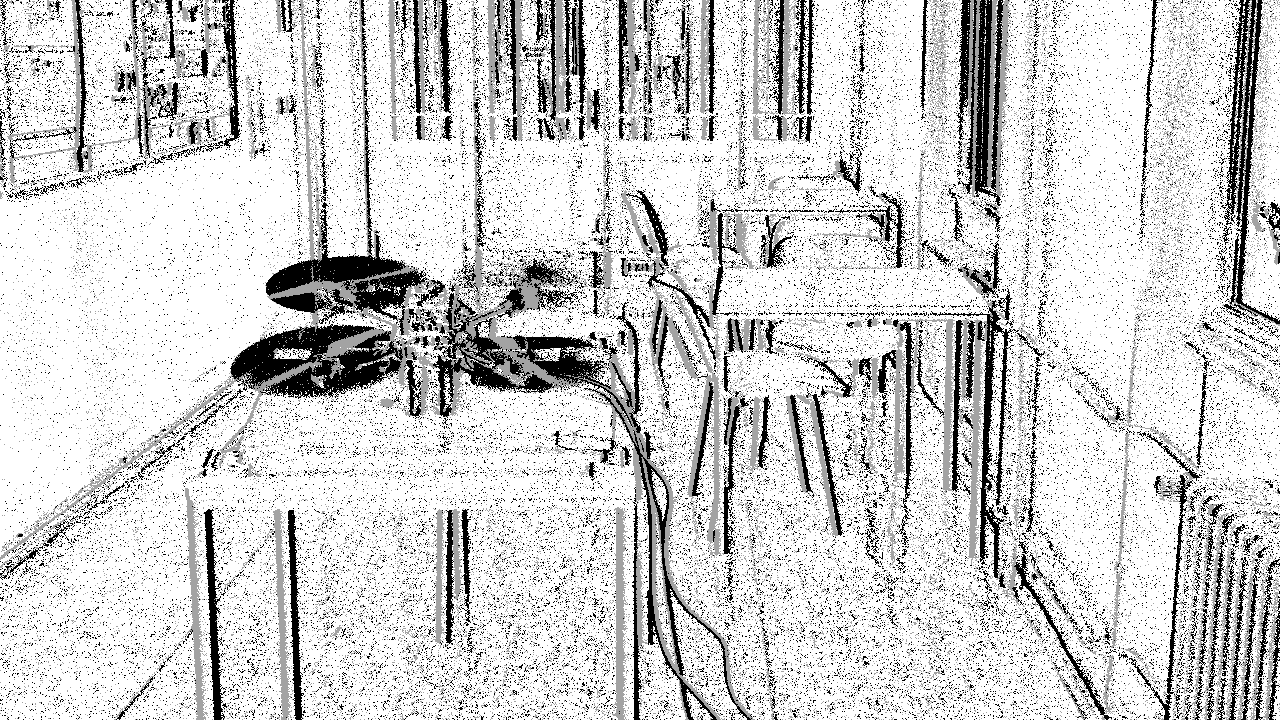} & \includegraphics[width=0.195\linewidth]{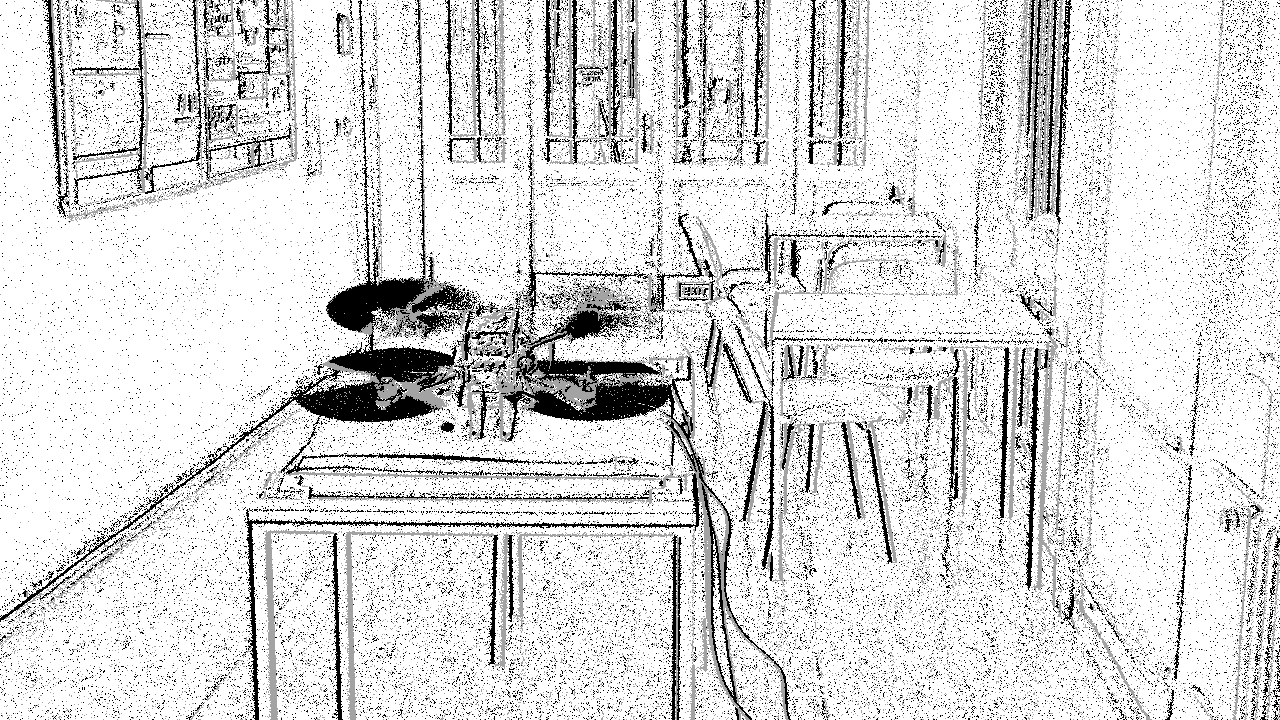} & \includegraphics[width=0.195\linewidth]{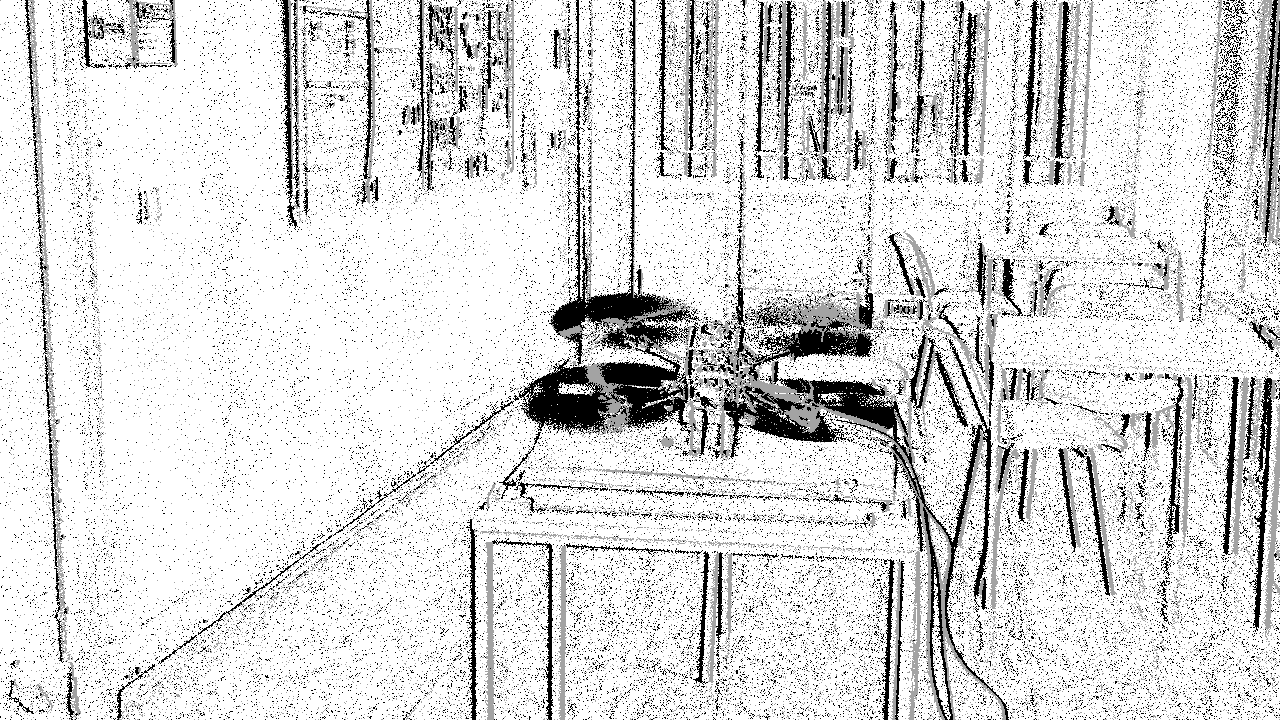} & \includegraphics[width=0.195\linewidth]{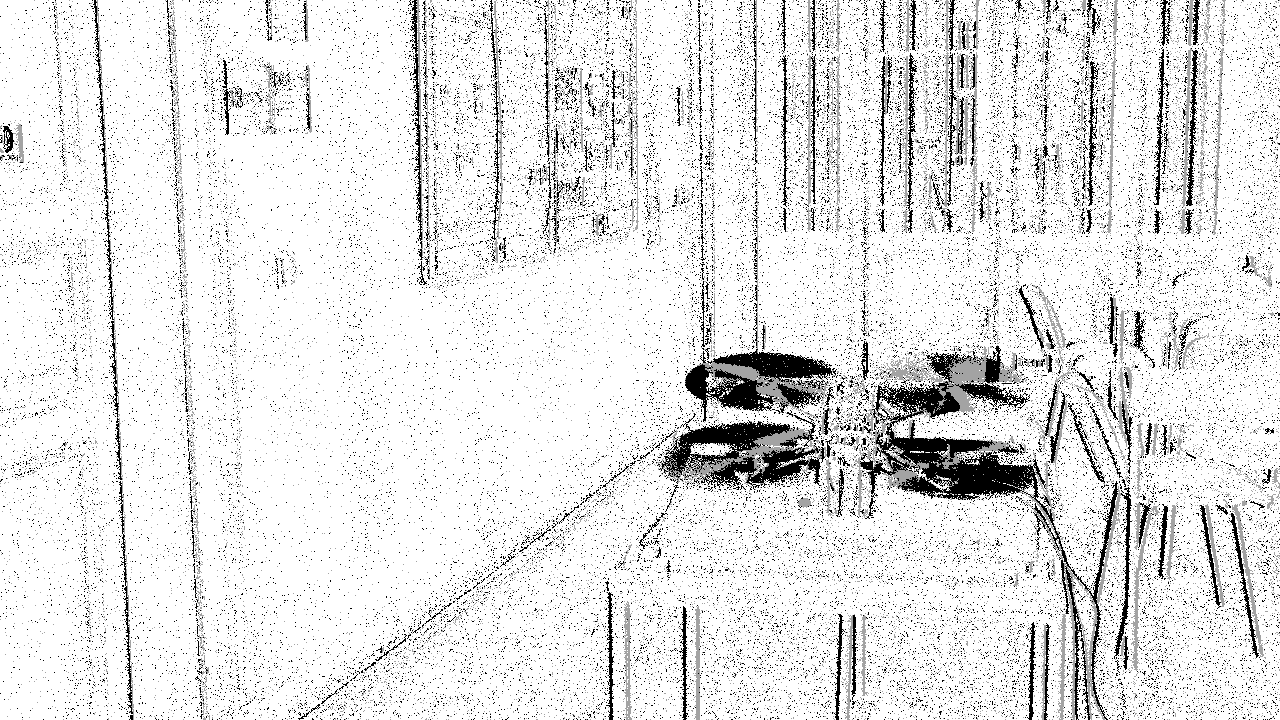}  \\
\hline
 \rotatebox{90}{\parbox{5em}{\centering $M_\text{ego}=4$}} &\includegraphics[width=0.195\linewidth]{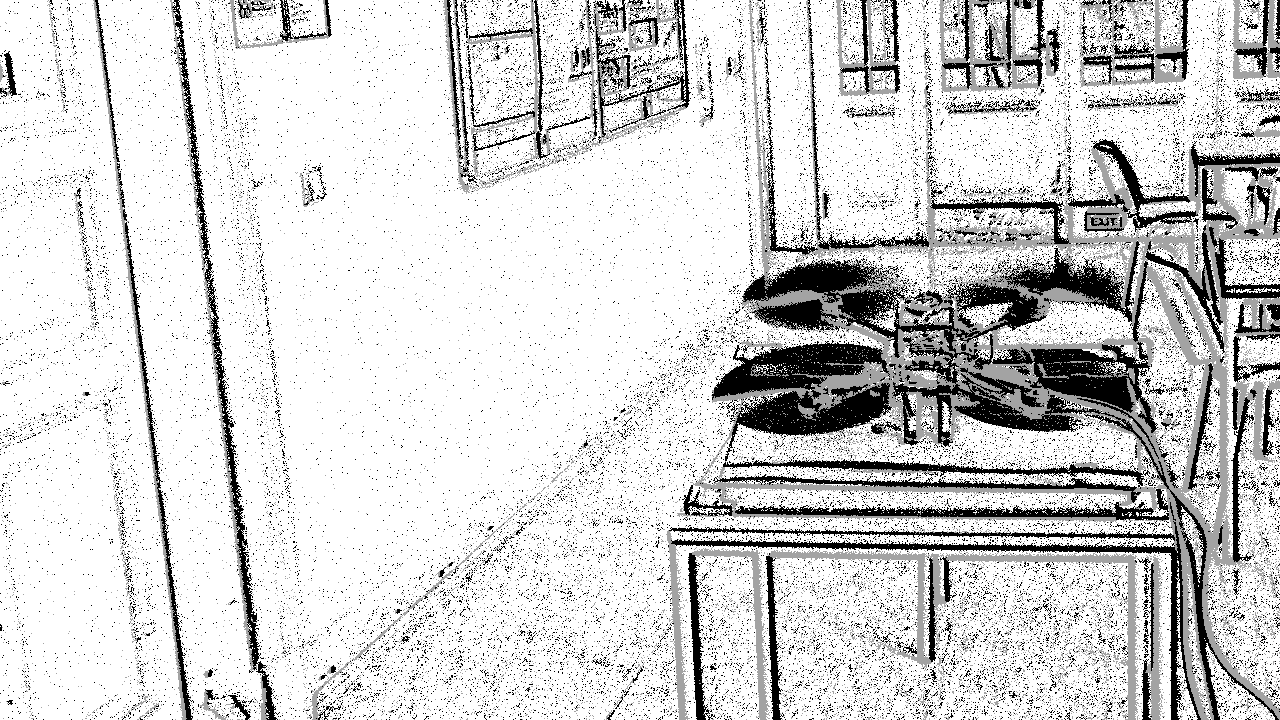} & \includegraphics[width=0.195\linewidth]{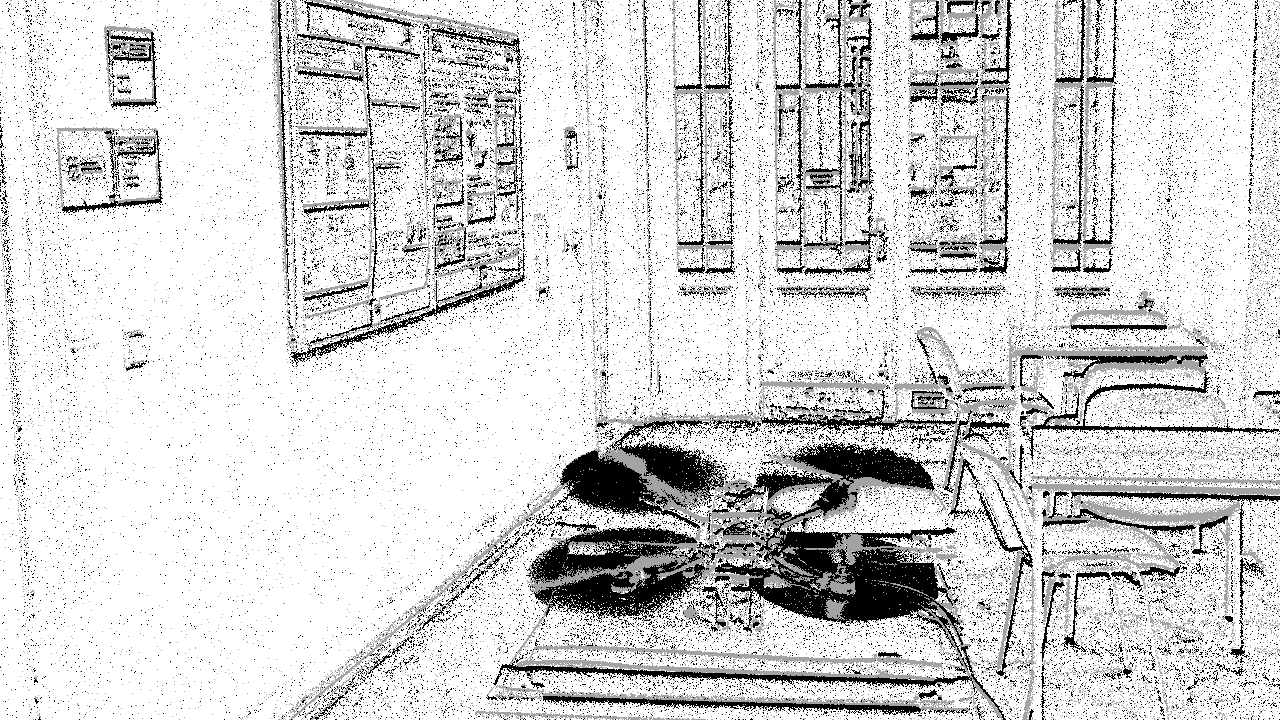} & \includegraphics[width=0.195\linewidth]{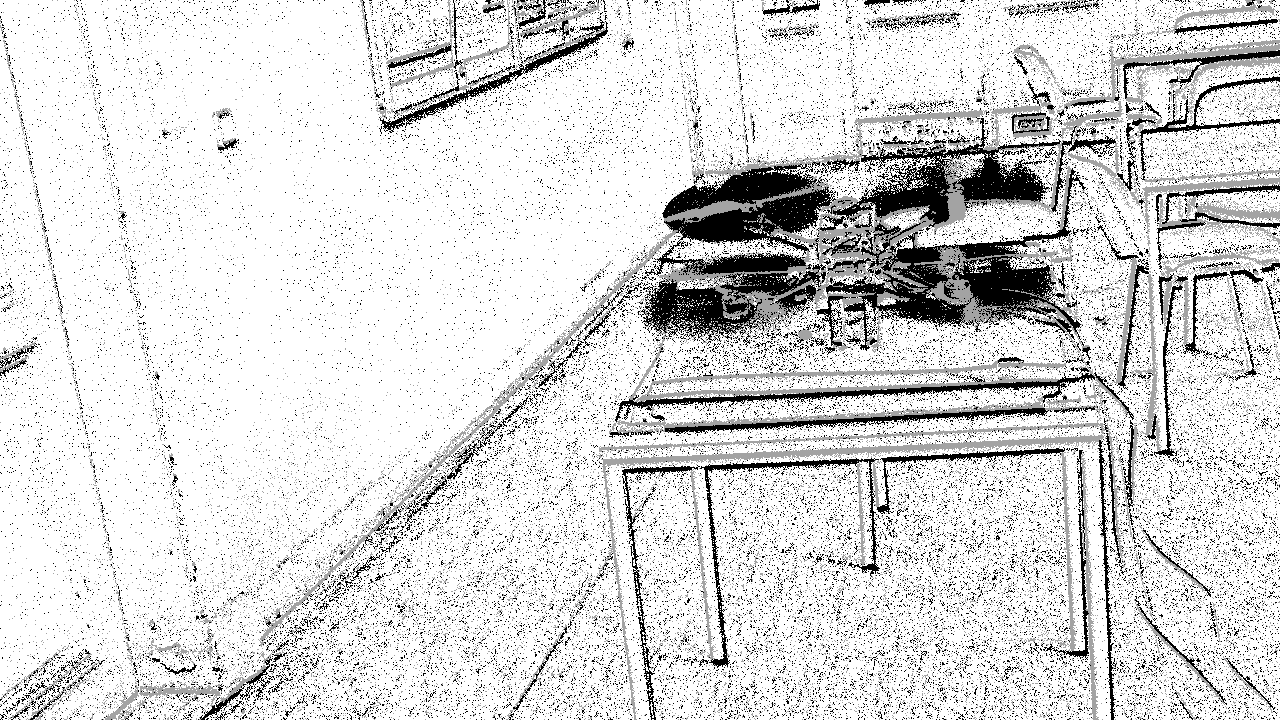} & \includegraphics[width=0.195\linewidth]{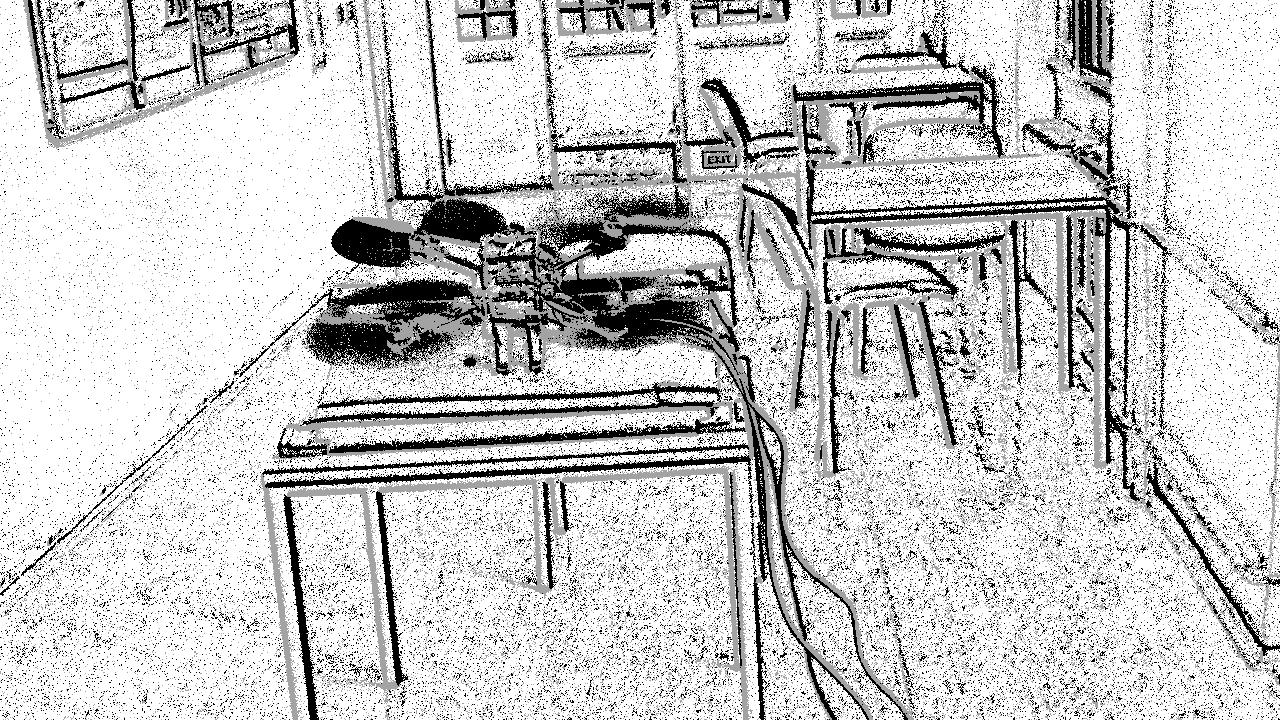} & \includegraphics[width=0.195\linewidth]{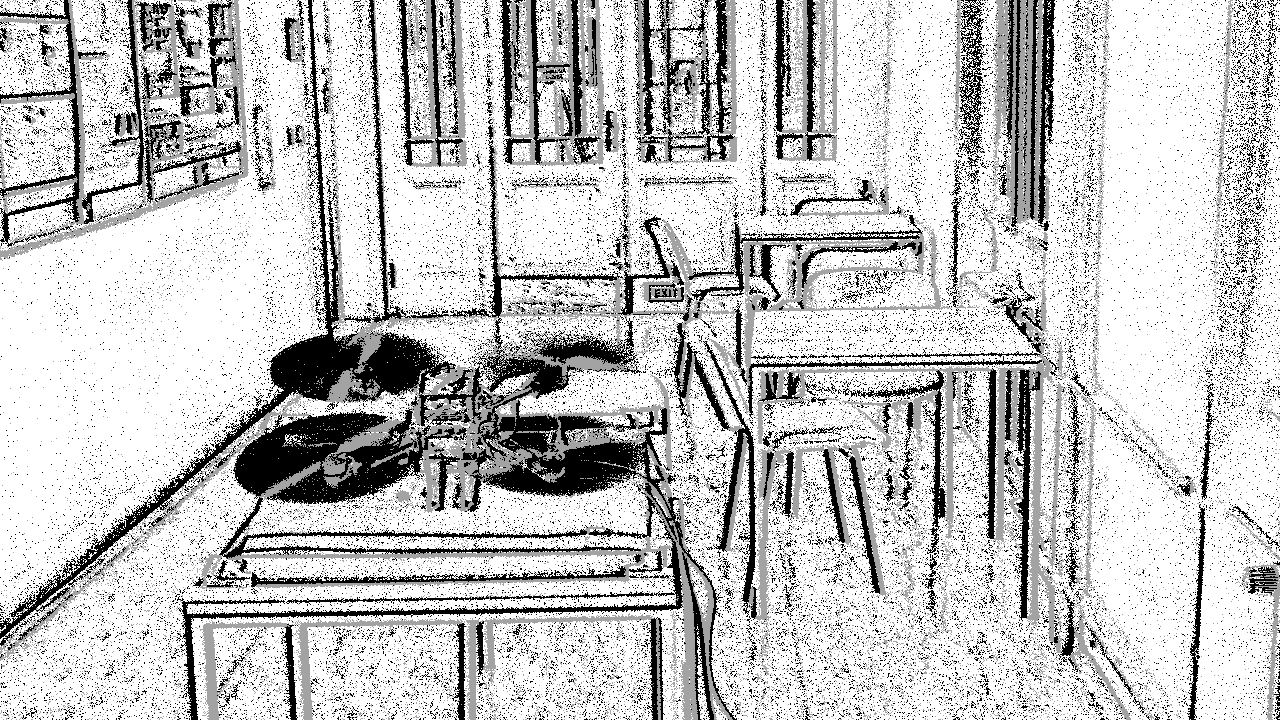}  \\
\hline
 \rotatebox{90}{\parbox{5em}{\centering $M_\text{ego}=5$}} &\includegraphics[width=0.195\linewidth]{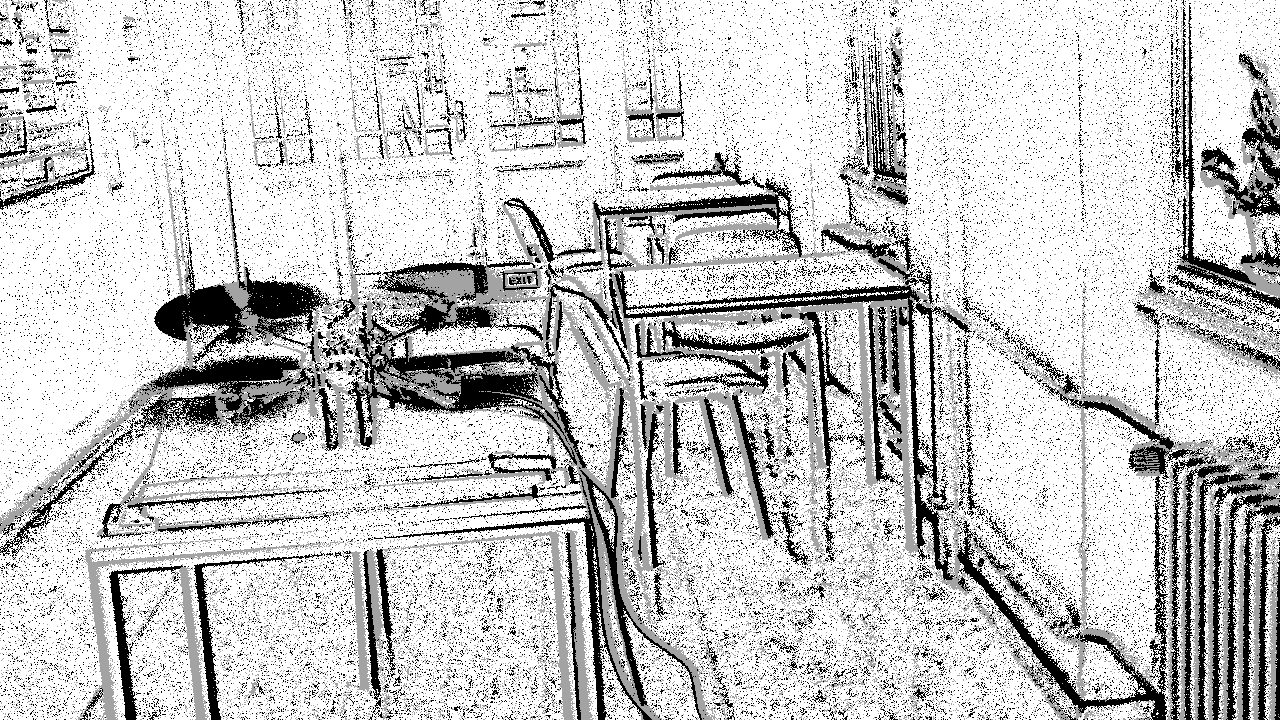} & \includegraphics[width=0.195\linewidth]{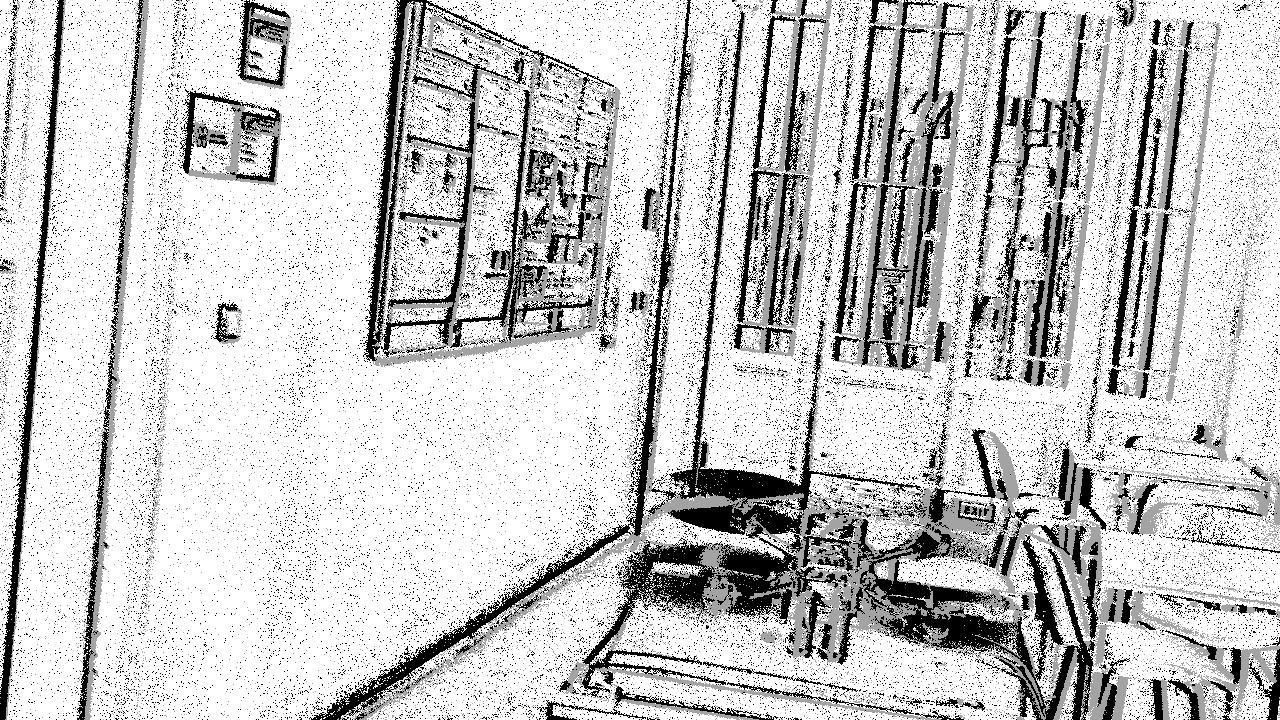} & \includegraphics[width=0.195\linewidth]{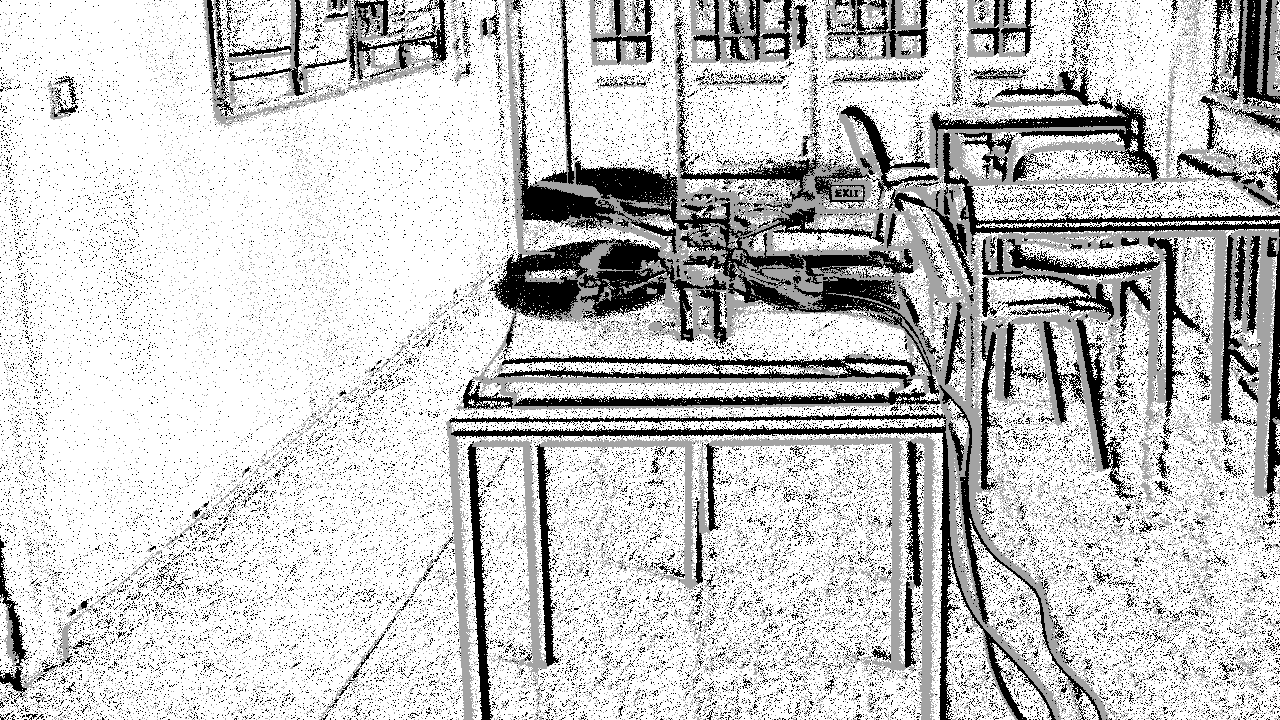} & \includegraphics[width=0.195\linewidth]{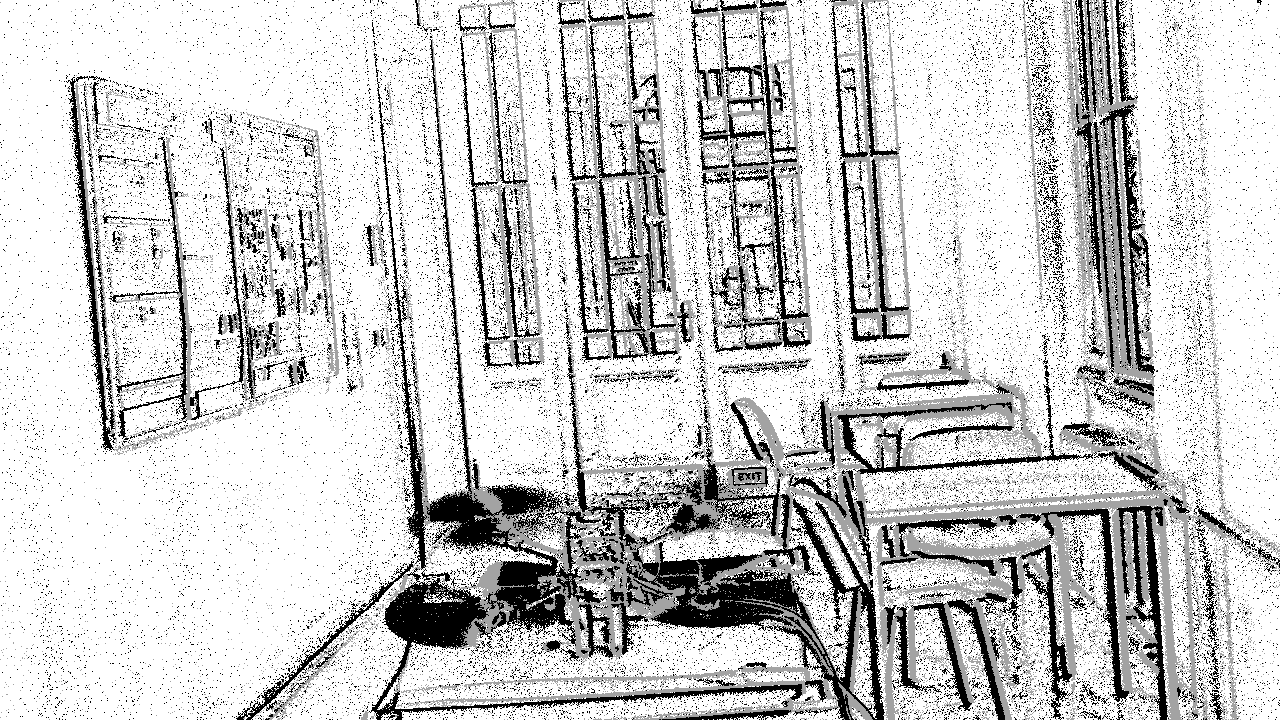} & \includegraphics[width=0.195\linewidth]{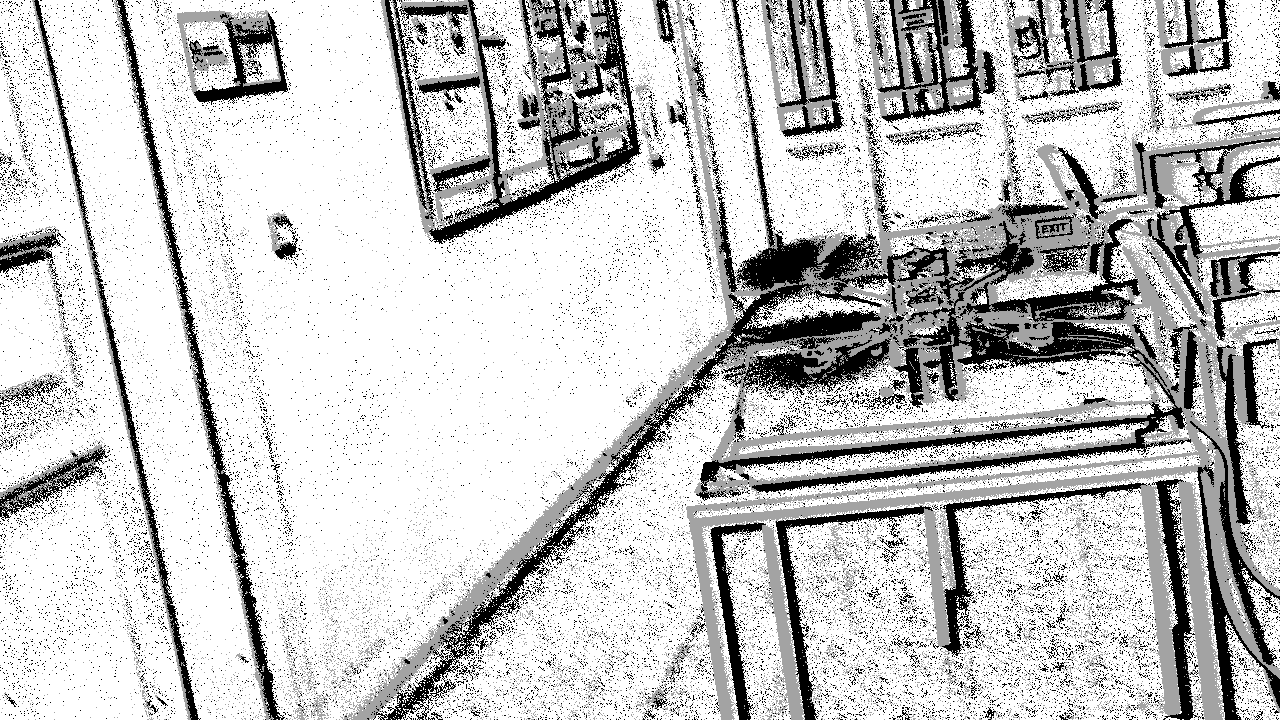}  \\
\hline
 \rotatebox{90}{\parbox{5em}{\centering $M_\text{ego}=6$}} &\includegraphics[width=0.195\linewidth]{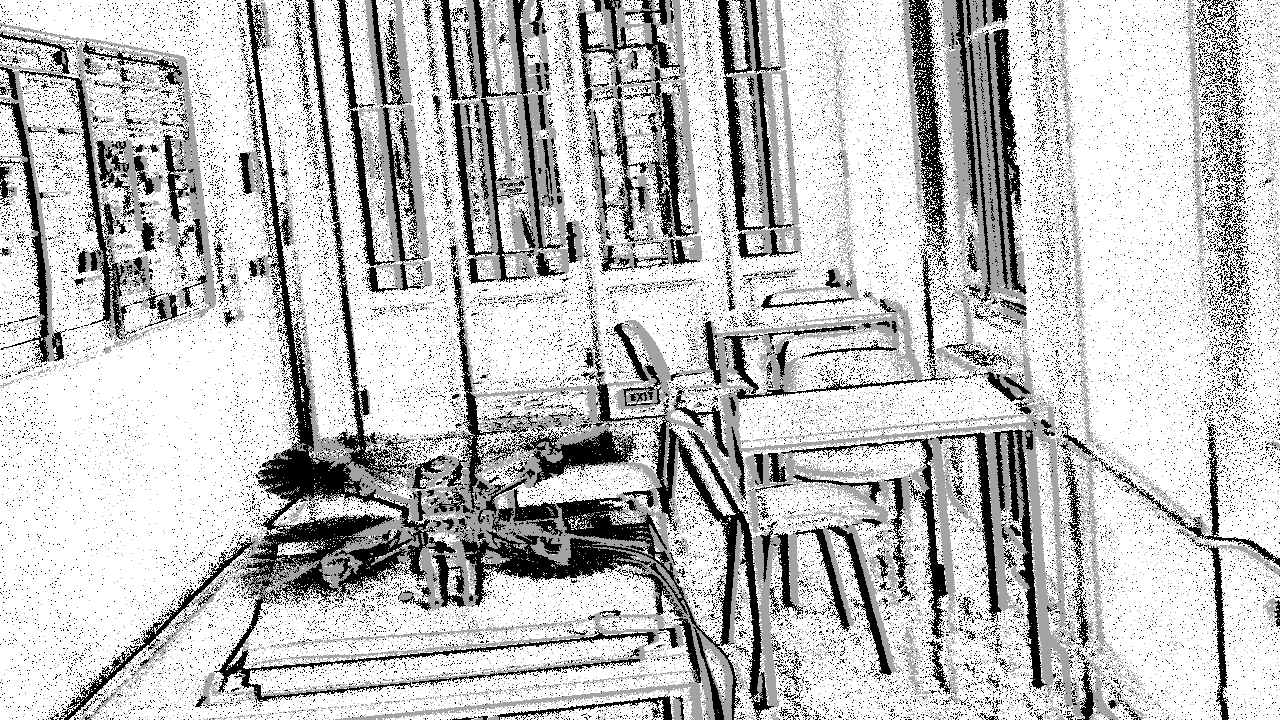} & \includegraphics[width=0.195\linewidth]{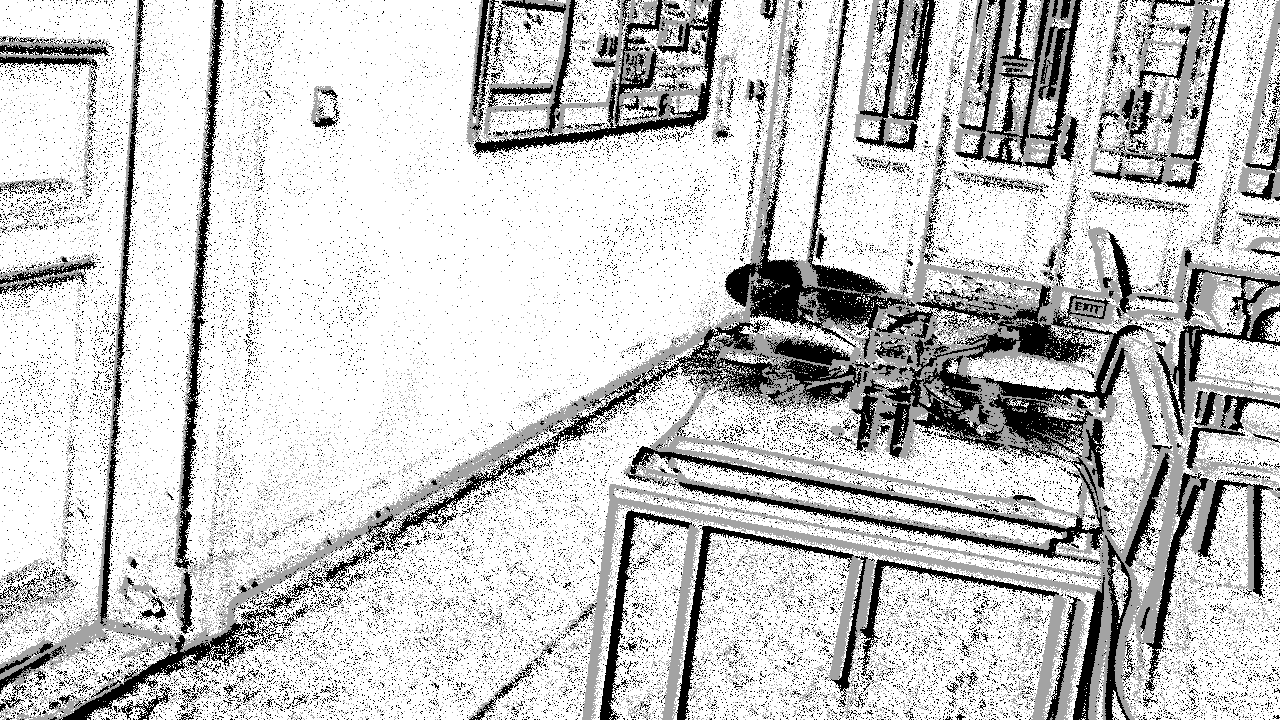} & \includegraphics[width=0.195\linewidth]{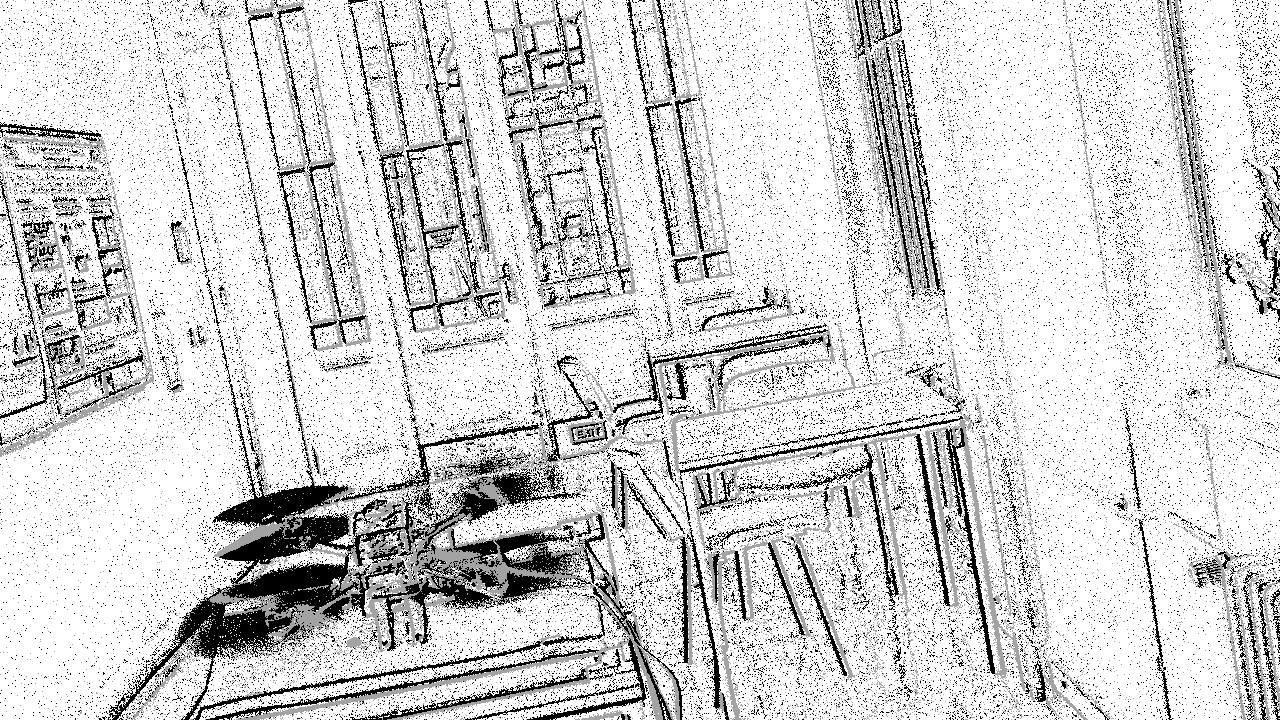} & \includegraphics[width=0.195\linewidth]{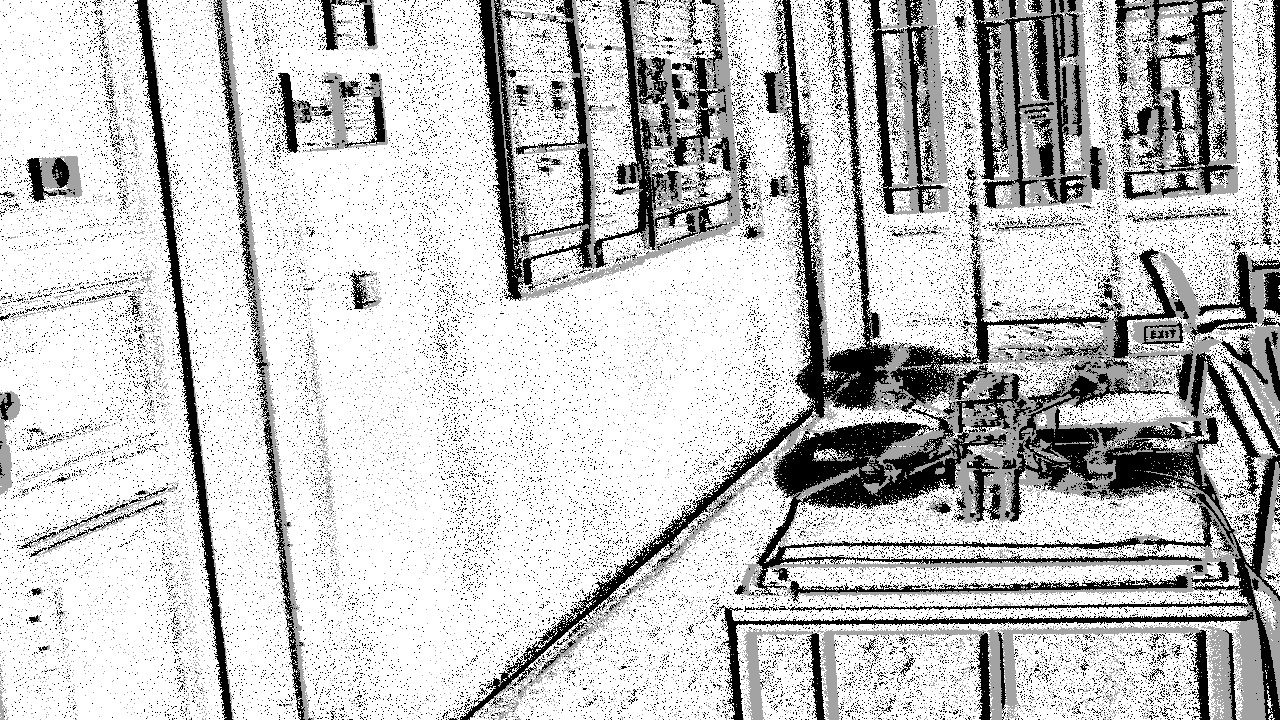} & \includegraphics[width=0.195\linewidth]{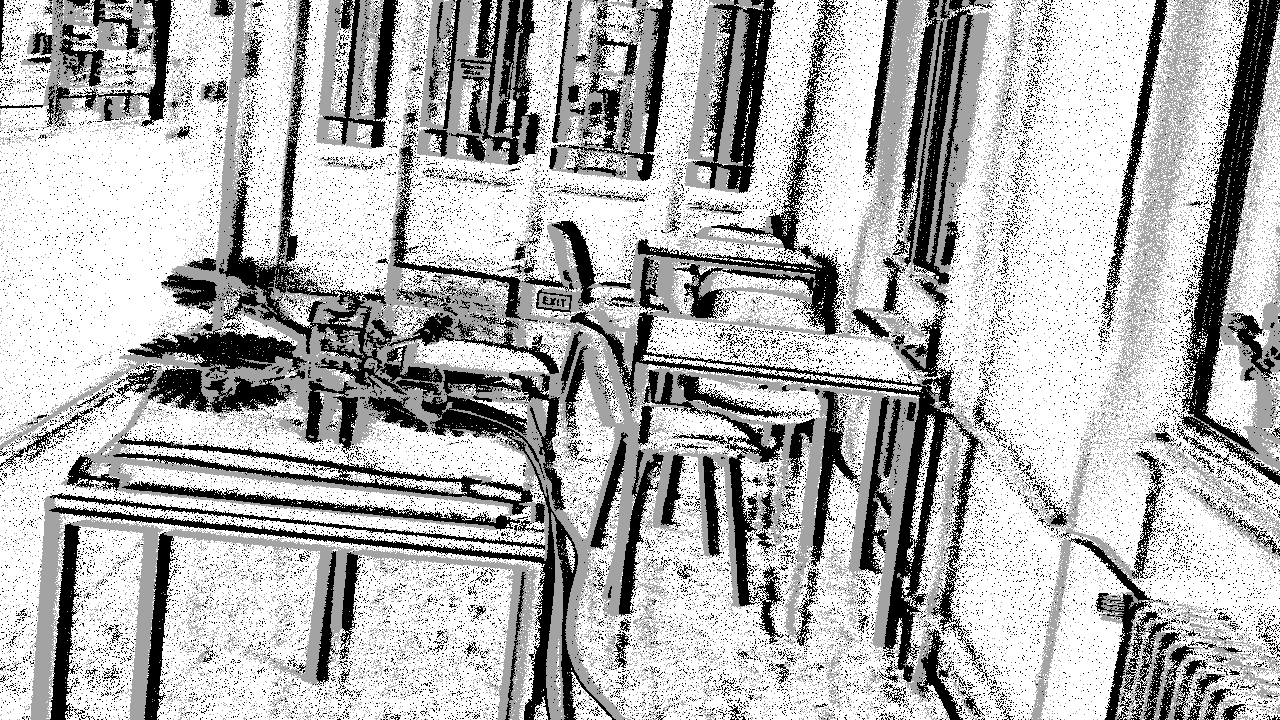}  \\
\hline
\end{tabular}

\caption{Aggregated events from all sequences in the \tquadcopterego{} dataset with camera-to-quadcopter distance of 4 m,
 from the smallest (top row, $M_{\text{ego}}$=1), to the largest (bottom row, $M_\text{ego}$=7) egomotion intensity.
 Events from the first 10 ms of five 1-second intervals are shown.
 Positive events displayed in black, negative in gray.
 }
\label{fig:dataset_ego_sequenced_overview_full_dist4}
\end{figure*}

In Fig.~\ref{fig:dataset_ego_sequenced_overview_full_dist2}, the aggregated events of the first 10 ms of five 1-second intervals are displayed for all recordings taken at a camera-to-quadcopter distance of 2~m. The intensity of egomotion begins at the top with $M_\text{ego} = 1$ and progresses to the last row with $M_\text{ego} = 7$. Fig.~\ref{fig:dataset_ego_sequenced_overview_full_dist4} is structured in the same way and presents recordings for a camera-to-quadcopter distance of 4~m.

\section{AEB-Tracker Parameters Grid Search}\label{sec:aeb_tracker_grid_search}

As described in Sec.~\ref{sec:aebtracker_modifications}, we first tuned the hyperparameters of the tracker qualitatively. Once a well-performing set of parameters was found on the first 10 milliseconds of the recording, we run the grid search described below.

We performed a comprehensive grid search over four AEB-tracker hyperparameters, 72 combinations in total, evaluated across all recordings and channels. The parameters are as follows.
\begin{enumerate}
    \item $rb$ (ring\_buffer\_len): Length of the per‑target event/state ring buffers. Larger values smooth estimates more but add inertia,
    \item $\Delta t$: Integration timestep (seconds) used in the state transition $F$ for $(x,y,\theta)$; in particular $\theta \leftarrow \theta + q\;dt$,
    \item $q_{xy}$: Sets both velocity process‑noise intensities $q_{v_x}$ and $q_{v_y}$ $[\mathrm{px}^2/\mathrm{s}^3]$. Larger values allow faster changes in velocity (more agility) at the cost of more noise/jitter.
    \item $q_q$: Sets the spin‑rate process‑noise intensity $q_q$ $[\mathrm{rad}^2/\mathrm{s}^3]$. Larger values let the angular velocity $q$ vary more between updates.
\end{enumerate}

We select the setting that minimizes mean absolute error (MAE). We also report std/median and mean absolute relative error (MArE).

The best configuration observed in Table~\ref{tab:aeb_grid} is $rb=22$, $\Delta t=2.5$, $q_{xy}=5.0$, $q_q=0.5$, achieving MArE 502.976 and MAE of 38763.807 (aggregated over all recordings and channels).

\section{\helixtrack{} Parameter and Grid Search}\label{sec:helixtrack_params}

We set the parameters of \helixtrack{} with a combination of element-wise search and grid search. The complete set of parameters we searched for and their final values are: $q_\text{jerk}=400$, $\lambda_\phi = 1.0$, $\lambda_r = 1.0$, $\lambda_\text{pol} = 0.25$, $\lambda_\text{band}=5.0$, $\lambda_\text{BTO} = 1.0$, $r_\text{in} = 0.2$, $r_\text{out} = 1.5$, $\gamma_\text{tx} = \gamma_\text{ty} = 6.0$, $\gamma_\text{persp} = 1.0$. For the complete set of all parameters, not just the ones we searched for, see the attached configuration file and codes.

The grid search was performed over $r_\mathrm{in}$ and $r_\mathrm{out}$ parameters, results are in Table~\ref{tab:helixtrack_grid}.

The element-wise search was performed with the following set of parameters:
\begin{itemize}[noitemsep, topsep=2pt]
    \item $q_\mathrm{jerk}$ (7): 50, 100, 200, 300, 400, 500, 600
    \item $\lambda_\phi$ (2): 0, 1
    \item $\lambda_r$ (2): 0, 1
    \item $\lambda_\mathrm{pol}$ (6): 0, 0.25, 0.5, 1, 2, 10
    \item $\lambda_\mathrm{band}$ (5): 0, 1, 5, 10, 20
    \item $\lambda_\mathrm{tip}$ (5): 0, 0.5, 1, 2, 10
    \item \textbf{$\gamma_x{=}\gamma_y$} (7): 1, 2, 3, 4, 5, 6, 7
    \item \textbf{$\gamma_\mathrm{persp}$} (5): 0.01, 0.1, 1, 2, 3
\end{itemize}

In total, \textbf{275} unique experiment configurations were examined, not including the grid search for $r_\mathrm{in}$ and $r_\mathrm{out}$.

\begin{table}
\centering
\label{tab:grid_rin_rout_mae}
\begin{tabular}{r r r@{ $\pm$ }l r}
\toprule
$\mathrm{r_\text{in}}$ & $\mathrm{r_{out}}$ & \multicolumn{2}{c}{MAE (RPM) $\downarrow$} & median \\
\midrule
0.0 & 1.4 & 25897.838 & 64560.838 & 6837.504 \\
0.0 & 1.5 & 31173.431 & 89359.956 & 6623.163 \\
0.0 & 1.6 & 21958.240 & 62330.271 & 7017.397 \\
0.0 & 1.7 & 33919.799 & 133795.000 & 7763.656 \\
0.0 & 1.8 & 17984.723 & 43510.488 & 6545.428 \\
0.0 & 1.9 & 15783.097 & 43241.885 & 7078.281 \\
0.1 & 1.4 & 4356.810 & 3985.889 & 4434.631 \\
0.1 & 1.5 & 4814.578 & 5064.586 & 4995.041 \\
0.1 & 1.6 & 4775.841 & 3918.283 & 6323.458 \\
0.1 & 1.7 & 4951.189 & 4428.663 & 5219.091 \\
0.1 & 1.8 & 4687.726 & 4220.879 & 5265.247 \\
0.1 & 1.9 & 10199.062 & 39763.741 & 5339.970 \\
0.2 & 1.4 & 10659.736 & 44302.860 & 5516.170 \\
\textbf{0.2} & \textbf{1.5} & \textbf{4052.139} & \textbf{3779.222} & \textbf{5200.247} \\
0.2 & 1.6 & 4626.934 & 3815.512 & 6050.976 \\
0.2 & 1.7 & 4574.203 & 4136.607 & 5391.578 \\
0.2 & 1.8 & 4691.997 & 4598.824 & 5790.502 \\
0.2 & 1.9 & 4500.675 & 3892.523 & 5436.960 \\
0.3 & 1.4 & 6470.363 & 12126.974 & 6095.369 \\
0.3 & 1.5 & 4401.135 & 3951.049 & 5629.956 \\
0.3 & 1.6 & 43303.116 & 274072.299 & 5758.449 \\
0.3 & 1.7 & 4846.586 & 5777.108 & 5273.668 \\
0.3 & 1.8 & 5826.368 & 8064.029 & 5181.593 \\
0.3 & 1.9 & 29329.222 & 179744.825 & 5588.616 \\
\bottomrule
\end{tabular}\caption{Grid search over $r_\mathrm{in}$ and $r_\mathrm{out}$ with
$q_\mathrm{jerk}{=}400$, $\lambda_{\phi}{=}1.0$, $\lambda_\mathrm{r}{=}1.0$,
$\lambda_\mathrm{pol}{=}0.25$, $\lambda_\mathrm{band}{=}5.0$, $\lambda_\mathrm{BTO}{=}1.0$,
$\gamma_x{=}\gamma_y{=}6.0$, $\gamma_\mathrm{persp}{=}1.0$.
The best setting (lowest MAE mean) is highlighted in bold. See Sec.~\ref{sec:helixtrack_params}}\label{tab:helixtrack_grid}.
\end{table}

\section{Benchmark Complete Results}\label{sec:benchmark_complete_results}

Across all cells of the Table~\ref{tab:benchmark_complete}, \helixtrack{} is consistently the most accurate. 
It achieves the lowest error in 43/52 (MArE), 46/52 (MAE), and 45/52 (RMSE) cases, while AEB-Tracker wins 8/6/5 and DeepEv 1/0/2, respectively. When we compute medians to mitigate heavy-tailed failures, HelixTrack reduces error dramatically relative to AEB‑Tracker --- MArE: 400.05 $\rightarrow$ 58.95 (-85.3\%), MAE: 29,859.50 $\rightarrow$ 5,200.3 (-82.6\%), RMSE: 51,662.7 $\rightarrow$ 6,332.25 (-87.7\%).
DeepEv’s medians for MAE/RMSE (14,459.0~/~19,504.4) are below the medians of AEB‐Tracker, but its means are much higher due to frequent catastrophic failures. 

\noindent \textbf{Effect of distance $d$.} Performance improves with increased camera–scene distance for HelixTrack. At 
$d=2\mathrm{m}$ vs.\ $4\mathrm{m}$, HelixTrack’s medians drop from 63.65 $\rightarrow$ 29.65 (MArE), 5,233.10 $\rightarrow$ 2,697.55 (MAE), and 6,461.30$\rightarrow$3,358.60 (RMSE). Compared to AEB‑Tracker, the improvement widens at longer distance: 
$d=2\mathrm{m}$: -84.1\%~/~82.5\%~/~87.5\% (MArE/MAE/RMSE); 
$d=4\mathrm{m}$: -94.9\%~/~94.0\%~/~95.5\%. A plausible explanation is that the propeller occupies a smaller angular extent and generates fewer disruptive events at larger $d$, which \helixtrack{} exploits more effectively than the baselines.

\noindent \textbf{Effect of ego‑motion magnitude $M_\text{ego}$.} As motion increases, all methods degrade, but \helixtrack{} degrades more gracefully. For MAE (medians), HelixTrack stays low for $M_\text{ego}=1-3$, and rises for $M_\text{ego}=4-7$ --- still well below the results of AEB‑Tracker. AEB-Tracker shows a pronounced failure at moderate‑to‑high motion (also reflected in RMSE), while DeepEv exhibits a non‑monotonic pattern with moderate medians at certain bins but very heavy tails overall.

\noindent \textbf{Sensitivity to propeller position.} The position of a propeller with respect to other propellers / distractors, matters., \helixtrack{}’s easiest configuration is front‑left. The most challenging is rear‑right, followed by front‑right and rear‑left. Despite this, HelixTrack still secures the majority of wins in every placement: for example, in front‑left it wins 13/13 cells for each metric, and even in rear‑right it wins 12/13 (MAE/RMSE) and 11/13 (MArE).

AEB-Tracker collects isolated wins (\eg, low motion with favorable geometry), and DeepEv takes two RMSE cells in specific configurations. However, these wins are not robust: the corresponding 95th percentiles and maxima reveal frequent large errors and outright failures, especially for DeepEv. In contrast, \helixtrack{} shows low variance in most of the grid.

\section{Loss ablation}\label{sec:loss_ablation_full}

The results of the loss term ablation in Table~\ref{tab:loss_ablation} show that removing either the phase-geometry coupling \((\lambda_\phi=0)\) or the radial centering term \((\lambda_r=0)\) leads to failure or order‑of‑magnitude error  increases. 
For example, at 2~m, \(M_{\text{ego}}=2\), MAE jumps from 90.1~$\pm$~16.9 (full loss) to 9327.1~$\pm$~175.7 when \(\lambda_r=0\); at 4~m, \(M_{\text{ego}}=3\) it rises from 3449.3~$\pm$~3961.3 to 9775.1~$\pm$~129.5 when \(\lambda_r=0\). 
Tying the Gauss-Newton pose update to phase residuals \(r_\phi\) and keeping events centered on the blade ring via \(r_r\) are important for stabilizing geometry and phase jointly (Sec.~\ref{sec:gn_loss}, Eq.~\ref{eq:loss}).

\begin{table*}[ht]
    \setlength{\tabcolsep}{5pt}
\fontsize{9.6pt}{11.3pt}\selectfont
\begin{tabular}{ c @{ } c @{ } r@{ $\pm$ }l r@{ $\pm$ }l r@{ $\pm$ }l r@{ $\pm$ }l r@{ $\pm$ }l r@{ $\pm$ }l}
\toprule
$d$ & $M_{\text{ego}}$ & \multicolumn{2}{c}{all} & \multicolumn{2}{c}{$\lambda_\phi=0$} & \multicolumn{2}{c}{$\lambda_r=0$} & \multicolumn{2}{c}{$\lambda_\text{pol}=0$} & \multicolumn{2}{c}{$\lambda_\text{band}=0$} & \multicolumn{2}{c}{$\lambda_\text{BTO}=0$} \\
\midrule
$2$ m  & 1 & \bfseries 92.8 & \bfseries 18.4 & 98.8 & 21.5 & \multicolumn{2}{c}{$F$} & 5701.9 & 3786.9 & \bfseries 92.8 & \bfseries 18.4 & 92.8 & 18.4 \\
$2$ m  & 2 & 90.1 & 16.9 & 2347.5 & 2766.6 & 9327.1 & 175.7 & 2103.3 & 4024.6 & \bfseries 89.7 & \bfseries 16.6 & 90.1 & 16.9 \\
$2$ m  & 3 & 680.2 & 974.9 & 3883.1 & 7372.4 & 8430.9 & 1589.7 & 7438.6 & 1700.5 & 1319.7 & 1522.6 & \bfseries 612.5 & \bfseries 841.1 \\
$2$ m  & 4 & 7212.5 & 332.2 & 7807.8 & 3357.2 & 8720.0 & 1068.0 & 7427.7 & 950.7 & \multicolumn{2}{c}{$F$} & \bfseries 6821.1 & \bfseries 475.0 \\
$2$ m  & 5 & 7488.4 & 2162.5 & \bfseries 6718.4 & \bfseries 2246.1 & \multicolumn{2}{c}{$F$} & 7142.7 & 1866.7 & 7452.8 & 1056.6 & 8434.2 & 974.4 \\
$2$ m  & 6 & 7981.4 & 1705.1 & 8902.2 & 307.7 & 8210.8 & 2099.4 & 8182.7 & 1695.0 & 7997.8 & 1939.9 & \bfseries 7868.9 & \bfseries 899.4 \\
$2$ m  & 7 & \bfseries 6497.4 & \bfseries 1534.1 & \multicolumn{2}{c}{$F$} & 9194.3 & 240.5 & 7873.5 & 1519.3 & \multicolumn{2}{c}{$F$} & 8032.8 & 573.4 \\
$4$ m  & 1 & \bfseries 1721.3 & \bfseries 3263.3 & 2176.4 & 4171.0 & 9056.5 & 197.7 & \multicolumn{2}{c}{$F$} & 1783.8 & 3388.9 & 2078.1 & 3976.9 \\
$4$ m  & 2 & 2210.3 & 4234.0 & 2319.0 & 4439.3 & 9144.2 & 202.0 & 4762.0 & 4837.7 & 2263.6 & 4340.8 & \bfseries 1926.3 & \bfseries 3666.0 \\
$4$ m  & 3 & 3449.3 & 3961.3 & 4670.2 & 3608.3 & 9775.1 & 129.5 & 5903.4 & 4194.7 & 4409.7 & 5026.2 & \bfseries 3402.7 & \bfseries 3953.5 \\
$4$ m  & 4 & \bfseries 2152.1 & \bfseries 4126.9 & 4795.1 & 5431.6 & 9617.7 & 224.8 & 8055.4 & 1451.6 & 4582.6 & 5204.8 & 2301.9 & 4426.5 \\
$4$ m  & 5 & 6969.2 & 4582.9 & \bfseries 5955.7 & \bfseries 3973.0 & 9573.0 & 229.7 & 8630.4 & 2381.1 & 6913.1 & 4549.0 & 6705.8 & 4456.2 \\
$4$ m  & 6 & \bfseries 6132.8 & \bfseries 842.3 & 8208.1 & 1158.6 & \multicolumn{2}{c}{$F$} & 7688.5 & 623.2 & 6251.3 & 2277.5 & 7990.8 & 1147.7 \\
\midrule
\multicolumn{2}{l}{\textbf{Average}} & \bfseries 4052.1 & \bfseries 2135.0 & \multicolumn{2}{c}{$F$} & \multicolumn{2}{c}{$F$} & 7070.4 & 3146.3 & \multicolumn{2}{c}{$F$} & 4335.2 & 1955.8 \\
\bottomrule
\end{tabular}\caption{Loss term ablation. Mean Absolute Error (MAE) of RPM (lower is better) under 4‑fold cross‑validation on the \emph{\tquadcopterego{}} (\tquadcopteregoshort{}) dataset, reported for two camera–target distances (2~m / 4~m) and seven egomotion levels \(M_{\text{ego}}\). ``All'' uses the full batch objective in Eq.~(\ref{eq:loss}) with: phase–geometry alignment \(r_\phi\), radial centering \(r_r\), polarity alignment \(r_{\text{pol}}\), soft annulus barriers \(r_{\text{in}}, r_{\text{out}}\), and Balanced Tip Occupancy \(\ell_{\text{BTO}}\). Each ablation zeroes the indicated coefficient \((\lambda_\phi, \lambda_r, \lambda_{\text{pol}}, \lambda_{\text{band}}, \lambda_{\text{BTO}})\). 
    \tquadcopteregoshort{} ground-truth RPM has $\mu\pm\sigma = 11193.5\pm3540.2$.
    Values are mean~$\pm$~std.}\label{tab:loss_ablation}
\end{table*}

Disabling the soft band-edge barriers \((\lambda_{\text{band}}=0)\) frequently causes failures in the harder settings
(average column reports failure; e.g., 2~m, \(M_{\text{ego}}=4\) and 2~m, \(M_{\text{ego}}=7\) both fail), indicating that the annulus constraints are necessary to keep the optimization from leaking inside/outside the informative blade band when background motion and perspective effects are large.

Dropping the polarity term \((\lambda_{\text{pol}}=0)\) degrades the overall average from 4052.1~$\pm$~2135.0 to 7070.4~$\pm$~3146.3 and can sharply increase error in challenging cases (e.g., 4~m, \(M_{\text{ego}}=4\): 2152.1~$\pm$~4126.9 to 8055.4~$\pm$~1451.6), suggesting that softly aligning polarity clusters with the expected blade edge improves robustness.

Removing BTO \((\lambda_{\text{BTO}}=0)\) changes the overall average modestly. This is consistent with BTO’s role in discouraging degenerate ``only‑inside/only‑outside'' fits rather than driving the solution on its own.

Overall, the \emph{phase} and \emph{radial} data terms provide the backbone for stable pose/phase coupling. The \emph{band} constraints and \emph{polarity} alignment significantly boost robustness under distance and high \(M_{\text{ego}}\); and \emph{BTO} offers a small but reliable safety margin against occupancy imbalance. Together, these components are necessary to keep the asynchronous EKF and batched GN updates coherent.

\begin{table*}[ht]
    \centering
\fontsize{8.5pt}{9.0pt}\selectfont
\begin{tabular}{ c @{ } c @{ } r @{ } l @{ } r r r r @{\hspace{0.5em}} r r r r r}
\toprule
& & & & \multicolumn{3}{c}{MArE (\%) $\downarrow$} & \multicolumn{3}{c}{MAE (RPM) $\downarrow$} & \multicolumn{3}{c}{RMSE (RPM) $\downarrow$} \\
\cmidrule{5-7}\cmidrule(l){8-10}\cmidrule(l){11-13}
$d$ & $M_{\text{ego}}$ & \multicolumn{2}{c}{propeller} & ours & \footnotesize AEB~\cite{wang_asynchronous_2024} & \footnotesize DeepEv~\cite{messikommer_data-driven_2025} & ours & \footnotesize AEB~\cite{wang_asynchronous_2024} & \footnotesize DeepEv~\cite{messikommer_data-driven_2025} & ours & \footnotesize AEB~\cite{wang_asynchronous_2024} & \footnotesize DeepEv~\cite{messikommer_data-driven_2025} \\
\midrule
\multirow{28}*{$2$ m\hspace{0.3em}} & \multirow{4}*{1} & rear & right & \bfseries 1.3 & 61.4 & 8564.8 & 93.7 \ok{} & 5980.8 & 652423.7 & 215.9 & 7186.8 & 785896.2\\
 &  & rear & left & 1.3 & \bfseries 0.9 & 402.4 & 81.4 \ok{} & 73.0 & 32090.3 & 241.4 & 241.2 & 46466.0\\
 &  & front & left & \bfseries 1.5 & 35.3 & 4132.0 & 118.4 \ok{} & 2487.0 & 303141.3 & 194.1 & 3302.9 & 359054.3\\
 &  & front & right & \bfseries 1.2 & 65.5 & 94.6 & 77.8 \ok{} & 5650.8 & 8759.2 & 186.3 & 7673.3 & 9700.4\\
\cmidrule{2-13}
 & \multirow{4}*{2} & rear & right & \bfseries 1.2 & 52.1 & 9523.5 & 85.0 \ok{} & 4637.6 & 749060.3 & 178.9 & 5549.9 & 904654.7\\
 &  & rear & left & 1.3 & \bfseries 1.2 & 4811.7 & 87.0 \ok{} & 112.9 & 357676.3 & 277.2 & 457.4 & 455119.6\\
 &  & front & left & \bfseries 1.5 & 49.8 & 478.5 & 114.0 \ok{} & 3075.8 & 28118.0 & 223.5 & 3879.2 & 68394.5\\
 &  & front & right & \bfseries 1.1 & 34.5 & 7458.3 & 74.3 \ok{} & 2035.8 & 562988.0 & 160.9 & 3367.3 & 787219.5\\
\cmidrule{2-13}
 & \multirow{4}*{3} & rear & right & \bfseries 1.2 & 1198.7 & 5013.6 & 85.1 \ok{} & 102482.5 & 386360.6 & 223.2 & 192935.8 & 437954.9\\
 &  & rear & left & 31.7 & \bfseries 15.0 & 2059.8 & 2137.8 \fail{} & 1696.4 & 151319.8 & 4518.8 & 3482.9 & 184206.2\\
 &  & front & left & \bfseries 4.2 & 467.5 & 5236.2 & 230.2 \ok{} & 37691.4 & 376231.1 & 763.8 & 81016.3 & 448776.7\\
 &  & front & right & \bfseries 4.8 & 1347.9 & 195.7 & 267.9 \ok{} & 110056.2 & 14595.8 & 1031.9 & 202948.8 & 18910.5\\
\cmidrule{2-13}
 & \multirow{4}*{4} & rear & right & \bfseries 80.2 & 1241.2 & 12474.4 & 7634.3 \fail{} & 93995.4 & 934583.8 & 9184.5 & 182884.8 & 1141131.4\\
 &  & rear & left & 72.2 & \bfseries 24.5 & 8678.3 & 7135.2 \fail{} & 2828.9 & 674615.2 & 8699.8 & 4782.1 & 824811.5\\
 &  & front & left & \bfseries 79.3 & 410.3 & 89.2 & 7249.8 \fail{} & 29227.1 & 8417.0 & 8825.0 & 68684.3 & 9362.8\\
 &  & front & right & 76.3 & \bfseries 59.5 & 115.2 & 6830.5 \fail{} & 5933.1 & 9027.6 & 8459.0 & 11580.7 & 10408.5\\
\cmidrule{2-13}
 & \multirow{4}*{5} & rear & right & \bfseries 62.3 & 531.3 & 106.7 & 6347.2 \fail{} & 41415.3 & 9244.0 & 7842.7 & 79321.0 & 10241.1\\
 &  & rear & left & \bfseries 100.9 & 480.0 & 117.9 & 9689.7 \fail{} & 30491.9 & 10624.2 & 10899.5 & 67097.7 & 12239.0\\
 &  & front & left & \bfseries 65.0 & 389.8 & 99.3 & 5045.3 \fail{} & 33458.3 & 8963.0 & 6266.7 & 69659.2 & 9836.3\\
 &  & front & right & \bfseries 97.7 & 1122.0 & 398.5 & 8871.4 \fail{} & 77151.0 & 32523.2 & 9710.6 & 111267.4 & 56823.0\\
\cmidrule{2-13}
 & \multirow{4}*{6} & rear & right & \bfseries 93.6 & 726.2 & 98.1 & 8928.2 \fail{} & 56816.8 & 9461.9 & 9874.9 & 72370.2 & 10223.4\\
 &  & rear & left & \bfseries 97.2 & 593.1 & 108.1 & 9221.9 \fail{} & 41752.5 & 9932.7 & 10100.1 & 56316.4 & 11232.3\\
 &  & front & left & \bfseries 89.5 & 885.2 & 103.3 & 8283.4 \fail{} & 61735.0 & 9163.9 & 9331.8 & 72061.8 & 10279.6\\
 &  & front & right & \bfseries 57.2 & 833.4 & 96.8 & 5492.3 \fail{} & 58682.1 & 8881.7 & 6655.9 & 72261.9 & 9697.5\\
\cmidrule{2-13}
 & \multirow{4}*{7} & rear & right & \bfseries 72.2 & 609.8 & 10417.8 & 7194.7 \fail{} & 44713.1 & 780191.6 & 8292.1 & 51904.1 & 894151.1\\
 &  & rear & left & 119.1 & 617.6 &\bfseries 97.0 & 8328.7 \fail{} & 45454.3 & 9180.8 & 10071.3 & 51421.3 & 10022.4\\
 &  & front & left & \bfseries 75.7 & 218.3 & 96.4 & 5341.7 \fail{} & 16859.6 & 8811.8 & 6817.1 & 28904.6 & 9686.0\\
 &  & front & right & \bfseries 72.2 & 248.4 & 3638.0 & 5124.5 \fail{} & 17502.1 & 253240.6 & 6262.9 & 29616.6 & 315329.1\\
\midrule
\multirow{24}*{$4$ m\hspace{0.3em}} & \multirow{4}*{1} & rear & right & \bfseries 73.1 & 285.5 & 834.5 & 6616.3 \fail{} & 21029.7 & 57801.3 & 8517.8 & 46759.1 & 73901.3\\
 &  & rear & left & \bfseries 1.2 & 57.2 & 89.6 & 76.4 \ok{} & 6038.7 & 8341.9 & 158.4 & 20115.0 & 9422.8\\
 &  & front & left & \bfseries 1.4 & 48.9 & 88.7 & 104.8 \ok{} & 4676.4 & 7853.7 & 255.3 & 16769.4 & 8925.9\\
 &  & front & right & \bfseries 1.3 & 165.4 & 835.4 & 87.8 \ok{} & 13041.8 & 58054.3 & 319.4 & 36853.1 & 74833.5\\
\cmidrule{2-13}
 & \multirow{4}*{2} & rear & right & 91.1 & \bfseries 43.1 & 93.1 & 8561.2 \fail{} & 3847.8 & 8671.9 & 9537.1 & 4496.9 & 9642.0\\
 &  & rear & left & \bfseries 1.3 & 37.4 & 3242.1 & 85.7 \ok{} & 3529.9 & 241975.2 & 273.2 & 5020.2 & 334534.1\\
 &  & front & left & \bfseries 1.5 & 72.1 & 93.8 & 112.9 \ok{} & 6545.9 & 8536.3 & 254.1 & 17753.2 & 9595.1\\
 &  & front & right & \bfseries 1.3 & 77.3 & 111.7 & 81.4 \ok{} & 4335.3 & 8969.8 & 248.9 & 17551.4 & 10429.0\\
\cmidrule{2-13}
 & \multirow{4}*{3} & rear & right & 70.9 & \bfseries 58.8 & 95.8 & 5770.0 \fail{} & 5904.9 & 9773.3 & 7008.1 & 11479.0 & 10725.7\\
 &  & rear & left & \bfseries 1.3 & 46.8 & 1801.4 & 85.0 \ok{} & 3960.0 & 140787.4 & 211.7 & 4845.4 & 181653.8\\
 &  & front & left & \bfseries 1.3 & 303.1 & 1937.2 & 109.4 \ok{} & 21349.0 & 146814.9 & 210.8 & 51357.9 & 174482.0\\
 &  & front & right & 79.8 & \bfseries 39.7 & 97.4 & 7832.8 \fail{} & 4091.3 & 9594.5 & 9394.6 & 5306.3 & 10465.4\\
\cmidrule{2-13}
 & \multirow{4}*{4} & rear & right & \bfseries 83.5 & 1194.9 & 95.2 & 8342.3 \fail{} & 101987.0 & 9517.6 & 9491.6 & 201102.2 & 10384.6\\
 &  & rear & left & \bfseries 1.1 & 915.1 & 5251.8 & 77.6 \ok{} & 76275.0 & 426131.2 & 222.5 & 179370.3 & 537350.6\\
 &  & front & left & \bfseries 1.4 & 1392.2 & 389.6 & 111.6 \ok{} & 98307.7 & 26933.9 & 288.4 & 201185.3 & 46853.8\\
 &  & front & right & \bfseries 1.2 & 1166.3 & 13.3 & 76.9 \ok{} & 86077.0 & 693.4 & 209.4 & 191384.9 & 2370.2\\
\cmidrule{2-13}
 & \multirow{4}*{5} & rear & right & \bfseries 95.3 & 994.6 & 97.8 & 9688.6 \fail{} & 83816.5 & 9942.3 & 10614.2 & 127547.1 & 10817.5\\
 &  & rear & left & \bfseries 93.9 & 1043.3 & 94.8 & 9316.3 \fail{} & 76882.9 & 9332.0 & 10400.6 & 133235.2 & 10351.1\\
 &  & front & left & \bfseries 1.7 & 1092.8 & 3339.1 & 119.1 \ok{} & 81598.8 & 260136.5 & 241.1 & 126119.1 & 310936.4\\
 &  & front & right & \bfseries 88.3 & 1126.9 & 95.6 & 8752.7 \fail{} & 93336.6 & 9351.7 & 9711.9 & 134050.9 & 10281.2\\
\cmidrule{2-13}
 & \multirow{4}*{6} & rear & right & \bfseries 62.1 & 862.2 & 167.5 & 6778.5 \fail{} & 68419.4 & 14322.1 & 8015.9 & 97991.7 & 20098.2\\
 &  & rear & left & \bfseries 57.6 & 958.9 & 3772.0 & 5547.2 \fail{} & 71680.8 & 290173.7 & 6746.4 & 101283.8 & 359570.3\\
 &  & front & left & \bfseries 71.5 & 995.9 & 11718.5 & 6929.4 \fail{} & 76394.7 & 918479.2 & 8522.6 & 105034.9 & 1024071.7\\
 &  & front & right & \bfseries 60.3 & 856.0 & 130.4 & 5276.0 \fail{} & 68594.4 & 12657.2 & 6397.8 & 102062.7 & 18155.9\\
\bottomrule
\end{tabular}\caption{Benchmark, \tquadcopterego{} (\tquadcopteregoshort{}). 
    For each presented method and setting (camera-target distance $d$, egomotion level $M_\text{ego}$), the average shaft-RPM mean absolute relative error (MArE), mean absolute error (MAE), and root mean squared error (RMSE) are reported over 4‑fold cross‑validation across the four propellers (one training propeller per fold).
    Both AEB-Tracker~\cite{wang_asynchronous_2024} and DeepEv~\cite{messikommer_data-driven_2025} methods were modified to allow RPM estimation (see Sec.~\ref{sec:existing_trackers_modifications}). \tquadcopteregoshort{} ground-truth RPM has $\mu\pm\sigma\!=\!11193\pm3540$, $\min$ $917$, $\max$ $15768$.
    $M_\text{ego} = 1$ roughly means ``slow camera movement'', while $M_\text{ego} = 5$ means ``fast camera movement'' (see Fig.~\ref{fig:dataset_ego_sequenced_overview}). 
    As $M_\text{ego}$ increases, the camera experiences stronger egomotion, leading to increase of the number of events generated in the background.
    }\label{tab:benchmark_complete}
\end{table*}

\begin{table*}[ht]
    \centering
\fontsize{7.5pt}{8.5pt}\selectfont
\begin{tabular}{rrrr r@{ $\pm$ }l r r@{ $\pm$ }l r}
\toprule
$rb$ &  $\Delta t$ & $q_{xy}$ &   $q_q$ & \multicolumn{2}{c}{MArE (\%)} $\downarrow$ & median &   \multicolumn{2}{c}{MAE (RPM)}  $\downarrow$ & median \\
\midrule
22 &   2 & 0.2 & 0.25 &     496.419 &    402.402 &       428.662 &  39026.775 &  35704.452 &  27149.123 \\
22 &   2 & 0.2 &  0.5 &     538.978 &    532.238 &       335.813 &  41383.706 &  40575.234 &  24252.682 \\
22 &   2 & 0.2 &    1 &     650.760 &    610.255 &       580.914 &  49669.433 &  45197.183 &  44281.303 \\
22 &   2 & 0.2 &    2 &     697.412 &    668.144 &       537.904 &  53110.765 &  50873.870 &  39796.295 \\
22 &   2 &   1 & 0.25 &     587.555 &    478.454 &       411.100 &  39086.051 &  29527.280 &  26570.284 \\
22 &   2 &   1 &  0.5 &     544.618 &    510.582 &       401.498 &  41860.949 &  39341.854 &  30303.074 \\
22 &   2 &   1 &    1 &     615.645 &    577.918 &       553.496 &  47675.376 &  44307.127 &  42378.293 \\
22 &   2 &   1 &    2 &     694.885 &    662.955 &       577.022 &  52290.860 &  49815.795 &  42225.829 \\
22 &   2 &   5 & 0.25 &     510.618 &    388.804 &       397.497 &  39608.748 &  29903.648 &  29931.931 \\
22 &   2 &   5 &  0.5 &     506.789 &    472.468 &       400.054 &  38952.001 &  36484.847 &  29859.506 \\
22 &   2 &   5 &    1 &     637.420 &    590.571 &       597.896 &  48856.215 &  44965.901 &  44515.639 \\
22 &   2 &   5 &    2 &     664.801 &    623.023 &       618.496 &  50721.102 &  47445.002 &  46145.155 \\
22 & 2.5 & 0.2 & 0.25 &     599.204 &    405.217 &       428.662 &  39154.585 &  31844.812 &  27149.123 \\
22 & 2.5 & 0.2 &  0.5 &     541.204 &    535.575 &       335.813 &  41498.917 &  40751.855 &  24252.682 \\
22 & 2.5 & 0.2 &    1 &     656.356 &    608.960 &       626.036 &  50086.419 &  45131.954 &  46723.292 \\
22 & 2.5 & 0.2 &    2 &     700.457 &    672.777 &       537.904 &  53245.557 &  51105.783 &  39796.295 \\
22 & 2.5 &   1 & 0.25 &     582.944 &    578.059 &       404.861 &  39748.436 &  29535.466 &  26218.892 \\
22 & 2.5 &   1 &  0.5 &     546.194 &    513.463 &       401.498 &  41776.215 &  39284.539 &  30303.074 \\
22 & 2.5 &   1 &    1 &     619.680 &    584.761 &       553.496 &  48030.629 &  44997.063 &  42378.293 \\
22 & 2.5 &   1 &    2 &     696.192 &    666.004 &       577.022 &  52501.734 &  50255.610 &  42225.829 \\
22 & 2.5 &   5 & 0.25 &     511.004 &    391.626 &       362.751 &  39689.789 &  30294.397 &  29321.017 \\
\bfseries 22 &\bfseries  2.5 &\bfseries    5 &\bfseries   0.5 &\bfseries      502.976 &\bfseries     462.531 &\bfseries        400.054 & \bfseries  38763.807 & \bfseries  35860.448 & \bfseries  29859.506 \\
22 & 2.5 &   5 &    1 &     635.099 &    588.028 &       597.896 &  48705.017 &  44782.988 &  44515.639 \\
22 & 2.5 &   5 &    2 &     667.087 &    627.251 &       618.496 &  50801.457 &  47640.824 &  46145.155 \\
22 &   3 & 0.2 & 0.25 &     537.926 &    339.939 &       388.690 &  40797.556 &  25179.970 &  32766.673 \\
22 &   3 & 0.2 &  0.5 &     527.948 &    330.960 &       482.435 &  39522.724 &  23976.905 &  36248.414 \\
22 &   3 & 0.2 &    1 &    1572.458 &   1439.686 &      1071.014 & 118593.768 & 108749.883 &  77281.079 \\
22 &   3 & 0.2 &    2 &    1729.390 &   1080.637 &      1360.253 & 130145.474 &  80259.290 & 110851.655 \\
22 &   3 &   1 & 0.25 &     510.972 &    337.896 &       387.988 &  39089.727 &  25213.444 &  29948.624 \\
22 &   3 &   1 &  0.5 &     522.499 &    311.308 &       439.585 &  39133.979 &  22301.822 &  32878.498 \\
22 &   3 &   1 &    1 &    1623.040 &   1543.883 &       999.876 & 122206.230 & 115777.126 &  76014.113 \\
22 &   3 &   1 &    2 &    1819.391 &   1076.640 &      1690.690 & 137646.186 &  80980.053 & 131528.893 \\
22 &   3 &   5 & 0.25 &     528.778 &    327.504 &       393.982 &  39501.077 &  24184.137 &  30356.179 \\
22 &   3 &   5 &  0.5 &     620.653 &    478.684 &       477.139 &  46596.672 &  35946.071 &  36226.416 \\
22 &   3 &   5 &    1 &    1775.582 &   1666.385 &      1223.899 & 133680.214 & 125213.233 &  87960.026 \\
22 &   3 &   5 &    2 &    1868.451 &   1043.308 &      1612.597 & 140173.319 &  78815.962 & 128391.421 \\
33 &   2 & 0.2 & 0.25 &     532.076 &    542.658 &       313.137 &  40817.897 &  42290.509 &  23969.480 \\
33 &   2 & 0.2 &  0.5 &     637.817 &    621.567 &       359.301 &  49911.826 &  48999.356 &  28496.135 \\
33 &   2 & 0.2 &    1 &     722.345 &    689.412 &       658.855 &  54920.949 &  53254.591 &  47154.050 \\
33 &   2 & 0.2 &    2 &     801.631 &    764.621 &       785.360 &  61030.529 &  59447.827 &  53700.933 \\
33 &   2 &   1 & 0.25 &     539.595 &    519.382 &       358.879 &  41270.586 &  40010.637 &  28240.227 \\
33 &   2 &   1 &  0.5 &     628.204 &    611.694 &       431.265 &  48692.147 &  47599.814 &  29409.219 \\
33 &   2 &   1 &    1 &     756.455 &    724.681 &       638.275 &  57933.012 &  56141.403 &  47235.750 \\
33 &   2 &   1 &    2 &     764.616 &    729.480 &       607.390 &  58027.384 &  56269.086 &  46369.599 \\
33 &   2 &   5 & 0.25 &     594.884 &    547.826 &       405.281 &  45212.128 &  41849.225 &  29930.397 \\
33 &   2 &   5 &  0.5 &     610.897 &    618.423 &       408.550 &  47681.510 &  48051.032 &  30428.694 \\
33 &   2 &   5 &    1 &     767.419 &    722.414 &       736.989 &  58523.058 &  55635.939 &  54296.780 \\
33 &   2 &   5 &    2 &     730.990 &    714.165 &       662.121 &  55352.472 &  55168.873 &  49709.951 \\
33 & 2.5 & 0.2 & 0.25 &     543.074 &    522.441 &       313.137 &  41890.431 &  41373.974 &  24346.624 \\
33 & 2.5 & 0.2 &  0.5 &     639.535 &    619.235 &       401.370 &  50020.820 &  48816.418 &  30022.854 \\
33 & 2.5 & 0.2 &    1 &     727.481 &    698.813 &       668.252 &  55288.661 &  54127.417 &  47256.535 \\
33 & 2.5 & 0.2 &    2 &     796.948 &    755.814 &       785.360 &  60555.252 &  58419.579 &  53700.933 \\
33 & 2.5 &   1 & 0.25 &     532.410 &    506.342 &       358.879 &  41036.682 &  39777.801 &  28240.227 \\
33 & 2.5 &   1 &  0.5 &     628.658 &    612.606 &       431.265 &  48665.147 &  47531.149 &  29409.219 \\
33 & 2.5 &   1 &    1 &     743.469 &    715.875 &       624.827 &  57122.147 &  55425.865 &  45484.744 \\
33 & 2.5 &   1 &    2 &     768.513 &    746.872 &       601.761 &  58471.489 &  57907.022 &  45485.936 \\
33 & 2.5 &   5 & 0.25 &     580.653 &    485.517 &       490.506 &  44371.759 &  38134.757 &  35223.893 \\
33 & 2.5 &   5 &  0.5 &     611.000 &    617.790 &       408.550 &  47776.018 &  48238.017 &  30428.694 \\
33 & 2.5 &   5 &    1 &     760.575 &    742.399 &       681.596 &  58266.260 &  57509.586 &  48980.242 \\
33 & 2.5 &   5 &    2 &     727.158 &    722.614 &       597.588 &  54981.420 &  55610.817 &  46677.389 \\
33 &   3 & 0.2 & 0.25 &     780.803 &    691.614 &       539.572 &  56835.266 &  50675.677 &  40003.480 \\
33 &   3 & 0.2 &  0.5 &     711.595 &    534.618 &       538.977 &  52049.844 &  39029.338 &  41273.869 \\
33 &   3 & 0.2 &    1 &    1334.438 &   1139.468 &       903.428 &  98922.204 &  83349.423 &  68183.272 \\
33 &   3 & 0.2 &    2 &    1800.403 &   1083.555 &      1388.857 & 133261.705 &  80987.728 & 101892.368 \\
33 &   3 &   1 & 0.25 &     797.649 &    670.641 &       528.780 &  58814.987 &  49205.799 &  37539.753 \\
33 &   3 &   1 &  0.5 &     752.721 &    578.674 &       586.153 &  56295.285 &  43346.004 &  42539.879 \\
33 &   3 &   1 &    1 &    1371.708 &   1131.488 &       885.898 & 100812.434 &  83139.604 &  65690.880 \\
33 &   3 &   1 &    2 &    1882.050 &   1109.445 &      1727.309 & 138601.952 &  83147.089 & 128224.109 \\
33 &   3 &   5 & 0.25 &     705.389 &    592.629 &       559.554 &  52950.606 &  43547.516 &  42432.900 \\
33 &   3 &   5 &  0.5 &     713.524 &    573.891 &       534.925 &  52982.647 &  42424.617 &  39459.804 \\
33 &   3 &   5 &    1 &    1401.862 &   1125.218 &       929.897 & 102499.465 &  82047.435 &  64251.491 \\
33 &   3 &   5 &    2 &    1843.070 &   1090.486 &      1565.962 & 136415.532 &  80887.240 & 122453.756 \\
\bottomrule
\end{tabular}\caption{AEB-Tracker grid search results. MAE and MArE reported as $\mu\pm\sigma$. See Sec.~\ref{sec:aeb_tracker_grid_search} for description.}\label{tab:aeb_grid}
\end{table*}


\end{document}